%% file: arxiv.tex
\documentclass{ARXIV}   %

\usepackage{latexsym}
\usepackage{graphicx}
\usepackage{multicol,multirow}
\usepackage{amsmath,amssymb,amsfonts}
\usepackage{mathrsfs}
\usepackage{amsthm}
\usepackage{rotating}
\usepackage{appendix}
\usepackage[authoryear]{natbib}
\usepackage{ifpdf}
\usepackage[T1]{fontenc}
\usepackage{times}
\usepackage{sourcesanspro}
\usepackage{newtxmath}
\usepackage{textcomp}%
\usepackage{xcolor}%
\usepackage{hyperref}
\usepackage{booktabs}

\usepackage{comment}
\makeatletter
\ifcsname arxiv\endcsname
  \excludecomment{mybackmatter}
\else
  \includecomment{mybackmatter}
\fi
\makeatother

\articletype{RESEARCH ARTICLE}
\jname{Data-Centric Engineering}
\jyear{2023}
\jdoi{10.1017/dce.2023.7}

\newcommand{\dint}{\,\mathrm{d}}
\newcommand{\Transp}{\top}

\newcommand{\gaussian}[2]{\mathcal{N}\left(#1,#2\right)}

\newcommand{\expq}{\exp_\mathsf{q}}

\DeclareMathOperator{\diag}{diag}
\renewcommand{\mid}[0]{\,|\,}

\usepackage{xspace}
\newcommand{\eg}{\textit{e.g.}\xspace}
\newcommand{\ie}{\textit{i.e.}\xspace}

\newcommand{\cf}{\textit{cf.}\xspace}

\usepackage{subcaption}

\usepackage{fontawesome5}

\definecolor{primarycolor}{RGB}{0, 93, 141}

\usepackage{tikz,pgfplots}
\pgfplotsset{compat=newest}
\usetikzlibrary{arrows.meta,decorations.markings}
\usetikzlibrary{calc}
\usetikzlibrary{external}
\tikzsetexternalprefix{tikzfigures/}
\usepgfplotslibrary{fillbetween}
\usepgfplotslibrary{statistics}
\newlength\figureheight 
\newlength\figurewidth 

  \usepackage[ruled, vlined, nofillcomment]{algorithm2e}
  \usepackage{algorithmic}  
  
  \SetCommentSty{mycommfont}
  \SetAlgoCaptionSeparator{:}

\usepackage[capitalise,noabbrev,nameinlink]{cleveref}
\crefname{equation}{Eq.}{Eqs.}
\crefname{appsec}{appendix}{appendices}

\usepackage{framed}
\usepackage{calc}

\begin{document}

\begin{Frontmatter}

\title[Rao-Blackwellized SLAM]
{Rao-Blackwellized Particle Smoothing for \\ Simultaneous Localization and Mapping}

\author*[1]{Manon Kok}\email{m.kok-1@tudelft.nl}\orcid{0000-0002-2441-2240}
\author[2]{Arno Solin}\orcid{0000-0002-0958-7886}
\author[3]{Thomas B. Sch\"on}\orcid{0000-0001-5183-234X}

\authormark{Kok, Solin, and Sch\"on}

\address*[1]{\orgdiv{Delft Center for Systems and Control}, \orgname{Delft University of Technology}, \orgaddress{\street{Mekelweg 2}, \postcode{2628 CD Delft}, \country{the Netherlands}}}

\address[2]{\orgdiv{Department of Computer Science}, \orgname{Aalto University}, \orgaddress{\street{Konemiehentie 2}, \postcode{FI-00076 Aalto}, \country{Finland}}}

\address[3]{\orgdiv{Department of Information Technology}, \orgname{Uppsala University}, \orgaddress{\street{Box 337}, \postcode{SE-751 05 Uppsala}, \country{Sweden}}}

\received{xx xx xxxx}
\revised{xx xx xxxx}
\accepted{xx xx xxxx}


\abstract{
Simultaneous localization and mapping (SLAM) is the task of building a map representation of an unknown environment while at the same time using it for positioning. A probabilistic interpretation of the SLAM task allows for incorporating prior knowledge and for operation under uncertainty. Contrary to the common practice of computing point estimates of the system states, we capture the full posterior density through approximate Bayesian inference. This dynamic learning task falls under state estimation, where the state-of-the-art is in sequential Monte Carlo methods that tackle the forward filtering problem. In this paper, we introduce a framework for probabilistic SLAM using particle smoothing that does not only incorporate observed data in current state estimates, but it also back-tracks the updated knowledge to correct for past drift and ambiguities in both the map and in the states. Our solution can efficiently handle both dense and sparse map representations by Rao-Blackwellization of conditionally linear and conditionally linearized models. We show through simulations and real-world experiments how the principles apply to radio (BLE/Wi-Fi), magnetic field, and visual SLAM. The proposed solution is general, efficient, and works well under confounding noise.
}

\begin{policy}[Impact Statement]
Simultaneous localization and mapping (SLAM) methods constitute an essential part of the backbone for autonomous robots, vehicles, and aircraft that operate in an environment with only on-board sensing, without external sensor support. The methods proposed in this paper help characterize the uncertainty of the pose and the map representation, and contribute to making SLAM methods capable of handling unexpected real-world conditions such as changing weather and lighting, signal attenuation, and confounding artifacts in sensor data.
\end{policy}

\end{Frontmatter}

\section{Introduction}
\label{sec:introduction}
In ego-motion estimation, simultaneous localization and mapping (SLAM) is a ubiquitous approach of simultaneously estimating the time-varying pose of a robot, person, or object and a map of the environment~\citep{Durrant-Whyte+Bailey:2006,Bailey+Durrant-Whyte:2006,cadenaCCLSNRL:2016,stachnissLT:2016}. SLAM is widely used for autonomous robots, vehicles, and aircraft. The wide adoption is due to the general nature of the concept behind SLAM, which makes it applicable across different sensor modalities and use case scenarios. Still, different setups for SLAM typically motivate different representations of the map data structure, thereby constraining how inference and learning is done under the SLAM paradigm. 

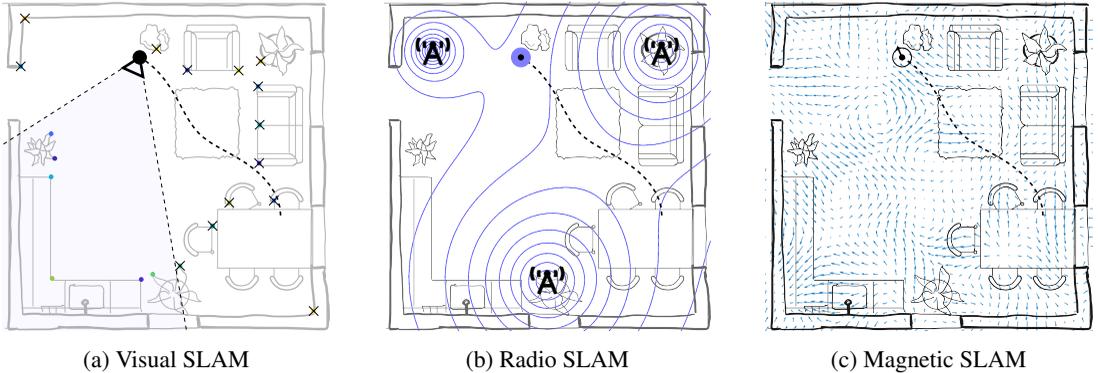
\begin{figure}
\centering
\tikzexternaldisable
\setlength{\figurewidth}{.6\textwidth}
\setlength{\figureheight}{\figurewidth}
\def\datapath{./fig/}
\begin{subfigure}{.3\textwidth}
  \centering
  \tikzsetnextfilename{fig-1a}
  \scalebox{.5}{\input{./fig/visual.tex}}
  \caption{Visual SLAM}
  \label{fig:intro-a}
\end{subfigure}
\hfill
\begin{subfigure}{.3\textwidth}
  \centering
  \tikzsetnextfilename{fig-1b}
  \scalebox{.5}{\input{./fig/radio.tex}}
  \caption{Radio SLAM}
\end{subfigure}
\hfill
\begin{subfigure}{.3\textwidth}
  \centering
  \tikzsetnextfilename{fig-1c}
  \scalebox{.5}{\input{./fig/magnetic.tex}}
  \caption{Magnetic SLAM}
\end{subfigure}\\
\caption{Illustration of the different SLAM modalities used throughout this paper. (a)~Visual SLAM uses a sparse map representation of distinctive corner points observed as projections onto the camera frustum; (b)~radio SLAM either uses a sparse map of the radio emitter locations or models the RSSI anomalies as a dense field; (c)~magnetic SLAM  leverages anomalies in the local magnetic vector field} %
\label{fig:intro}
\end{figure}

In {\bf visual SLAM}, the map is traditionally represented by a set of sparse landmark points which are observed as projections by a camera rigidly attached to the moving coordinate frame \citep[\eg,][]{Davison+Reid+Molton+Statsse:2007,Hartley+Zisserman:2004}. Alternatively, dense representations are used, which can take the form of continuous surfaces  \citep[\eg,][]{Whelan+Leutenegger+SalasMoreno+Glocker+Davison:2015,Bloesch+Czarnowski+Clark+Leutenegger+Davison:2018,Kerl+Sturm+Cremers:2013}. These two extremes motivate the alternative views we bring into the map representation also in other sensor modalities. In {\bf magnetic SLAM} \citep[\eg,][]{Kok+Solin:2018}, the map is inherently dense as it characterizes the anomalies of the Earth's magnetic vector field which are observed by a three-axis magnetometer. In {\bf radio-based SLAM} \citep[\eg,][]{Ferris+Fox+Lawrence:2007}, the map can either represent point-wise radio emitter sources (typically Wi-Fi base stations or Bluetooth low-energy, BLE, beacons) or a dense anomaly map of receiver signal strength indicator (RSSI) values. To this end, dense maps of these anomaly fields have been constructed using Gaussian process regression \citep{Ferris+Fox+Lawrence:2007}. In this work we consider both sparse and dense SLAM, as illustrated in \cref{fig:intro}, with the intent of presenting general principles for SLAM that extend across sensor modalities. Our interest lies in settings where the map is static (not changing over time), but initially unknown.
 
The SLAM problem is inherently nonlinear. To perform inference and learning under the SLAM paradigm, all approaches resort to approximate inference of some kind. Typically, methods either approximate the estimation problem using linearization and Gaussian approximations or using Monte Carlo sampling methods. More specifically, when considering online recursive algorithms, the former approximation results in EKF-SLAM~\citep{Bailey+Nieto+Guivant+Stevens+Nebot:2006,Barrau+Bonnabel:2015,Mourikis+Roumeliotis:2007}, while the latter results in particle filter-based SLAM. The most well-known particle filter SLAM algorithm is FastSLAM~\citep{Montemerlo+Thrun+Koller+Wegbreit:2002}. An extension of this algorithm is FastSLAM 2.0~\citep{Montemerlo+Thrun+Koller+Wegbreit:2003}, which recognizes that the bootstrap particle filter implementation from~\citet{Montemerlo+Thrun+Koller+Wegbreit:2002} suffers from particle degeneracy, and hence it does not capture the full posterior distribution. In this work, we will instead overcome this limitation using particle smoothing. 

In recent years, it has become more common to consider smoothing problems for SLAM, where all information acquired up to the current time is back-tracked through the model to ensure better global consistency of the map and the past movement. This is closely related to graphSLAM \citep{Grisetti+Kummerle+Stachniss+Burgard:2010,Thrun+Montemerlo:2006} which turns the problem into a global (sparse) optimization task. Even though these smoothing approaches allow for updated knowledge to correct for past drift and ambiguity in the map and path (bundle-adjustment), they are currently almost exclusively linearization-based as opposed to sampling-based. This means that rather than characterizing the full posterior, only point estimates are considered. In this work, we present a new approach for particle smoothing, \ie smoothing using sequential Monte Carlo sampling. We show that it outperforms both particle filtering and extended Kalman filtering approaches. It is specifically more robust to unexpected real-world conditions, for instance the case that the initial map estimates are quite off or that the sensors are not properly calibrated. 

We develop a solution with similarities to FastSLAM and speed up the computations of our particle smoother by exploiting the conditionally linear substructure in the problem using Rao-Blackwellization~\citep{Schon+Gustafsson+Nordlund:2005,gustafssonGBFJKN:2002,doucetFMR:2000}. To this end, we leverage progress in the field of Rao-Blackwellized particle smoothing~\citep{Svensson+Schon+Lindsten:2014,Svensson+Schon+Kok:2015}. For these existing algorithms, however, it is not possible to assume a constant map which prevents using these algorithms for standard SLAM problems. To the best of our knowledge, the only existing work on particle smoothing for SLAM is~\citet{Berntorp+Nordh:2014}, which assumes that the map is slowly time-varying. This can, however, deteriorate the map quality, and have a negative effect on the estimation accuracy, especially in challenging applications. Our algorithm overcomes these limitations by explicitly assuming a constant map. An alternative interpretation of our SLAM problem is therefore in terms of joint state and parameter estimation, see \citet{WigrenWLWS:2022} for a recent tutorial on the parameter estimation problem. This problem has traditionally been handled using maximum likelihood estimation~\citep{Schon+Wills+Ninness:2011}, and a bit more than a decade ago, an interesting Bayesian solution was derived by \citet{AndrieuDH:2010}. We will derive a fully Bayesian solution to the SLAM problem, which allows for prior information to be included also on the map. This becomes particularly important when building Gaussian process maps. Our algorithm can be seen as a special case of~\cite{wigrenRML:2019}, where for the specific form of the SLAM problem turns out to take a particularly nice form. 

The contributions of this paper can be summarized as follows. {\em (i)}~We present a general framework for SLAM which is agnostic to the map representation; {\em (ii)}~We derive a Rao-Blackwellized particle smoothing method for SLAM, which exploits the conditionally linear or conditionally linearized substructure and circumvents typical pitfalls in particle filtering approaches; {\em (iii)}~We demonstrate the applicability of the proposed SLAM method in a range of simulated and real-world examples. A reference implementation of our method and code to reproduce the experiments can be found on \url{https://github.com/manonkok/Rao-Blackwellized-SLAM-smoothing}.

\section{Background}
In this work, we introduce a framework for probabilistic SLAM using particle smoothing. As our approach extends existing work on sequential Monte Carlo (SMC) methods---specifically particle filters---for localization and SLAM, we will introduce this concept in \cref{sec:background-smc}. We consider both sparse (feature-based) as well as dense maps. In terms of dense maps, we specifically focus on maps represented using Gaussian processes (GPs). Because of this, we briefly introduce the concept of GPs and their use to construct maps of the environment in \cref{sec:background-gp}.

\subsection{Sequential Monte Carlo for localization and SLAM}
\label{sec:background-smc}
\begin{figure}
\tikzsetnextfilename{fig-2}
\begin{tikzpicture}

  \setlength{\figurewidth}{.45\textwidth}

  \node[minimum width=\figurewidth,minimum height=0.5625\figurewidth,rounded corners=2pt,draw=none,path picture={
    \node at (path picture bounding box.center) {
      \includegraphics[width=\figurewidth]{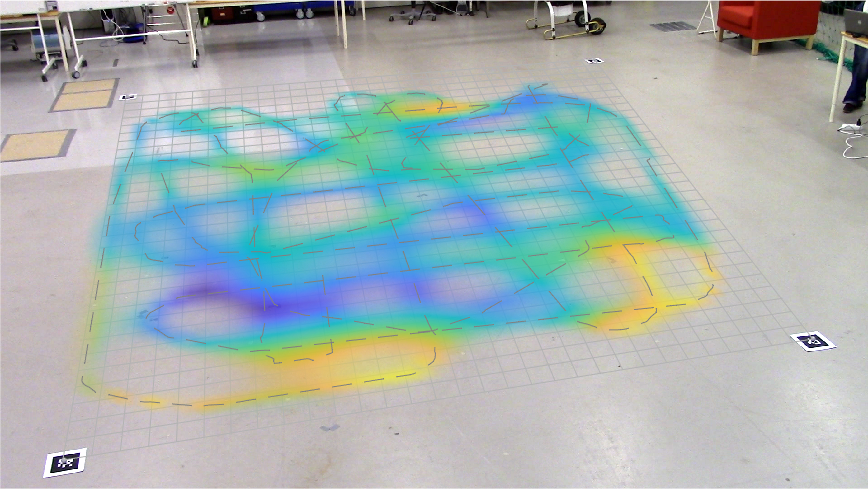}
    };}] (mag) at (0,0) {};

  \node[anchor=north west] (magx) at (mag.south west) {\includegraphics[width=.333\figurewidth]{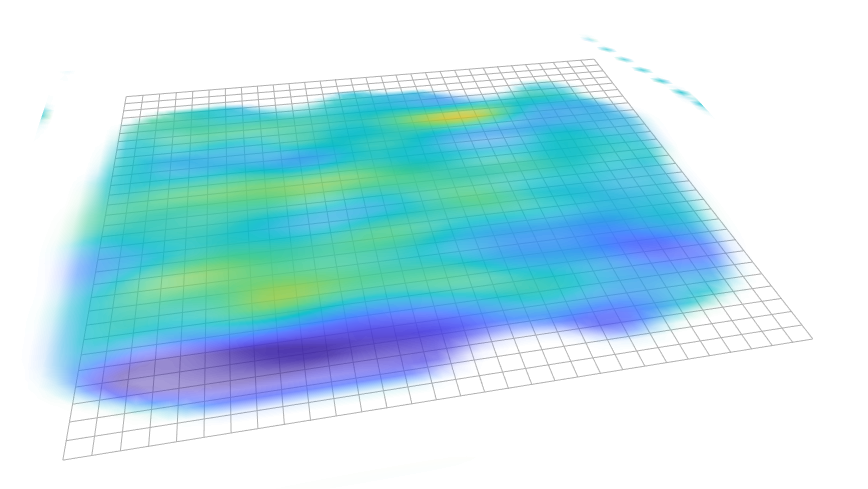}};
  \node[anchor=north] (magy) at (mag.south) {\includegraphics[width=.333\figurewidth]{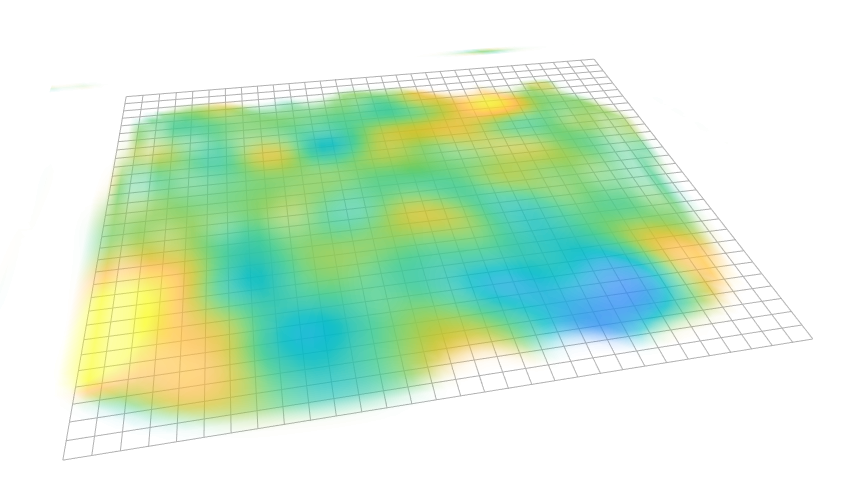}};
  \node[anchor=north east] (magz) at (mag.south east) {\includegraphics[width=.333\figurewidth]{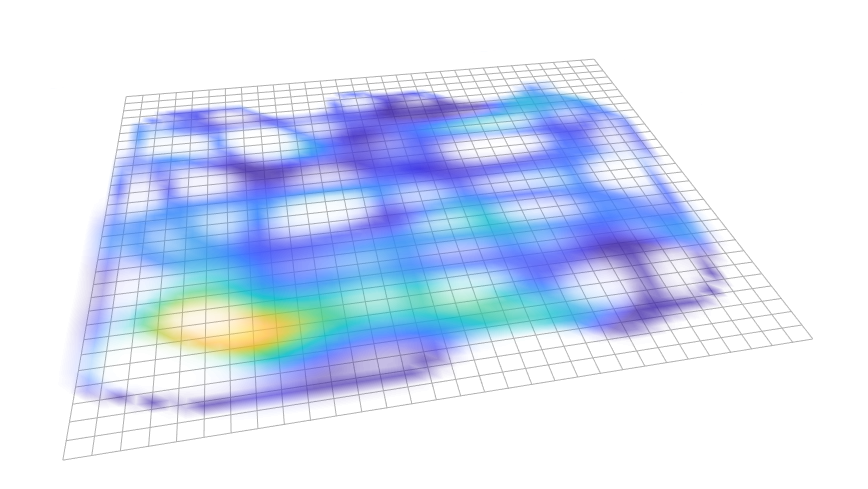}};

  \node[below of=magx,yshift=6pt] {\footnotesize $x$}; 
  \node[below of=magy,yshift=6pt] {\footnotesize $y$}; 
  \node[below of=magz,yshift=6pt] {\footnotesize $z$};

  \node[right of=magz,xshift=16pt,inner sep=0] (cbar) {\includegraphics[width=2mm,height=12mm]{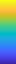}};
  \node[anchor=south west,inner sep=0] at (cbar.south east) {\tiny ~~$-60\,\mu\text{T}$};
  \node[anchor=north west,inner sep=0] at (cbar.north east) {\tiny ~~$30\,\mu\text{T}$};
  \node[left of=cbar,rotate=90,yshift=-2.2em] {\tiny Components};

  \node[above of=cbar,yshift=36pt,inner sep=0] (cbar) {\includegraphics[width=2mm,height=22mm]{fig/parula}};
  \node[anchor=south west,inner sep=0] at (cbar.south east) {\tiny ~~$10\,\mu\text{T}$};
  \node[anchor=north west,inner sep=0] at (cbar.north east) {\tiny ~~$90\,\mu\text{T}$};
  \node[left of=cbar,rotate=90,yshift=-2.2em] {\tiny Magnetic field strength};

  \setlength{\figurewidth}{.33\figurewidth}
  \setlength{\figureheight}{.5625\figurewidth}

  \foreach \x/\s [count=\i from 0] in {001/0,021/20,041/40,061/60,101/100,121/120,141/140,161/160,181/180}
    \node[anchor=north west,text width=\figurewidth,align=center,inner sep=0] (\i) at 
        ($(mag.north east) + (.1\textwidth,0) +
        ({\figurewidth*mod(\i,3)},{-1.6\figureheight*int((\i)/3)})$)
		{\includegraphics[width=\figurewidth]{fig/robot-frame-\x} \\[-4pt] \footnotesize$t=\s\,$s};

  \node[minimum width=2cm,circle,draw=black!90,path picture={
    \node at (path picture bounding box.center){
      \includegraphics[width=2cm]{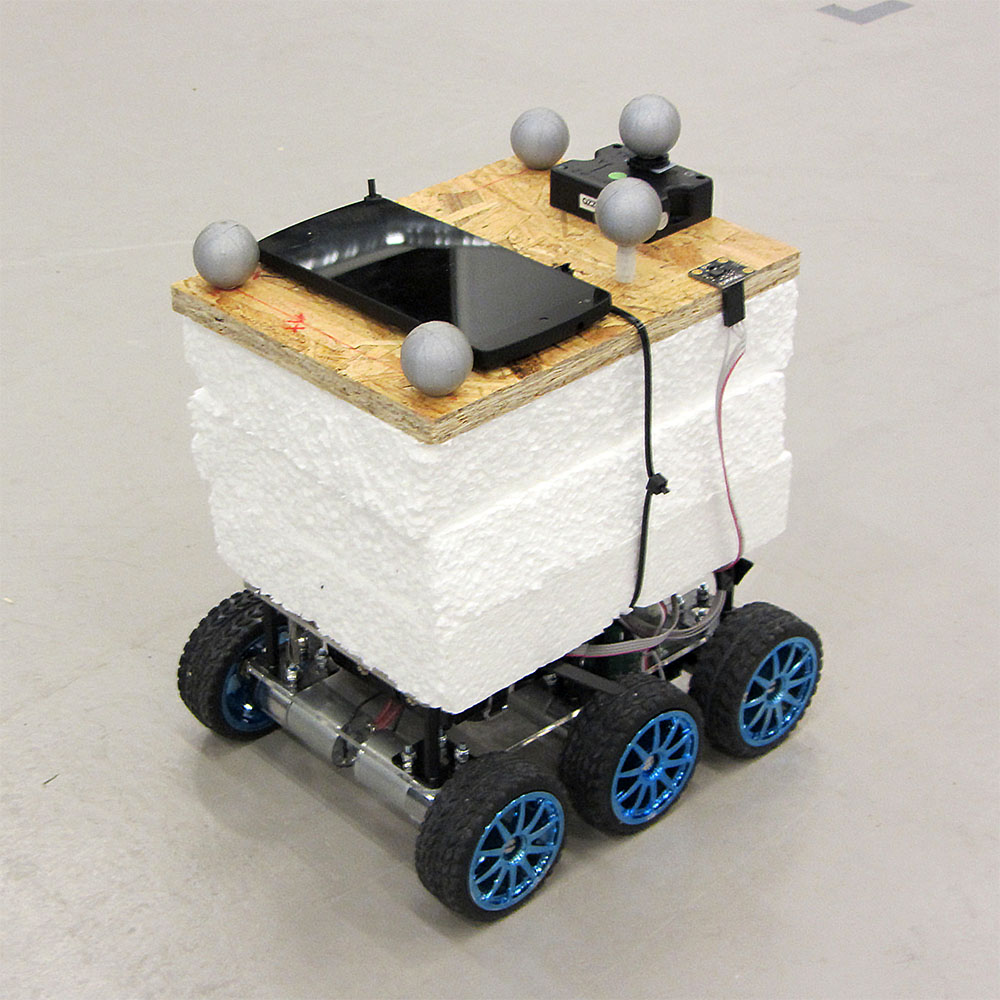}
    };}] at (0.25\textwidth,.14\textwidth) {};

  \node[align=center,text width=.45\textwidth] at (0,.175\textwidth) 
    {{\bf Magnetic mapping} \\ \small\it Magnetic map from known \\ sensor positions};
  \node[align=center,text width=.45\textwidth] at (.55\textwidth,.175\textwidth) 
    {{\bf Magnetic localization} \\ \small\it Movement path estimation by matching \\ magnetic measurements to the magnetic map };

\end{tikzpicture}
\caption{Left:~A magnetic anomaly map mapped by a robot (smartphone for scale) equipped with a 3-axis magnetometer. Map opacity follows the marginal variance (uncertainty), and mapping (training) paths shown by dashed lines. Right:~Localization by map matching. Current estimate is characterized by a particle cloud, the dashed line shows the ground-truth, and the solid line the weighted mean path}%
\label{fig:LAM}
\end{figure}

In sequential Monte Carlo, a posterior distribution is approximated using samples, also referred to as particles. Assuming for simplicity that we are interested in estimating the posterior $p(x_{1:t} \mid y_{1:t})$ of a set of time-varying states $x_{1:t} = ( x_1, x_2, \hdots, x_t )$ given a set of measurements $y_{1:t}$, the posterior is approximated as
\begin{equation}
\hat{p}(x_{1:t} \mid y_{1:t}) = \sum_{i=1}^N w^i_t  \delta_{x_{1:t}^i} (x_{1:t}),
\end{equation}
where $N$ denotes the number of particles, $w^i_t$ denotes the importance weight of each particle $x_{1:t}^i$ at time $t$ and $\delta_{x_{1:t}^i}$ denotes the Dirac measure at $x_{1:t}^i$. Using SMC to approximate the posterior distribution $p(x_{1:t} \mid y_{1:t})$ is also synonymously called particle filtering. The idea was proposed in the beginning of the 1990s \citep{Gordon:1993,Kitagawa:1993,StewartM:1992} and there are by now several tutorials \citep{DoucetJ:2011,NaessethLS:2019fnt} and textbooks \citep{ChopinP:2020,Sarkka:2013} available on the subject. 
The use of a particle filter for localization in a known magnetic field map is illustrated in \cref{fig:LAM} (estimated magnetic anomaly map to the left, localization result to the right). Here, a map of the strongly position-dependent indoor magnetic field is used to give information about how likely each particle is. The approach illustrated in the figure follows \citet{solin2016terrain}, where the idea of {\em map matching} is cast as positioning in a vector-valued magnetic field map. As can be seen, the posterior is highly multi-modal to start with, but becomes unimodal when time passes and a unique trajectory through the magnetic map can be reconstructed. The fact that SMC can handle multi-modal distributions makes it especially well-suited for both localization and SLAM problems.  

The challenge in SMC is to ensure that the particles represent the full posterior distribution. Because of this, the particles are regularly resampled, discarding the particles with low weight and replicating the particles with high weights. This, however, results in the phenomenon of particle degeneracy, where all particles at a certain time $t$ descend from the same ancestor some time $t - s$ in the past. To overcome this issue, a number of particle smoothing algorithms have been developed---see~\citet{Lindsten+Schon:2013} for a tutorial.

Using SMC for higher-dimensional state spaces typically requires a larger number of particles and hence increases the computational complexity. Because of this, a number of algorithms have been developed to exploit a conditionally linear substructure in the models and treat this using a Kalman filter \citep{AndrieuD:2002,Chen:2000,Schon+Gustafsson+Nordlund:2005}. This has resulted in the so-called marginalized or Rao-Blackwellized particle filters. The current work is inspired by these approaches and develops a Rao-Blackwellized particle smoothing algorithm for SLAM. 

\subsection{Representing maps using Gaussian process priors}
\label{sec:background-gp}
Gaussian processes \citep[GPs,][]{Rasmussen+Williams:2006} represent distributions over functions. A GP $h(x)$ is fully specified by its mean function $\mu(x)$ and covariance function $\kappa(x,x')$ as 
\begin{equation}
h(x) \sim \mathcal{GP}(\mu(x),\kappa(x,x')).
\end{equation}
The covariance function (kernel) can encode prior information about the properties of the function, such as continuity, smoothness, or physical properties of a random field. Observational data $\mathcal{D} = \{(x_t,y_t)\}_{t = 1}^T$ is coupled with the GP prior through a likelihood model. The Gaussian likelihood model reduces to observing noise-corrupted versions of the Gaussian process
\begin{equation}
y_t = h(x_t) + \varepsilon_{t},
\label{eq:gp-MeasModel}
\end{equation}
where $y_t \in \mathbb{R}^{n_y}$, and $\varepsilon_{t} \sim \mathcal{N}(0,\sigma^2 \mathcal{I}_{n_y})$ denotes zero-mean Gaussian i.i.d.\ measurement noise with covariance $\sigma^2 \mathcal{I}_{n_y}$, with $\mathcal{I}_{n_y}$ being an identity matrix of size $n_y$. Under a Gaussian (conjugate) likelihood, GP regression can be conducted in closed form.

In recent years, GPs have been used to construct dense maps of the magnetic field or of the Wi-Fi/BLE RSSI field to be used for localization~\citep{Ferris+Fox+Lawrence:2007,Kok+Solin:2018,visetHK:2022,coulinGGJF:2021,vallivaaraHKR:2010}. In these approaches, the field is modeled as a GP and the input to the GP, $x$, is in this case the position. The GP map is used to make predictions of the field at previously unseen locations. Since the GP map also provides information about the uncertainty of these predictions, such approaches are  suitable to handle situations of uninformative maps at certain locations. In \cref{fig:LAM}, we visualize the magnitude of the magnetic field predicted using a GP that has learned the magnetic field based on measurements along the visualized path. The transparency visualizes the marginal variance, where larger transparency indicates a more uncertain prediction. 

One of the main challenges with GPs is their computational complexity which scales cubically with the number of data points. Because of this, there is a large literature dedicated to the construction and use of computationally efficient GP regression \citep[see, \eg,][]{QuinoneroCandela+Rasmussen:2005,hensman2013gaussian}. 
One method that is particularly relevant for our work is the reduced-rank GP regression approach by \citet{Solin+Sarkka:2020} which rewrites
the GP model in terms of a Hilbert space representation. This approach approximates the covariance function in terms of an eigenfunction expansion of the Laplace operator in a confined domain as
\begin{equation}
\kappa(x,x') \approx \sum_{j = 1}^m S(\sqrt{\lambda_j}) \phi_j(x) \phi_j(x') = \Phi \Lambda \Phi^\Transp,
\label{eq:Hilbert-approx}
\end{equation}
where $\phi_j(x)$ denotes the orthonormal eigenfunctions and $\lambda_j$ the corresponding eigenvalues. Furthermore, $S( \cdot )$ is the spectral density function of the kernel. For rectangular domains, the expressions for the eigenfunctions and eigenvalues can be computed in closed-form \citep{Solin+Sarkka:2020}. For more general domains, they can be approximated numerically \citep[see, \eg,][]{solinK:2019}. The spectral density of the kernel can be computed in closed form for any stationary kernel \citep{Solin+Sarkka:2020,Rasmussen+Williams:2006}. The approximation in \cref{eq:Hilbert-approx} has been shown to converge to the exact GP solution when the number of eigenfunctions as well as the size of the domain tend towards infinity. However, it is typically a good approximation already for a relatively small number of eigenfunctions as long as we do not come too close to the boundary. 

Using the approximation in \cref{eq:Hilbert-approx}, allows us to write the measurement model~\cref{eq:gp-MeasModel} as
\begin{equation}
y_t \approx \Phi(x_t) \theta + \varepsilon_{t},
\label{eq:gp-MeasModel-approx}
\end{equation}
with $\Phi(x_t) = \begin{pmatrix} \phi_1(x_t) & \phi_2(x_t) & \cdots & \phi_m(x_t) \end{pmatrix}^\Transp$, and to write the GP prior in terms of a mean $\theta_0$ and covariance 
\begin{equation}
P_0 = \text{diag}\begin{pmatrix} S(\lambda_1) & S(\lambda_2) & \cdots & S(\lambda_m)\end{pmatrix}.
\label{eq:prior-gp-basis}
\end{equation}
The posterior can be computed recursively as new data arrives as~\citep{Sarkka:2013}
\begin{subequations}
\begin{align}
\hat{\theta}_{t} &= \hat{\theta}_{t-1} + K_t ( y_t - \hat{y}_t ), &\hat{y}_t &= \Phi(x_t) \hat{\theta}_{t-1},  \\
P_{t} &= P_{t-1} - K_t S_t K_t^\Transp, & K_t &= P_{t-1} (\Phi(x_t))^\Transp S_t^{-1}, \qquad S_t = \Phi(x_t) P_{t-1} (\Phi(x_t))^\Transp + \sigma^2 \mathcal{I}.
\end{align}
\label{eq:recursiveUpdateMap}%
\end{subequations}%
 In \cref{eq:recursiveUpdateMap}, we can recognize that the recursive map updating is now posed as a recursive linear parameter estimation problem. We will use this representation for our SLAM approach with dense maps. 

\section{Models}
\label{sec:models}
We are interested in jointly estimating time-varying states $x_{1:T}$ at times $t = 1,2, \hdots, T$ and a constant map, which we denote in terms of constant parameters $\theta$. The sensor pose, $x_t$ at least consists of the sensor's position $p_t^\text{n}$ and its orientation $q^\text{nb}_{t}$. The superscript $n$ in $p_t^\text{n}$ specifically indicates that we represent the position in a fixed navigation frame $n$. We focus both on planar localization where $p_t^\text{n} \in \mathbb{R}^2$ and on full 3D localization, \ie\ $p_t^\text{n} \in  \mathbb{R}^3$. The double superscript on $q^\text{nb}_{t}$ specifically indicates that we consider a rotation from a body-fixed frame $b$ to the navigation frame $n$. The origin of this body-fixed frame $b$ coincides with that of the sensor triads and the axes of the body-fixed frame are aligned with the sensor axes. We represent the orientation $q^\text{nb}_{t}$ using unit quaternions when considering full 3D localization while we represent it using a single heading angle when considering planar localization. Additional states can be included in $x_t$ such as the sensor's velocity or sensor biases. 

We model the dynamics of the state as 
\begin{equation}
x_{t+1} =  f(x_t, u_t, w_t), 
\label{eq:dynModel}
\end{equation}
where $w_t$ denotes the process noise and $u_t$ denotes a possible input to the dynamic model which can be used to incorporate available odometry. Note the generality of the model \cref{eq:dynModel} which can contain both nonlinearities as well as non-Gaussian noise. Hence, all dynamic models that are commonly used in SLAM can be written in the form of \cref{eq:dynModel}.

We additionally assume that we have prior information on the map $\theta$ as
\begin{equation}
p(\theta) = \mathcal{N}(\theta \, ; \, \mu_\theta, P_\theta). 
\label{eq:prior}
\end{equation}
This prior information is particularly crucial when doing SLAM with GP-based maps, since the prior on the map is equal to the GP prior. Other examples are prior information available from previous data collections. 

We furthermore assume that if the states $x_{1:T}$ would be known, inferring $\theta$ would be a linear or an almost linear estimation problem. In other words, we assume that there is a conditionally linear substructure or a conditionally approximately linear substructure in the measurement model 
\begin{equation}
y_t = C(x_t) \theta + \varepsilon_t, \qquad \text{or} \qquad y_t \approx C(x_t) \theta + \varepsilon_t,
\label{eq:measModel}
\end{equation}
and that the measurement noise $\varepsilon_t$ is Gaussian i.i.d.\ white noise with $\varepsilon_t \sim \mathcal{N}(0,\Sigma)$. The model \cref{eq:measModel} is fairly general and diverse measurement models typically considered in SLAM scenarios can be written in this form. In \cref{sec:methods} we will present our smoothing algorithm for SLAM that will assume the model structures in \cref{eq:dynModel,eq:prior,eq:measModel}. To highlight the generality of the method, or, in other words, to highlight the generality of \cref{eq:measModel} in the context of SLAM, in this work we will focus on three types of models: visual, radio and magnetic SLAM. We will introduce specific measurement models for these three cases in \cref{sec:magSLAM,sec:radioSLAM,sec:visualSLAM}. In \cref{sec:magSLAM} we will first introduce a measurement model for magnetic field SLAM that we have used in our previous work \citep{Kok+Solin:2018}. In \cref{sec:radioSLAM} we will then show that for dense radio SLAM a similar model structure can be obtained. Both the measurement models in \cref{sec:magSLAM,sec:radioSLAM} are conditionally linear. In \cref{sec:visualSLAM} we will subsequently introduce a widely-used model for visual SLAM that has a conditionally approximately linear substructure. 

\subsection{Dense magnetic SLAM}
\label{sec:magSLAM}
In magnetic SLAM, at each time instance $t$ we receive a three-dimensional measurement $y_{\text{m},t}$ of the magnetic field. These three dimensions are not independent since the ambient magnetic field follows laws of physics captured by Maxwell's equations. One way to take this into account is to model the magnetic field measurements as~\citep{Kok+Solin:2018}
\begin{subequations}
\begin{align}
\varphi(p) &\sim \mathcal{GP}(0,\kappa_\text{lin.}(p,p') + \kappa_\text{SE}(p,p')), \label{eq:measModel-mag-gp} \\
y_{\text{m},t} &= R (q_t^\text{bn}) \nabla \varphi(p) \big|_{p=p^\text{n}_t} + \varepsilon_{\text{m},t}, \label{eq:measModel-mag-meas}
\end{align}%
\label{eq:measModel-mag}%
\end{subequations}%
where $\varepsilon_{\text{m},t} \sim \mathcal{N}(0,\sigma_\text{m}^2 \mathcal{I}_3)$ and 
\begin{equation}
\kappa_\text{lin.}(p,p') = \sigma_{\text{lin.}}^2 p^\Transp p', \qquad \kappa_\text{SE}(p,p') = \sigma_\text{f}^2 \exp \bigg( - \tfrac{\| p - p' \|_2^2 }{2 \ell^2}\bigg).
\label{eq:gp-se}
\end{equation} 
In \cref{eq:measModel-mag-gp}, $\varphi(\cdot)$ denotes a scalar potential field, the GP depends on the position $p$ and $\kappa(p,p')$ represents the covariance function~\citep{Rasmussen+Williams:2006}. The linear term $\kappa_\text{lin.}(p,p')$ in the covariance function models the local Earth's magnetic field while the squared exponential term $\kappa_\text{SE}(p,p')$ models the magnetic field anomalies. The Gaussian process in \cref{eq:measModel-mag-gp} has three hyperparameters: the lengthscale $\ell$, the magnitude of the squared exponential term $\sigma_\text{f}$ and the magnitude of the linear term $\sigma_\text{lin.}$. The measurements $y_{\text{m},t}$ of the magnetic field in~\cref{eq:measModel-mag-meas} are the gradient of the scalar potential field, expressed in the body frame $b$. We use the notation $R(q^{\text{nb}}_t)$ to denote the rotation matrix representation of the state $q^\text{nb}_t$ and $R(q^{\text{nb}}_t) = (R (q^{\text{bn}}_t))^\Transp$. 

Making use of the reduced-rank Gaussian process regression introduced in \cref{sec:background-gp}, via \cref{eq:gp-MeasModel-approx}, we can write \cref{eq:measModel-mag-meas} as
\begin{equation}
\label{eq:measModel-mag-rr}
y_{\text{m},t} = R (q^\text{bn}_t) \nabla \Phi(p_t^\text{n}) \theta + \varepsilon_{\text{m},t},
\end{equation}
where $\Phi(p_t^\text{n}) \in \mathbb{R}^{m+3}$ consists of $m+3$ basis functions evaluated at $p_t^\text{n}$, and $\theta$ represents the weight of each of these basis functions. The first three basis functions model the linear kernel, while the other $m$ basis functions approximate the squared exponential kernel. The prior \cref{eq:prior} on the weights $\theta$ is in this case given by $\mu_\theta = 0_{m+3}$, with $0_{m+3}$ being an $m$-dimensional zero-vector, and $P_\theta$ given by 
\begin{equation}\label{eq:basis-Lambda}
    P_\theta = \diag( {\sigma_\text{lin.}^2, \sigma_\text{lin.}^2, \sigma_\text{lin.}^2, S_\text{SE}(\lambda_1), S_\text{SE}(\lambda_2), \ldots, S_\text{SE}(\lambda_m)} ).
\end{equation}
For more background concerning the model, we refer to \citet{Kok+Solin:2018}. Note the conditionally linear structure is in line with the more general measurement model \cref{eq:measModel}. 

\subsection{Dense radio SLAM}
\label{sec:radioSLAM}
In dense radio SLAM, a map of the RSSI values is constructed. Representing this map using a Gaussian process, the measurements can be modeled as~\citep{Ferris+Fox+Lawrence:2007}
\begin{subequations}
\begin{align}
h(p) &\sim \mathcal{GP}(0,\kappa(p,p')), \label{eq:measModel-radio-gp} \\
y_{\text{r},t} &= h(p) \big|_{p=p^\text{n}_t} + \varepsilon_{\text{r},t}, \label{eq:measModel-radio-meas}
\end{align}
\label{eq:measModel-radio}%
\end{subequations}
where $\varepsilon_{\text{r},t} \sim \mathcal{N}(0,\sigma_\text{r}^2)$. 
Note that $y_{\text{r},t} \in \mathbb{R}$ although in practice it is common to receive RSSI measurements from multiple Wi-Fi access points, resulting in multiple maps of the form of \cref{eq:measModel-radio}. 

Similarly to \cref{sec:magSLAM}, using \cref{eq:gp-MeasModel-approx} we can write \cref{eq:measModel-radio-meas} as
\begin{equation}
\label{eq:measModel-radio-rr}
y_{\text{r},t} = \Phi(p_t^\text{n}) \theta + \varepsilon_{\text{r},t},
\end{equation}
where $\Phi(p_t^\text{n}) \in \mathbb{R}^{m}$ consists of $m$ basis functions evaluated at $p_t^\text{n}$, and $\theta$ represents the weight of each of these basis functions. The prior \cref{eq:prior} on the weights $\theta$ is in this case given by $\mu_\theta = 0_{m}$, and $P_\theta$ given by \cref{eq:prior-gp-basis}. Note that \cref{eq:measModel-radio-rr} is again conditionally linear as in the more general measurement model \cref{eq:measModel}. 

\subsection{Sparse visual SLAM}
\label{sec:visualSLAM}
In sparse visual SLAM, the map is represented in terms of a number of landmark locations  which we denote by $p_{1:L}$. Using a pinhole camera model \citep{Hartley+Zisserman:2004}, a measurement $y_{j,t}, j=1,2,\ldots,L$ (tracked feature points in images) of landmark $p_j$ can be modeled as
\begin{equation}
  y_{j,t} = \frac{1}{\rho} \begin{pmatrix}
    l_{\text{u},t} \\ l_{\text{v},t} 
  \end{pmatrix} + \varepsilon_{\text{v},t},
  \label{eq:measModel-visual}
\end{equation}
where the points are based on the intrinsic and extrinsic camera matrices such that
\begin{equation}
  \begin{pmatrix}
    l_{\text{u},t} \\ l_{\text{v},t} \\ \rho
  \end{pmatrix} = 
  \begin{pmatrix}
    f_x & 0   & c_x \\
    0   & f_y & c_y \\
    0   & 0   & 1
  \end{pmatrix}
  \begin{pmatrix}
    R (q^{\text{bn}}_t) & -R (q^{\text{bn}}_t) p_t^\text{n}
  \end{pmatrix}
  \begin{pmatrix}
    p_j \\ 1
  \end{pmatrix}.
  \label{eq:measModel-visual-noProj}
\end{equation}
Here, $y_{j,t} \in \mathbb{R}^2$ denote the pixel coordinates of landmark $p_j$, $f_x$ and $f_y$ denote the focal lengths in the two different directions, $c_x$, $c_y$ denote the origin of the image plane, and $\varepsilon_{\text{v},t} \sim \mathcal{N}(0,\sigma_{v}^2 \mathcal{I}_2)$. Note that the model in \cref{eq:measModel-visual-noProj} is linear in $p_j$ conditioned on $p_t^\text{n}$ and $q^{\text{bn}}_t$. The projection in \cref{eq:measModel-visual} makes the model conditionally approximately linear. 

\section{Methods}
\label{sec:methods}
We derive a particle smoothing algorithm for SLAM to jointly estimate the time-varying pose $x_{1:T}$ of the sensor and the static but initially unknown map $\theta$ it is navigating in. We assume that the (possibly nonlinear, non-Gaussian) dynamics of the state is modeled according to \cref{eq:dynModel} and that the measurement model is nonlinear, but it is (approximately) linear when conditioned on the state, as modeled in \cref{eq:measModel}. Furthermore, we assume that the map is constant and that prior map information is available according to \cref{eq:prior}. The SLAM problem is inherently unobservable, since it is possible to shift or rotate the entire map and trajectory~\citep{gustafsson:2012}. To resolve this ambiguity, we assume that the initial pose is known, as is common practice in any SLAM formulation. 

The joint smoothing distribution $p( x_{1:T}, \theta \mid y_{1:T})$ for the SLAM problem cannot be computed in closed form due to the nonlinear nature of the models \cref{eq:dynModel,eq:measModel}. Using a similar strategy as in~\citet{Svensson+Schon+Kok:2015,wigrenRML:2019}, we instead approximate the joint smoothing distribution $p( x_{1:T}, \theta \mid y_{1:T})$ using Markov Chain Monte Carlo (MCMC). This approach has been shown to avoid the problem of particle degeneracy typically occurring in particle filters~\citep{Svensson+Schon+Kok:2015} and has been shown to be equivalent to backward simulation strategies commonly used for particle smoothing \citep{Lindsten+Jordan+Schon:2014}. Contrary to \citet{Svensson+Schon+Kok:2015,doucetFMR:2000}, we assume that the conditionally linear parameter vector (the map) is not time-varying. The case of static, conditionally linear parameters is also studied by \citet{wigrenRML:2019} and our algorithm can be considered to be a special case of their development. However, we will show that, contrary to \citet{wigrenRML:2019}, the specific structure of our model \cref{eq:dynModel,eq:measModel,eq:prior} for SLAM allows us to compute certain terms in closed form. 
 
In \cref{sec:smoother}, we first introduce our MCMC smoothing algorithm for SLAM. This algorithm makes use of two specific implementations of the  particle filter, which will subsequently be introduced in \cref{sec:rbpf-as,sec:crbpf-as}. For notational simplicity, throughout this section we assume that the measurement model \cref{eq:measModel} is conditionally linear and we omit the explicit dependence of the dynamic model \cref{eq:dynModel} on the inputs $u_{1:T}$. Note that the extension to a conditional approximately linear model is straightforward. 

\subsection{MCMC smoother for SLAM}
\label{sec:smoother}
Using a similar approach to~\citet{Svensson+Schon+Kok:2015,wigrenRML:2019}, we approximate the joint smoothing distribution $p( x_{1:T}, \theta \mid y_{1:T})$ by generating $K$ (correlated) samples using Markov Chain Monte Carlo (MCMC). In~\citet{Svensson+Schon+Kok:2015}, each iteration of the MCMC algorithm used a conditional particle filter with ancestor sampling. To also exploit the linear substructure in our problem, similar to \citet{wigrenRML:2019}, we use a conditional Rao-Blackwellized particle filter with ancestor sampling (CRBPF-AS) to generate samples $x_{1:T}[k], \theta[k]$, $k = 1,2, \hdots, K$. Our MCMC smoother is presented in \cref{alg:mcmc}. It starts with an initial state trajectory $x_{1:T}[0]$ and then draws $K$ samples from the Markov chain by running the CRBPF-AS for SLAM provided in \cref{alg:cpf-as}. This algorithm makes use of the previously sampled state trajectory as an input, and outputs a new sampled state trajectory and its corresponding map. A natural choice for the initial state trajectory is the trajectory from a Rao-Blackwellized particle filter with ancestor sampling (RBPF-AS). In \cref{sec:rbpf-as}, we will first describe our RBPF-AS for SLAM. We will then describe the extension to a CRBPF-AS in \cref{sec:crbpf-as}.

\begin{algorithm}[t!]
  \caption{MCMC smoother for SLAM}
  \label{alg:mcmc}
  \begin{minipage}{\linewidth-14.45pt}
  \begin{algorithmic}[1]
    \REQUIRE Initial state trajectory $x_{1:T}[0]$ from the RBPF-AS described in \cref{sec:rbpf-as}.
    \ENSURE $K$ samples from the Markov chain with state trajectories $x_{1:T}[1], \hdots, x_{1:T}[K]$ \\ and the corresponding maps $\theta[1], \hdots, \theta[K]$.
    \STATE \textbf{for} $k = 1,2, \hdots, K$ \textbf{do}
    \STATE \quad Run the CRBPF-AS from \cref{alg:cpf-as} conditional on $x_{1:T}[k-1]$ to obtain $x_{1:T}[k]$ and $\theta[k]$.
    \STATE \textbf{end for}
  \end{algorithmic}
  \end{minipage}
\end{algorithm}

\subsection{RBPF-AS for SLAM}
\label{sec:rbpf-as}
In this section we will first focus on deriving an RBPF-AS for SLAM that approximates the joint smoothing distribution $p( x_{1:T}, \theta \mid y_{1:T})$. We use a similar Rao-Blackwellization approach as \citet{wigrenRML:2019,Schon+Gustafsson+Nordlund:2005} and write the joint smoothing distribution at time $t$ as
\begin{equation}
p( x_{1:t}, \theta \mid y_{1:t}) = p(\theta \mid  x_{1:t}, y_{1:t}) p(x_{1:t} \mid y_{1:t}).
\label{eq:raoblackwelliseparams}
\end{equation}
Contrary to \citet{wigrenRML:2019}, the specific form of our model for SLAM \cref{eq:measModel,eq:prior} opens up for computing the distribution $p(\theta \mid  x_{1:t}, y_{1:t})$ in the closed form  
\begin{equation}
p(\theta \mid  x_{1:t}, y_{1:t}) = \mathcal{N}(\theta \, ; \, \hat{\theta}_t, P_t).
\label{eq:posteriorMap}
\end{equation}
More specifically, $\hat{\theta}_t$ and $P_t$ can be computed recursively as
\begin{align}
\hat{\theta}_t &= \hat{\theta}_{t-1} + K_t ( y_t - C(x_t) \hat{\theta}_{t-1} ), \qquad K_t = P_{t-1} ( C (x_t) )^\Transp (C (x_t) P_{t-1} C^\Transp (x_t) + \Sigma )^{-1} ,\nonumber \\
P_t &= P_{t-1} - K_t C(x_{t}) P_{t-1}, 
\label{eq:measUpdateParam}
\end{align}
with $\hat{\theta}_0$, $P_0$, respectively, equal to $\mu_\theta$, $P_\theta$ from \cref{eq:prior}. 

In line with \citet{wigrenRML:2019}, we compute the distribution $p(x_{1:t} \mid y_{1:t})$ from \cref{eq:raoblackwelliseparams} using SMC as~\citep[\cf,][]{Lindsten+Jordan+Schon:2014}
\begin{equation}
\hat{p}( x_{1:t} \mid y_{1:t}) = \sum_{i=1}^{N} w_t^i \delta_{x_{1:t}^i}(x_{1:t}).
\label{eq:smc}
\end{equation}
Note that the Rao-Blackwellization allows us to use the particles in the SMC approach \cref{eq:smc} to only represent the nonlinear states $x_t$. The Dirac delta in \cref{eq:smc} implies that for each particle, $x_t$ is deterministic and due to the conditionally linear structure of our model the map can be computed in closed form using \cref{eq:posteriorMap}. Hence, our approximation of the joint smoothing distribution $p( x_{1:t}, \theta \mid y_{1:t})$ makes use of $N$ particles to represent the state trajectory $x_{1:t}$, where each particle carries its own map that is updated at each time step using \cref{eq:measUpdateParam}. By writing $p(x_{1:t} \mid y_{1:t})$ as 
\begin{equation}
p(x_{1:t} \mid y_{1:t}) = \frac{ p(y_t \mid x_{1:t}, y_{1:t-1}) p(x_t \mid x_{1:t-1}, y_{1:t-1}) }{ p(y_t \mid y_{1:t-1}) } p(x_{1:t-1} \mid y_{1:t-1}) ,
\end{equation}
it can be seen that also the particles can be updated recursively. Using the Markov property of the state and the fact that our model \cref{eq:dynModel} assumes that the dynamics of the state is independent of the map $\theta$, $p(x_t \mid x_{1:t-1}, y_{1:t-1}) = p(x_t \mid x_{t-1})$ (note again that for notational simplicity we omitted the explicit conditioning on the input $u_{t-1}$). Furthermore, contrary to e.g.\ \citet{wigrenRML:2019,Svensson+Schon+Kok:2015}, the measurement model \cref{eq:measModel} for SLAM can be used to compute $p(y_t \mid x_{1:t}, y_{1:t-1})$ in closed-form as
\begin{align}
p(y_t \mid x_{1:t}, y_{1:t-1}) &= \int p(y_t \mid x_{1:t}, y_{1:t-1}, \theta) p(\theta \mid x_{1:t}, y_{1:t-1}) \dint \theta \nonumber \\
&= \int p(y_t \mid x_{t}, \theta) p(\theta \mid x_{1:t-1}, y_{1:t-1}) \dint \theta \nonumber \\
&= \mathcal{N}( y_t \, ; \, C(x_t) \hat{\theta}_{t-1}, C(x_t) P_{t-1} ( C(x_t) )^\Transp + \Sigma ).
\label{eq:weights}
\end{align}
Note that if the measurement model is only approximately linear, \cref{eq:weights} will be an approximation. Similar to the commonly used approach of bootstrap particle filtering~\citep{Doucet+deFreitas+Gordon:2001}, in our RBPF-AS we propagate the particles using the dynamic model \cref{eq:dynModel}, compute their weights using \cref{eq:weights} and resample the particles using systematic resampling. Specifically implementing this filter in terms of ancestor sampling~\citep{Lindsten+Jordan+Schon:2014} allows us to keep track of the history of each particle. The RBPF-AS for SLAM can be found in lines~\ref{as:initPose},~\ref{as:initParams},~\ref{as:initWeights},~\ref{as:resTime}--\ref{as:time},~\ref{as:shuffleDyn},~\ref{as:computeWeights} and \ref{as:updateParams} of \cref{alg:cpf-as}.

\begin{algorithm}[!t]
  \caption{Conditional Rao-Blackwellized particle filter with ancestor sampling for SLAM}
  \label{alg:cpf-as}
  \begin{minipage}{\linewidth-14.45pt}
  \begin{algorithmic}[1]
    \REQUIRE Reference state trajectory $x'_{1:T} = x_{1:T}[k-1]$, prior map mean and covariance $\mu_\theta$, $P_\theta$, and initial pose $x_0$.
    \ENSURE Sampled state trajectory $x_{1:T}[k]$ and its corresponding map $\theta[k]$. 
    \STATE Initialize $N-1$ particles at initial pose as $x_1^i = x_0$ for $i = 1, \hdots, N-1$.\label{as:initPose}
    \STATE Add reference state trajectory by setting $x_t^N = x'_t$ for $t = 1, \hdots, T$. \label{as:addref}
    \STATE Initialize parameters using the prior $p(\theta)$ as $\hat{\theta}^i_0 = \mu_\theta$ and $P^i_0 = P_\theta$ for $i = 1, \hdots, N$. \label{as:initParams}
    \STATE Set the initial weights equal to $w_1^i = 1/N$ for $i = 1, \hdots, N$. \label{as:initWeights}
    \STATE Optionally, precompute $C(x'_{t})$ for $t = 2, \hdots, T$.
    \FOR{$t = 1,2, \hdots, T$}
    \IF{$t > 1$}
    \STATE \textit{Resampling and time update:} \label{as:resTime}
    \STATE  Draw $a_t^i$ with $\mathbb{P}(a_t^i = j) = w_{t-1}^j$ for $i = 1, \hdots, N-1$. \label{as:res}
    \STATE  Draw $x_t^i$ by sampling $x_t^i \sim p(x_t^i \mid x_t^{a_t^i})$ for $i = 1, \hdots, N-1$. \label{as:time}
    \STATE \textit{Compute ancestor weights reference trajectory:} \\ Draw $a_t^N$ with $\mathbb{P}(a_t^N = i) \propto w_{t-1}^i p(y_{t:T} \mid x^i_{1:t-1}, x'_{t:T}, y_{1:t-1}) p(x'_{t} \mid x^i_{t-1})$, \\ where $p(y_{t:T} \mid x^i_{1:t-1}, x'_{t:T}, y_{1:t-1})$ is computed using~\cref{eq:likFutureMeasRef}. \label{as:ancWeights}
    \STATE Set $x_{1:t}^i = \big\{x_{1:t-1}^{a_t^i}, x_t^i \big\}$, $\hat{\theta}^i_{t-1} = \hat{\theta}_{t-1}^{a_t^i}$ and $P^i_{t-1} = P_{t-1}^{a_t^i}$ for $i = 1, \hdots, N$. \label{as:shuffleDyn}
    \ENDIF
    \STATE \textit{Compute weights:} \\ Set $w_t^i \propto p(y_t \mid x_{1:t}^i, y_{1:t-1})$ using~\cref{eq:weights} for $i = 1, \hdots, N$ and normalize to $\sum_{i = 1}^N w_1^i = 1$. \label{as:computeWeights}
    \STATE \textit{Update parameters:} \\ Compute $\hat{\theta}_t^i$ and $P_t^i$ using~\cref{eq:measUpdateParam}. \label{as:updateParams}
    \ENDFOR
    \STATE Sample a state trajectory and its corresponding map as $x_{1:T}[k] = x_{1:T}^J$ and $\theta[k] = \theta_T^J$ with $\mathbb{P}(J = j) = w_T^j$.
  \end{algorithmic}
  \end{minipage}
\end{algorithm}

\subsection{CRBPF-AS for SLAM}
\label{sec:crbpf-as}
During each iteration of the MCMC smoother from \cref{alg:mcmc}, we run a CRBPF-AS to obtain a sample of $p( x_{1:T}, \theta \mid y_{1:T})$ using the weights of the trajectories and maps at time $T$. 
This sampled trajectory will be used as a so-called reference trajectory in the next iteration of the MCMC smoother \citep{AndrieuDH:2010}. In practice this means that we run an RBPF-AS as described in \cref{sec:rbpf-as} with $N-1$ particles and assign the reference trajectory to particle $N$ for each time $t = 1, \hdots, T$~\citep{Svensson+Schon+Kok:2015,wigrenRML:2019}. At each time instance a history (ancestor index $a_t^N$) of this reference trajectory is sampled. In other words, we find a history for $x'_t$, where $x'_t$ denotes the reference trajectory at time instance $t$. The ancestor index is drawn with probability~\citep{Svensson+Schon+Lindsten:2014,wigrenRML:2019}
\begin{align}
\mathbb{P}(a_t^N = i) &\propto p( x_{1:t-1}^i \mid x'_{t:T}, y_{1:T}) \nonumber \\
&\propto p( y_{t:T}, x'_{t:T} \mid x^i_{1:t-1}, y_{1:t-1}) p( x^i_{1:t-1} \mid y_{1:t-1}),
\label{eq:ancestorProbRef}
\end{align}
where $p( x^i_{1:t-1} \mid y_{1:t-1})$ is equal to the importance weight $w^i_{t-1}$ of particle $i$, $i = 1, \hdots, N$. For our model, the term $p( y_{t:T}, x'_{t:T} \mid x^i_{1:t-1}, y_{1:t-1})$ can be rewritten according to  
\begin{align}
p(y_{t:T}, x'_{t:T} \mid x^i_{1:t-1}, y_{1:t-1}) 
&= p(y_{t:T} \mid x^i_{1:t-1}, x'_{t:T}, y_{1:t-1}) p(x'_{t:T} \mid x^i_{1:t-1}, y_{1:t-1}) \nonumber \\
&= p(y_{t:T} \mid x^i_{1:t-1}, x'_{t:T}, y_{1:t-1}) p(x'_{t} \mid x^i_{t-1}) \prod_{\tau = t+1}^{T} p(x'_{\tau} \mid x'_{\tau-1}) \nonumber \\
&\propto p(y_{t:T} \mid x^i_{1:t-1}, x'_{t:T}, y_{1:t-1}) p(x'_{t} \mid x^i_{t-1}).
\label{eq:ancestorProbRef-y}
\end{align}
Note that the Markovian structure of the dynamics in the second line of \cref{eq:ancestorProbRef-y} is due to the specific structure of our dynamic model \cref{eq:dynModel}. If the dynamic model would also depend on the map $\theta$, as for instance assumed by \citet{wigrenRML:2019}, the marginalized dynamic model would not be Markovian or, in other words, in that case $p(x'_{t:T} \mid x^i_{1:t-1}, y_{1:t-1})$ would not be equal to $p(x'_{t} \mid x^i_{t-1}) \prod_{\tau = t+1}^{T} p(x'_{\tau} \mid x'_{\tau-1})$. Note also that in line 3 of \cref{eq:ancestorProbRef-y} we have omitted the term $\prod_{\tau = t+1}^{T} p(x'_{\tau} \mid x'_{\tau-1})$ due to the fact that it is equal for all particles and hence irrelevant for computing \cref{eq:ancestorProbRef}. The interpretation of \cref{eq:ancestorProbRef,eq:ancestorProbRef-y} is that the probability of drawing the ancestor index~$i$ depends on {\em (i)}~the importance weight $w_{t-1}^i$, {\em (ii)}~the probability of transitioning from $x_{t-1}^i$ to the reference trajectory state $x_t'$ according to the dynamic model in \cref{eq:dynModel}, and {\em (iii)}~how well the map learned from $x_{1:t-1}^i, y_{1:t-1}$ can predict the current and future measurements $y_{t:T}$ at locations $x_{t:T}'$. The latter can be seen more clearly when writing $p(y_{t:T} \mid x^i_{1:t-1}, x'_{t:T}, y_{1:t-1})$ as
\begin{align}
p(y_{t:T} \mid x^i_{1:t-1}, x'_{t:T}, y_{1:t-1}) &= \int p(y_{t:T}  \mid x_{1:t-1}^i, x'_{t:T}, \theta) p(\theta  \mid x^i_{1:t-1}, x'_{t:T}, y_{1:t-1}) \dint \theta \nonumber \\
&= \int p(y_{t:T}  \mid x'_{t:T}, \theta) p(\theta  \mid x^i_{1:t-1}, y_{1:t-1}) \dint \theta,
\end{align}
where the explicit dependence on $\theta$ is introduced to write this probability distribution in terms of the measurement model $p(y_{t:T}  \mid x'_{t:T}, \theta)$ and the map estimate $p(\theta  \mid x^i_{1:t-1}, y_{1:t-1})$. Note that the conditioning of the first term inside the integral on $x_{1:t-1}^i,y_{1:t-1}$ is dropped since given $\theta$ all history is contained in $x_t$ and that the conditioning of the second term inside the integral on $x^i_{1:t-1}$ is dropped since locations without measurements do not affect the map estimate. For our state-space model~\cref{eq:dynModel,eq:prior,eq:measModel} it is possible to compute $p(y_{t:T} \mid x^i_{1:t-1}, x'_{t:T}, y_{1:t-1})$ in closed form as 
\begin{equation}
p(y_{t:T} \mid x^i_{1:t-1}, x'_{t:T}, y_{1:t-1}) 
= \mathcal{N}( \bar{y}_{t} \, ; \, \bar{C}(x'_{t:T}) \hat{\theta}^i_{t-1}, \bar{C}(x'_{t:T}) P^i_{t-1} (\bar{C}(x'_{t:T}))^\Transp + \bar{\Sigma} ),
\label{eq:likFutureMeasRef}
\end{equation}
with 
\begin{equation}
\bar{y}_{t} = \begin{pmatrix} y_{t}^\Transp & y_{t+1}^\Transp & \hdots & y_T^\Transp \end{pmatrix}^\Transp, \quad 
\bar{C}(x'_{t:T}) = \begin{pmatrix} (C(x'_{t}))^\Transp & (C(x'_{t+1}))^\Transp & \hdots & (C(x'_{T}))^\Transp \end{pmatrix}^\Transp, \quad \bar{\Sigma} = \mathcal{I}_{T-t} \otimes \Sigma,
\label{eq:yCRbar}
\end{equation}
where $\otimes$ denotes a Kronecker product. Note that if the measurement model is only approximately linear, \cref{eq:likFutureMeasRef} will be an approximation. Using \cref{eq:likFutureMeasRef} in \cref{eq:ancestorProbRef-y}, we can now sample ancestor indices for the reference trajectory at time $t = 1,2, \hdots, T$. Combining these steps with the RBPF-AS described in \cref{sec:rbpf-as}, the resulting conditional RBPF-AS for SLAM can be found in \cref{alg:cpf-as}.

\section{Results}
\label{sec:experiments}
To validate and study the properties of our approach, we present simulation and experimental results. We first study the properties of our smoothing algorithm in terms of particle degeneracy and by analyzing the shape of the estimated trajectories for a simulated radio SLAM example in \cref{sec:radioSLAM-results}. Subsequently, we compare our estimation results for simulated magnetic SLAM with results from a particle filter and from an extended Kalman filter in \cref{sec:magSLAM-results}. In a third simulation experiment, we compare our estimation results for visual SLAM with results from a particle filter, an extended Kalman filter and an extended Kalman smoother in \cref{sec:visualSLAM-results}. Finally, in \cref{sec:expResults} we illustrate the efficacy of our algorithm to estimate a trajectory using magnetic SLAM based on real-world experimental data collected using a smartphone. 

\begin{figure}[!t]
\tikzexternaldisable
\setlength{\figurewidth}{.28\textwidth}
\setlength{\figureheight}{\figurewidth}
\def\datapath{./fig/}
\centering\footnotesize
\begin{subfigure}[t]{\figurewidth}  
  \raggedright
  \tikzsetnextfilename{fig-6a}
  \fboxsep=0pt
  \fbox{%
  \begin{tikzpicture}[outer sep=0,inner sep=0]

    \node[outer sep=0,inner sep=0] at (0,0) {\includegraphics[width=\figurewidth]{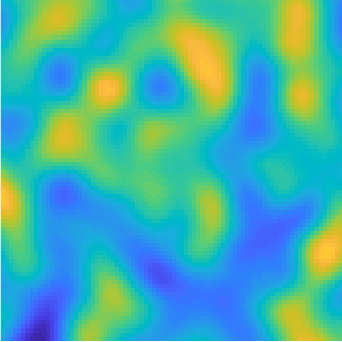}};

    \newlength{\pathradius}
    \setlength{\pathradius}{.2941\figurewidth}

    \draw[line width=2pt,draw=primarycolor,-latex] (-\pathradius,-\pathradius) -- (-\pathradius,\pathradius) -- (\pathradius,\pathradius) -- (\pathradius,-\pathradius) -- (-.9\pathradius,-\pathradius);

    \node at (0,0) {(RSSI field)};

    \draw[line width=2pt,primarycolor] (-\pathradius,.8\pathradius) arc (-90:0:.2\pathradius);
    \node at (-.73\pathradius,.73\pathradius) {?};
    
    \draw[line width=2pt,primarycolor] (\pathradius,.8\pathradius) arc (-90:-180:.2\pathradius);
    \node at (.73\pathradius,.73\pathradius) {?};

    \draw[line width=2pt,primarycolor] (\pathradius,-.8\pathradius) arc (90:180:.2\pathradius);   
    \node at (.73\pathradius,-.73\pathradius) {?};
    
  \end{tikzpicture}}
  \caption{Simulation setup}
  \label{fig:degeneracy-path}    
\end{subfigure}
\hfill
\begin{subfigure}[t]{\figurewidth}
  \centering
  \tikzsetnextfilename{fig-6b}
  \pgfplotsset{xticklabel pos=top}
  \input{./fig/degeneracy-filter.tex}
  \caption{Filter degeneracy}
  \label{fig:degeneracy-filter} 
\end{subfigure}
\hfill
\begin{subfigure}[t]{.38\textwidth}
  \raggedleft
  \tikzsetnextfilename{fig-6c}
  \pgfplotsset{xticklabel pos=top,colorbar style={
    width=1em,
    ytick style={draw=none},
    ytick={-4,-2,0,2,4},
    yticklabels={{$-4$},{$-2$},{$\phantom{-}0$},{$\phantom{-}2$},{$\phantom{-}4$}}}}
  \input{./fig/degeneracy-smoother.tex}
  \caption{Smoother non-degeneracy}
  \label{fig:degeneracy-smoother} 
\end{subfigure}
\caption{Highlight of the non-degeneracy of the particle smoothing approach with (a) the simulation setup, and simulation results for radio SLAM with (b) filtering samples showing that the filter is degenerate and (c) non-degenerate samples of the smoother} %
\label{fig:degeneracy}
\end{figure}

\subsection{Simulation results for radio SLAM}
\label{sec:radioSLAM-results}
To study the properties of \cref{alg:mcmc}, we first simulate RSSI measurements and consider planar localization to estimate a time-varying position $p_t \in \mathbb{R}^2$ and a heading angle $q_t^\text{nb} \in \mathbb{R}$. We model the measurements according to the model \cref{eq:measModel-radio} and the GP approximation \cref{eq:measModel-radio-rr}, with a measurement noise covariance of $\sigma_\text{r}^2 = 0.01$. We define the GP prior through a squared exponential covariance function given by~\cref{eq:gp-se} that encodes a smoothness assumption on the model function. We use $128$ basis functions, and fix the model hyperparameters to $\sigma_\text{f}^2 = 2$ (magnitude) and $\ell = 0.25$ (lengthscale). Furthermore, we use a square domain to ensure that the eigenfunctions for the GP approximation, see \cref{sec:background-gp}, can be computed in closed-form \citep{Solin+Sarkka:2020}. To avoid boundary effects, we ensure that all ground-truth positions are at least two lengthscales from the boundary of the domain. 

\begin{figure}
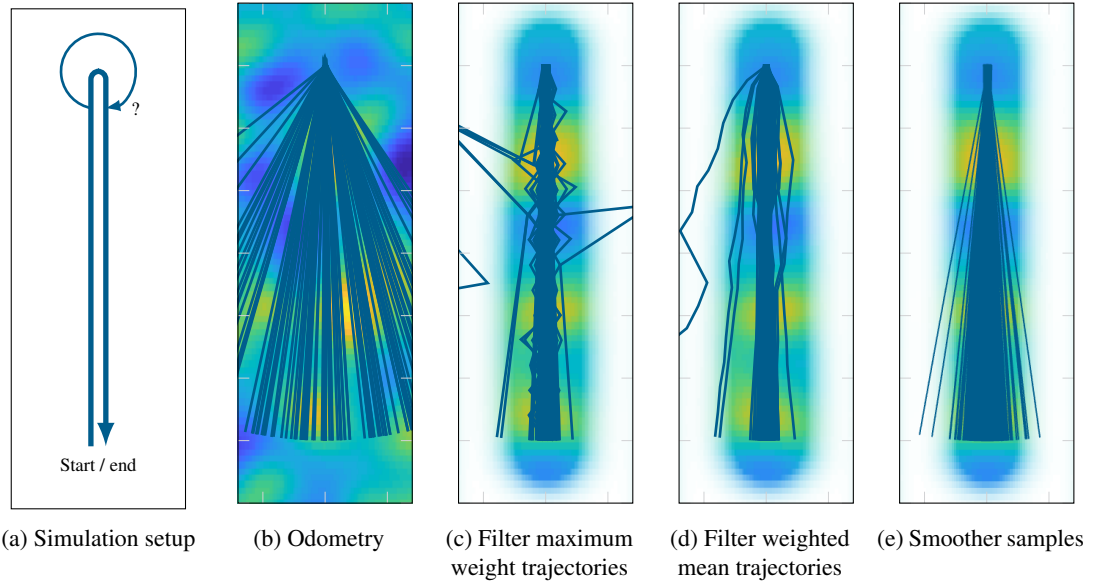

\tikzexternaldisable
\setlength{\figurewidth}{.16\textwidth}
\setlength{\figureheight}{2.8571\figurewidth}
\def\datapath{./fig/}
\captionsetup{justification=centering}
\centering\footnotesize
\pgfplotsset{xticklabel pos=top}
\begin{subfigure}[t]{.19\textwidth}  
  \centering
  \tikzsetnextfilename{fig-5a}
  \begin{tikzpicture}[outer sep=0,inner sep=0]

    \node[draw,minimum width=\figurewidth,minimum height=\figureheight,inner sep=0,outer sep=0] at (0,0) {};

    \newlength{\pathlength}
    \setlength{\pathlength}{.75\figureheight}

    \draw[line width=2pt,draw=primarycolor,-latex,rounded corners=3pt] (-.02\pathlength,-.5\pathlength) -- (-.02\pathlength,.5\pathlength) -- (.02\pathlength,.5\pathlength) -- (.02\pathlength,-.5\pathlength);

    \def\centerarc[#1](#2)(#3:#4:#5)%
    { \draw[#1] ($(#2)+({#5*cos(#3)},{#5*sin(#3)})$) arc (#3:#4:#5); }

    \centerarc[line width=1pt,draw=primarycolor,latex-](0,.5\pathlength)(-80:260:.1\pathlength)

    \node at (0,-.55\pathlength) {Start / end};
    \node at (.1\pathlength,.4\pathlength) {?};
        
  \end{tikzpicture}
  \caption{Simulation setup}
  \label{fig:line-setup}    
\end{subfigure}
\hfill
\begin{subfigure}[t]{.19\textwidth}
  \centering
  \tikzsetnextfilename{fig-5b}
  \input{./fig/line-odometry.tex}
  \caption{Odometry}
  \label{fig:line-odometry} 
\end{subfigure}
\hfill
\begin{subfigure}[t]{.19\textwidth}
  \centering
  \tikzsetnextfilename{fig-5c}
  \input{./fig/line-filter-max.tex}
  \caption{Filter maximum \\ weight trajectories}
  \label{fig:line-filter-max} 
\end{subfigure}
\hfill
\begin{subfigure}[t]{.19\textwidth}
  \centering
  \tikzsetnextfilename{fig-5d}
  \input{./fig/line-filter-mean.tex}
  \caption{Filter weighted \\ mean trajectories}
  \label{fig:line-filter-mean} 
\end{subfigure}
\hfill
\begin{subfigure}[t]{.19\textwidth}
  \centering
  \tikzsetnextfilename{fig-5e}
  \input{./fig/line-smoother.tex}
  \caption{Smoother samples}
  \label{fig:line-smoother} 
\end{subfigure}
\caption{(a)~Illustrative example of a back and forth path on an RSSI map; (b)~the only source of uncertainty in the odometry is the turn angle at the farthest most point, which leads to spread; (c--d)~Particle filtering SLAM reduces the spread, but does not backtrack the smooth odometry information; (e)~Our particle smoothing solution gives a tighter estimate and backtracks the smoothness along the sample paths}%
\label{fig:line}
\end{figure}

We simulate odometry measurements $\Delta p_t, \Delta q_t$ that provide information about the change in position and heading at each time instance $t$, respectively, and model the discrete-time dynamics according to
\begin{subequations}\label{eq:exp-dyn}
\begin{align}
p_{t+1}^\text{n} &= p_t^\text{n} + R(q_t^\text{nb}) \Delta p_t, \\
q_{t+1}^\text{nb} &= q_t^\text{nb} + \Delta q_t + w_{t}, \qquad w_{t} \sim \mathcal{N}(0, \Delta T Q_t).
\end{align}
\end{subequations}
Here, $\Delta T = 1$ and $R(q_t^\text{nb})$ denotes the $2 \times 2$ rotation matrix representing a rotation around the $z$-axis with an angle~$q_t^\text{nb}$. This model reflects the fact that odometry measurements are typically measured in a body-fixed frame while navigation is typically done in an earth-fixed frame. For illustrational purposes, we assume that the position odometry measurements $\Delta p_t$ are noiseless, but model a noise $w_t$ representing the error in the heading odometry measurements. More specifically, we simulate trajectories with $Q_t = 1 \cdot 10^{-6}$ rad$^2$ for all time instances where the user is moving straight. For the samples where a turn occurs, we model $Q_t$ to be significantly larger ($Q_t = 0.1^2$ and $Q_t = 0.3^2$ for the examples in \cref{fig:degeneracy,fig:line}, respectively). Such a model represents for instance a visual odometry where straight lines can be tracked very accurately but the odometry accuracy can have large uncertainties during fast turns. 

For illustrative sanity-checks, we consider two different scenarios. First, we focus on a square trajectory where $Q_t = 0.1^2$ rad$^2$ in the three $90^\circ$ turns. This scenario is visualized in \cref{fig:degeneracy-path} where both the true trajectory and the true RSSI field map are displayed. We then run the Rao-Blackwellized particle filter with ancestor sampling from \cref{sec:rbpf-as}. \cref{fig:degeneracy-filter} visualizes the history of the $100$ particles at time $T$ as well as the RSSI map estimated by the particle with the highest weight at this time instance. The transparency of the map indicates its estimation uncertainty. Finally, we generate $K = 50$ samples $x_{1:T}[1], \hdots, x_{1:T}[K], \theta[1], \hdots, \theta[K]$ using \cref{alg:mcmc} and visualize these $50$ trajectories as well as the RSSI map estimated during iteration $K$ in \cref{fig:degeneracy-smoother}. Again, the transparency of the map indicates the predictive marginal standard deviation (estimation uncertainty). In \cref{fig:degeneracy-filter} it can be seen that the filter suffers from particle degeneracy as all particles at time $T$ descend only from a small number of ancestors in the past. Please note that even though all particles seem to have a common ancestor at time $t = 1$, this is not due to particle degeneracy but due to how we initialize our SLAM problem: To overcome inherent unobservability in the problem, we use a standard approach in which we assume that the initial pose is known \citep{gustafsson:2012}. Nevertheless, there is clear particle degeneracy as  all particles at time $T$ descend from only 5 ancestors in the upper left corner and only 7 ancestors in the upper right corner. The posterior as visualized by MCMC samples in \cref{fig:degeneracy-smoother} can be seen to have a much wider spread representing possible trajectories. Note that the large spread is due to the large odometry uncertainty in each corner, and that this uncertainty decreases only when revisiting previous locations. This reduction of uncertainty will be illustrated in the next scenario.  

In a second scenario, we assume that a user moves straight, turns and returns to the starting point as visualized in \cref{fig:line-setup}. We run 100 Monte Carlo simulations where for each simulation we sample both the measurements of the field as well as the odometry. The 100 odometry trajectories as well as the true RSSI map are visualized in \cref{fig:line-odometry}. We run the filter from \cref{sec:rbpf-as} with 100 particles and for each of the Monte Carlo simulations we run 50 iterations of the smoother from \cref{alg:mcmc}. At each time instance of the filter, we save both the highest weight particle and the weighted mean particle. The resulting trajectories from the filter are shown in \cref{fig:line-filter-max,fig:line-filter-mean}. In \cref{fig:line-filter-max,fig:line-filter-mean,fig:line-smoother} we also visualize the estimated field for one of the Monte Carlo samples. The transparency again indicates the estimation uncertainty. For the smoother we visualize the field estimated by the $K$\textsuperscript{th} sample. As can be seen, the sampled trajectories from \cref{alg:mcmc} are significantly more smooth than the estimates of the filter from \cref{sec:rbpf-as}. Furthermore, the estimates from the filter include some trajectories which significantly deviate from the true path while this is not the case for the samples from the particle smoother. 

\subsection{Simulation results for magnetic SLAM}
\label{sec:magSLAM-results}
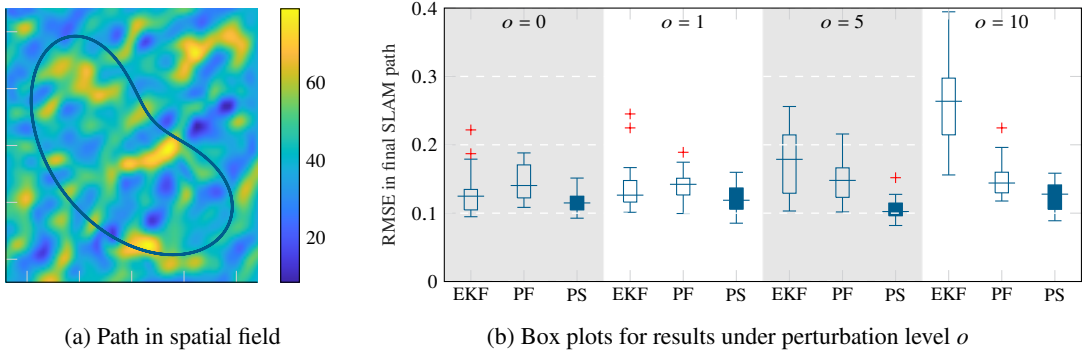
\begin{figure}[!t]
\def\datapath{./fig/}
\tikzexternaldisable
\setlength{\figureheight}{.25\textwidth}
\begin{subfigure}[b]{.33\textwidth}
  \raggedright\footnotesize
  \setlength{\figurewidth}{.7\textwidth}
  \pgfplotsset{axis on top,set layers,xticklabel pos=top,colorbar style={
    width=1em,ytick style={draw=none}},xticklabels={\phantom{E}}}
  \input{./fig/mag-path-field.tex}
  \caption{Path in spatial field}
  \label{fig:mag-setup} 
\end{subfigure}
\hfill
\begin{subfigure}[b]{.65\textwidth}
  \raggedleft\footnotesize
  \setlength{\figurewidth}{.9\textwidth}
  \pgfplotsset{width=\figurewidth,height=\figureheight,ytick align=inside,xtick align=inside,axis on top, scale only axis,
  legend style={font=\tiny},
  grid style={line width=.5pt,draw=white,dashed},
  ymajorgrids,set layers,ytick distance=1}
  \input{./fig/boxplot-mag.tex}%
  \caption{Box plots for results under perturbation level $o$}
  \label{fig:mag-rmse} 
\end{subfigure}\\[-2em]
\caption{Robustness study with magnetometer perturbed with a constant bias $o$. Left:~Setup showing the ground truth path and one realization of the magnetic field map. Right:~Box plots of RMSE in the final estimated SLAM path. Particle and extended Kalman filtering methods drawn with outlines, particle smoothing method in solid colours. The EKF performs well under negligible calibration errors, the particle filter (PF) and smoother (PS) perform well under large calibration errors}%
\label{fig:magSLAM-results}
\end{figure}

We also study the properties of our algorithm for magnetic field SLAM. For this, we simulate a magnetic field using 512 basis functions and $\sigma^2_\text{lin.} = 650$,  $\sigma_\text{m}^2 = 10$, $\ell = 1.3$ and $\sigma_\text{f}^2 = 200$. These hyperparameters are equal to the ones that we used previously in \citet{Kok+Solin:2018} and in line with values we found by estimating the hyperparameters from experimental data \citep{solinKWSS:2018}. We again use a square domain for the GP approximation and to avoid boundary effects, we ensure that the simulated trajectory, as shown in \cref{fig:mag-setup} is at least two lengthscales from the boundary of the domain. We simulate the odometry using the following dynamic model
\begin{subequations}
\begin{align}
p_{t+1} &= p_t + \Delta p_t + e_{\text{p},t}, \\
q^\text{nb}_{t+1} &= q^\text{nb}_{t+1} \odot \Delta q_t \odot \expq \big( e_{\text{q},t} \big),
\end{align}%
\label{eq:dynModel-mag}%
\end{subequations}
where $e_{\text{p},t} \sim \gaussian{0}{ \Delta T Q_\text{p}} $ and $e_{\text{q},t} \sim \gaussian{0}{ \Delta T Q_\text{q}}$, with $\Delta T = 0.01$~s, $Q_\text{p} = \text{diag}( 0.25, 0.25, 0.01 )$ and $Q_\text{q} = \text{diag}(0.01^2 \pi^2/180^2, 0.01^2 \pi^2/180^2, 0.3^2 \pi^2/180^2)$. We compare the results of our particle smoother with a particle filter similar to the one presented in~\citet{Kok+Solin:2018} and an extended Kalman filter similar to the one presented in~\citet{visetHK:2022}. To study the three algorithms for non-ideal, real-world scenarios, we add a constant, unmodeled magnetometer bias of varying magnitude~$o$ to the y-axis of the measurements. Because we are not interested in the true magnetic field map but only in one that aids in obtaining a correct position, an incorrect bias estimate would not have any negative effect on the localization performance if the sensor would not rotate. However, rotating the sensor results in measuring the bias in a different direction in the navigation frame and hence inconsistencies in the map. The sensitivity of the algorithms to erroneous magnetometer calibration therefore depends on the shape of the trajectory. We use the trajectory shown in \cref{fig:mag-setup}, which is not that sensitive for erroneous calibration, \eg because the same path is not traversed twice in opposite directions. Because of this, we simulate large calibration errors of $o = 0, 1, 5, 10$. For each of these cases, we run 20 Monte Carlo simulations and sample $K = 10$ sample trajectories from our Rao-Blackwellized particle smoother. One of the maps is visualized in \cref{fig:mag-setup}. The resulting RMSE can be found in \cref{fig:mag-rmse}. As can be seen, the RMSE of the particle smoother from \cref{alg:cpf-as} is smaller than that of the particle filter and remains consistently low, while that of the extended Kalman filter increases for larger calibration errors. 

\subsection{Simulation results for visual SLAM}
\label{sec:visualSLAM-results}
In a final set of simulations, we consider a visual SLAM problem. The previous experiments considered the case where our model was conditionally linear, whereas the sparse visual model is conditionally {\em linearized} (see \cref{sec:visualSLAM}). For simulation purposes, we consider the two-dimensional setup from \cref{fig:intro}. In two spatial dimensions, the pinhole camera observations become one-dimensional and the observation model becomes
\begin{equation}
  y_{j,t} = \frac{l_{t}}{\rho} + \varepsilon_{\text{v},t},
  \quad \text{with} \quad
  \begin{pmatrix}
    l_{t} \\ \rho
  \end{pmatrix} = 
  \begin{pmatrix}
    f & c \\
    0 & 1
  \end{pmatrix}
  \begin{pmatrix}
    R (q^{\text{bn}}_t) & -R (q^{\text{bn}}_t) p_t^\text{n}
  \end{pmatrix}
  \begin{pmatrix}
    p_j \\ 1
  \end{pmatrix},
\end{equation}
where $y_j \in \mathbb{R}$ denotes the observed pixel coordinate of landmark $p_j \in \mathbb{R}^2$, $f$ denotes the focal length, $c$ the origin of the image line, and $\varepsilon_{\text{v},t} \sim \mathcal{N}(0,\sigma_{v}^2)$. The task is to learn both the sparse map of $\theta = \{p_j\}_{j=1}^L$ and the 3-DoF trajectory of the camera. In the experiment, we use $f = 1.5$, $c = 0$ (the field-of-view is ${\sim}$67$^\circ$), and $\sigma_{v}^2 = 0.1^2$. The dynamical model for the camera movement follows the setup in \cref{eq:dynModel-mag} with $\Delta T = 1$ and a total of $T=197$ time steps. To corrupt the odometry signal, we add a constant drift of 1~cm/s and Gaussian noise with covariance $Q_\text{p} = 0.04^2\mathcal{I}_2$/s to the spatial increments $\Delta p_t$, and add a small amount of noise ($Q_\text{q} = 10^{-12}$/s) to the orientation increments. This setup corresponds to a typical visual-inertial odometry setup, where tracking orientation with the help of a gyroscope is easier than tracking the twice-integrated accelerometer signal. In the simulation setup, the camera traverses around the space two times, which means that learned landmark points are revisited and should improve the tracking. 

\begin{figure}[t!]
  \centering\footnotesize
  \setlength{\figurewidth}{.9\textwidth}
  \setlength{\figureheight}{.4\figurewidth}
  \pgfplotsset{width=\figurewidth,height=\figureheight,ytick align=inside,xtick align=inside,axis on top, scale only axis,
  legend style={font=\tiny},
  grid style={line width=.5pt,draw=white,dashed},
  ymajorgrids,set layers}
  \input{./fig/boxplot.tex}\\[-1em]
  \caption{Robustness to feature point initialization in visual SLAM. Simulation study with initial feature locations perturbed with Gaussian noise (variance $\sigma^2$). Box plots of RMSE in the final estimated SLAM path. Forward filtering methods drawn with outlines, smoothing methods in solid colours. The EKF/EKS perform well under negligible perturbation, the particle filter (PF) and smoother (PS) perform well under large uncertainty}
  \label{fig:boxplot}
\end{figure}
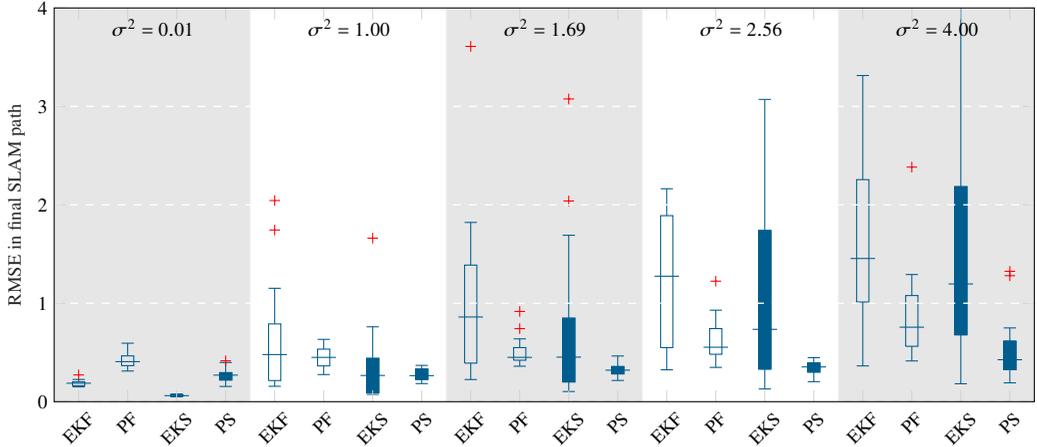

A recurring issue in visual SLAM problems is the initialization of the feature point locations $p_j$. The initial guess is typically vague and could be, \eg, based on triangulation from only two views, where errors in the pose of these views translate to large errors in the initialization. We study robustness to feature point initialization by controlling the initial error of the initialization in a simulated setting. We follow a similar setup as \citet[Sec.~IV.A]{solin2022alook}, where we initialize the points by taking their ground truth locations and corrupting them with Gaussian noise (with variance $\sigma^2$). The visual SLAM setting follows that visualized in \cref{fig:intro-a}, where we have $L=20$ feature points---most of which are not observed at the same time.

As explained in \cref{sec:visualSLAM}, the sparse visual SLAM model is a conditionally linearized filter/smoother. Thus, it falls natural to compare Rao-Blackwellized particle filtering/smoothing (denoted PF/PS) to a vanilla extended Kalman filter/smoother as baselines (denoted EKF/EKS). For this two-dimensional simulated SLAM problem, we use $N=100$ particles for the Rao-Blackwellized particle filter with ancestor sampling and sample $K=10$ sample trajectories from our Rao-Blackwellized particle smoother. We repeat the experiment with 20 random repetitions for each noise level $\sigma^2$. The prior state covariance corresponding to the feature location is initialized to $4^2\mathcal{I}_2$, which means that the scale of the corrupting noise can be considered moderate even for large noise scales.

\cref{fig:boxplot} shows results for controlling the feature point initialization. For each noise level, we show box plots for the final error of the weighted mean estimates compared to the ground truth translation of the moving camera. We apply standard Procrustes correction (rotation, translation, and scalar scaling) based on the learned map points due to possible lack of scale information in the visual-only observations. The results in the box plots are as expected: The effect of deviating from the vicinity of the `true' linearization point is apparent, and the EKF and EKS performance deteriorates quickly as the noise scale grows. The particle filter appears robust to the initialization when compared to the EKF, and the particle smoother shows clearly improved robustness over the EKS.

\begin{figure}
\tikzexternaldisable
\setlength{\figurewidth}{.18\textwidth}
\setlength{\figureheight}{1.8041\figurewidth}
\def\datapath{./fig/}
\captionsetup{justification=centering}
\centering\footnotesize
\pgfplotsset{xticklabel pos=top}
\begin{subfigure}[b]{.18\textwidth}  
  \centering
  \tikz\node[minimum width=0.5625\figureheight,minimum height=\figureheight,inner sep=0,rounded corners=.5em,clip] {\includegraphics[height=\figureheight]{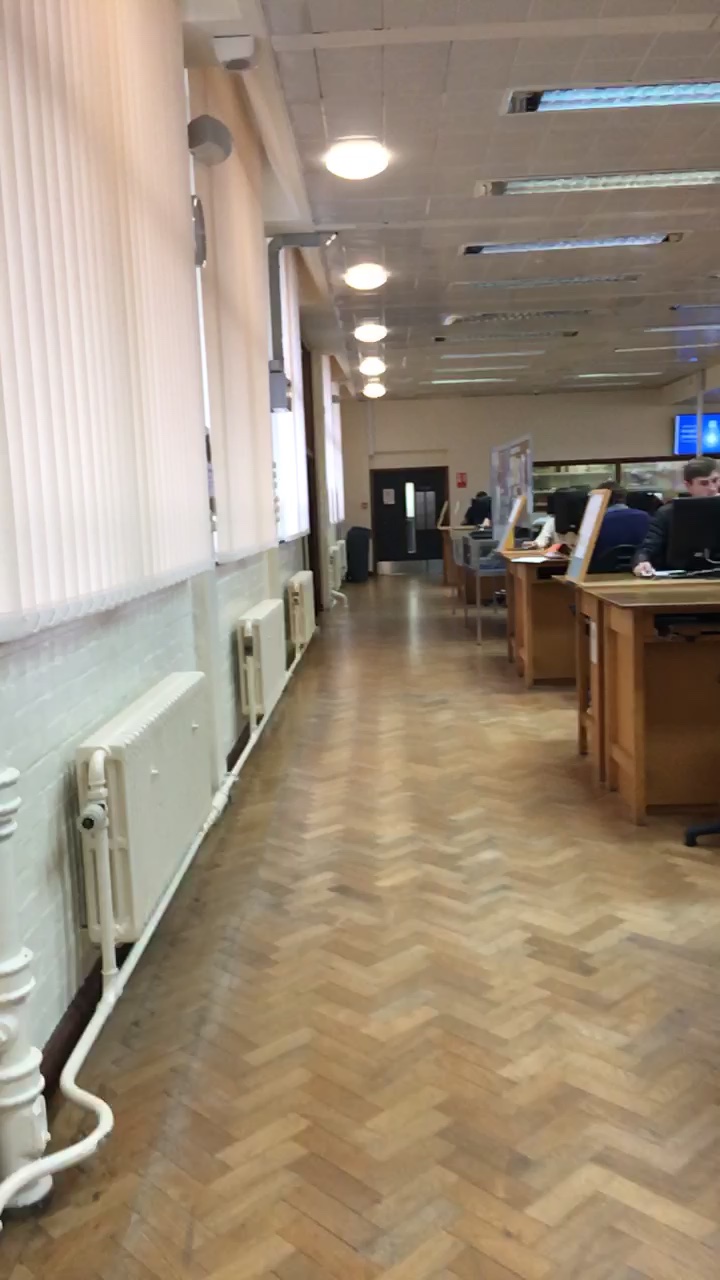}};
  \caption{Magnetic environment}
  \label{fig:expResults-scene}
\end{subfigure}
\hfill
\begin{subfigure}[b]{.18\textwidth}  
  \centering
  \tikzsetnextfilename{fig-6a}
  \input{./fig/dpo-odometry.tex}
  \caption{Drifting odometry}
  \label{fig:expResults-odo}
\end{subfigure}
\hfill
\begin{subfigure}[b]{.18\textwidth}
  \centering
  \tikzsetnextfilename{fig-6b}
  \input{./fig/dpo-smoother.tex}
  \caption{Smoother samples}
  \label{fig:expResults-ps} 
\end{subfigure}
\hfill
\tikzexternalenable
\begin{subfigure}[b]{.38\textwidth}
  \centering
  \begin{tikzpicture}
    \node[inner sep=0] (cbar) {\includegraphics[width=22mm,height=2mm]{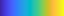}};
  \node[anchor=north,inner sep=0,outer sep=6pt] at (cbar.south west) {\tiny ~~$30\,\mu\text{T}$};
  \node[anchor=north,inner sep=0,outer sep=6pt] at (cbar.south east) {\tiny ~~$60\,\mu\text{T}$};
  \node[below of=cbar,rotate=0,yshift=1em] {\tiny Magnetic field strength};
  \end{tikzpicture}
  
  \includegraphics[width=.9\textwidth,trim=110 110 90 100,clip]{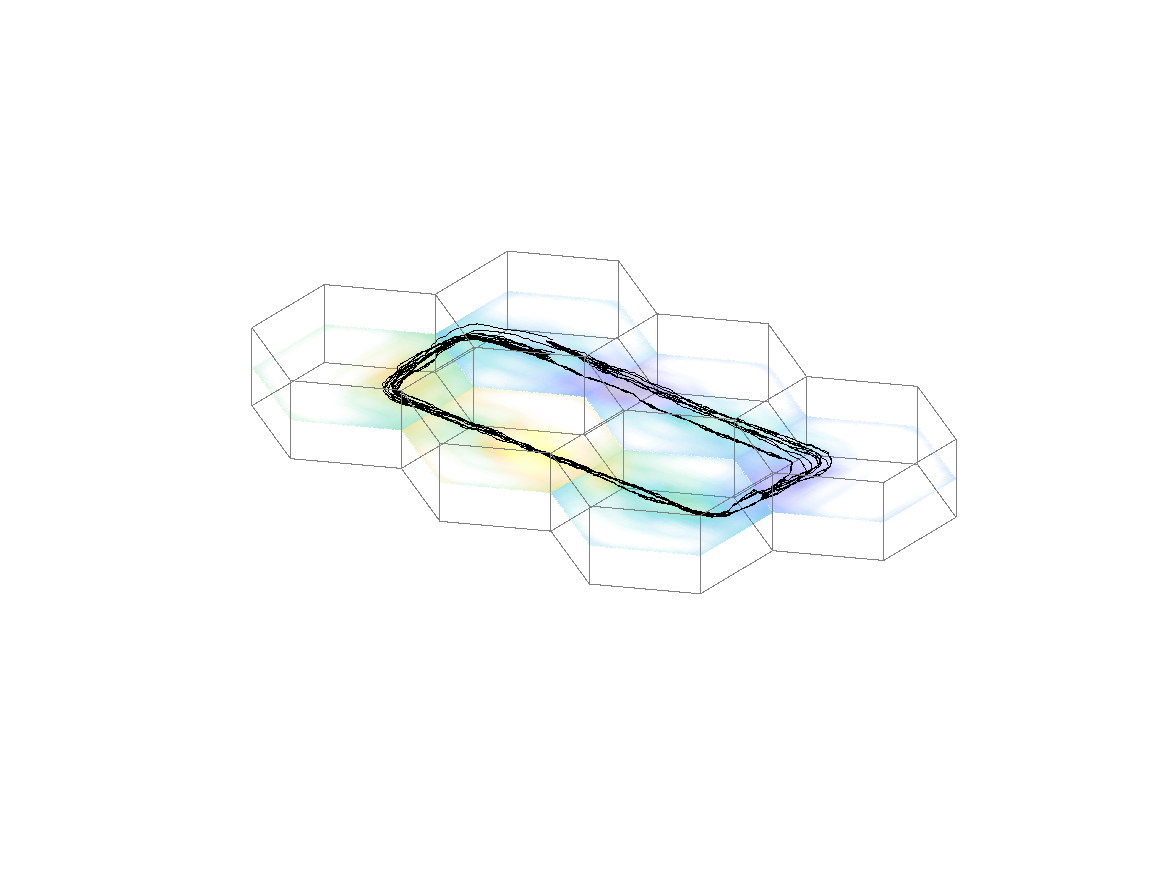}
  \caption{3D view \\ ~}
  \label{fig:dpo-3D} 
\end{subfigure}\\[-1em]
\caption{Empirical proof-of-concept magnetic indoor SLAM. Panel (a) shows a view through the mobile phone camera, (b) the drifting odometry, (c) samples of the smoothing distribution, and (d) a 3D view of these results. The color scaling visualizes the field strength of a learned magnetic map}%
\label{fig:experimentalResults}
\end{figure}
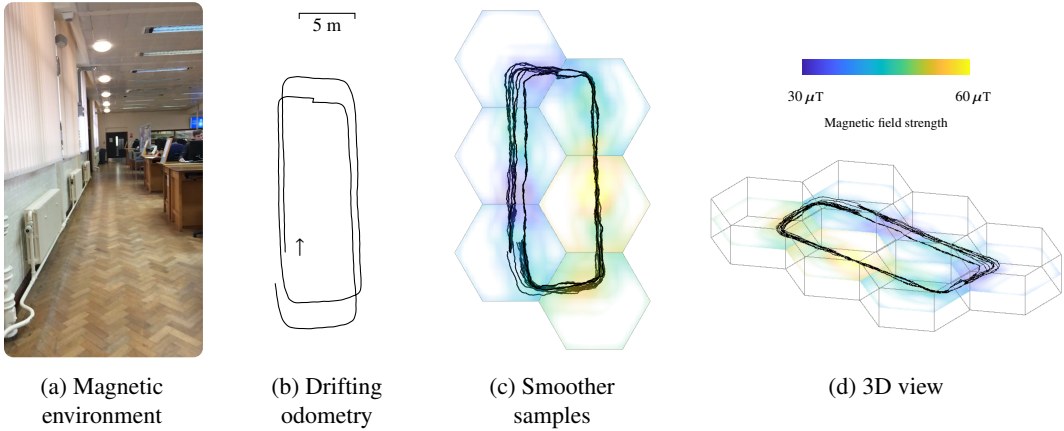

\subsection{Empirical results for magnetic field SLAM}
\label{sec:expResults}
To experimentally validate our algorithm, we use experimental data previously used for magnetic field-based SLAM in~\cite{Kok+Solin:2018}. In this experiment, we collected magnetometer and odometry data using an Apple
iPhone 6s. To obtain the odometry data we used the app ARKit which uses IMU and camera data to provide a 6-DoF (position and orientation) movement trajectory. ARKit visually recognizes previously visited locations and corrects the trajectory for this, resulting in discontinuities in the trajectories. We instead removed these discontinuities, resulting in a drifting odometry shown in \cref{fig:expResults-odo}. The goal of our SLAM algorithm is then to remove the drift in this trajectory. To this end, we use the dynamic model \cref{eq:dynModel-mag} and the measurement model \cref{eq:measModel-mag} with the reduced-rank approximation from \cref{eq:measModel-mag-rr} and downsample the data from the original 100 Hz to 10 Hz. We use the same hyperparameters as in~\citet{Kok+Solin:2018}: $\sigma^2_\text{lin.} = 650$,  $\sigma_\text{m}^2 = 10$, $\ell = 1.3$~m and $\sigma_\text{f}^2 = 200$. Furthermore, we have $\Delta T \approx 0.1$, $\Sigma_\text{p} = \text{diag}(0.1^2, 0.1^2, 0.02^2)$, $\Sigma_\text{q} = \text{diag}(0.01^2 \pi^2/180^2, 0.01^2 \pi^2/180^2, 0.24^2 \pi^2/180^2)$. 

Even though we downsample the data from $100$ Hz to $10$ Hz, the computational complexity of both the particle filter as well as the particle smoother for this example quickly becomes prohibitively large. The reason for this is not only the large number of data points but also the large area that is covered and the fact that the number of basis functions that is needed to make a good approximation scales with this~\citep{Solin+Sarkka:2020}. Because of this, similar to our approach in~\cite{Kok+Solin:2018}, we use smaller hexagonal block domains (each with a radius of $5$ m, a height of $2$ m and 256 basis functions), as visualised in \cref{fig:dpo-3D}. Details on the basis functions can be found in~\cite{Kok+Solin:2018}. Note that similar to~\cite{Kok+Solin:2018}, we model the actual hexagonal blocks to be slightly larger than the area that we use to reduce boundary effects. In principle, the hexagonal blocks can straightforwardly be included in \cref{alg:mcmc,alg:cpf-as}, except for line 10 in \cref{alg:cpf-as}. In that line, we compute the ancestor weights of the reference trajectory and hence compute the likelihood of every future measurement $y_{t:T}$ given the previous measurements $y_{1:t-1}$, the previous locations of that particle $x_{1:t-1}^i$ and the future locations of the reference trajectory~$x_{t:T}'$. Note that $x_{t:T}'$ can span multiple hexagons. We therefore check for the whole reference trajectory $x_{t:T}'$ in which hexagon it lies. If a map has been started in this hexagon by any of the particles, we compute the likelihood according to \cref{eq:likFutureMeasRef,eq:yCRbar} either using the estimated map or using the prior (in case that particle has not yet started that hexagon). The part of the reference trajectory that is in hexagons not created by any of the particles contributes only a constant offset on the weights and is therefore omitted.

\section{Discussion}
\label{sec:discussion}
We have presented a probabilistic approach for SLAM problems under the smoothing setup, \ie\ for conditioning the entire trajectory on all observed data. This is a particularly challenging problem as the state and parameter space become very high-dimensional, which makes the general problem largely intractable. Key to our setup is to leverage the conditionally linear structure---through Rao-Blackwellization---that separates the map parameters $\theta$ from the poses $x$.
The smoothing approach should in general not be considered a real-time method as the idea is to jointly consider all data over the entire time-horizon (incorporating knowledge from the future into past states). Compared to real-time filtering approaches, smoothing also adds to the computational load. Our method (see \cref{alg:mcmc}) requires running a conditional particle filter for each sample $k$, which would translate to a cost of $K$ times the cost of running a particle filter. However, Line~\ref{as:ancWeights} in \cref{alg:cpf-as} is the most computationally heavy step of our algorithm as the computation of $p(y_{t:T} \mid x^i_{1:t-1}, x'_{t:T}, y_{1:t-1})$ requires a prediction along the entire future trajectory at every time instance. As shown in \cref{eq:likFutureMeasRef}, this distribution is Gaussian with a covariance $\bar{C}(x'_{t:T}) P^i_{t-1} (\bar{C}(x'_{t:T}))^\Transp$. Computing this covariance has a computational complexity in the order of $\mathcal{O}(mT^2 + m^2 T)$ as $\bar{C}(x'_{t:T}) \in \mathbb{R}^{n_y (T-t) \times m}$, where $n_y$ is the dimension of the measurement at time $t$, and $P^i_{t-1} \in \mathbb{R}^{m \times m}$. This results in an overall computational complexity of \cref{alg:mcmc} of $\mathcal{O}(K N T^3 m + K N m^2 T^2)$. Note that this finding is contrary to the linear time complexity $\mathcal{O}(T)$ reported in \citet{wigrenRML:2019} even though our smoothing algorithm for SLAM can be seen as a special case of their algorithm. In their work, $p(y_{t:T} \mid x^i_{1:t-1}, x'_{t:T}, y_{1:t-1})$ cannot be computed in closed-form and because of this, they propagate sufficient statistics. We can use a similar strategy by implementing our recursive linear parameter estimation in information form. In other words, instead of estimating a mean $\hat{\theta}_t$ and covariance $P_t$, we can equivalently estimate an information vector $\iota_t = I_t \theta_t$ and an information matrix $I_t = P_t^{-1}$. The details on how our algorithm can be implemented in information form are available in \cref{appsec:informationForm}. This implementation has a computational complexity of $\mathcal{O}(K N T m^3)$. In other words, our algorithm either scales cubically with the number of time steps $T$ or cubically with the number of map parameters $m$. In practice, the implementation from \cref{appsec:informationForm} is therefore often preferred over direct implementation of \cref{alg:cpf-as}. However, in the case where the map is represented using a large number of basis functions approximating a Gaussian process, it highly depends on the exact number of basis functions and time steps which implementation is most efficient. In \cref{sec:expResults}, we partially overcame this computational complexity by only computing this quantity for hexagonal block domains that have previously been created, avoiding unnecessary computations. In future work it would be interesting to explore more ways of reducing the computational complexity, for example by only doing predictions into the near future, or by doing independent instead of joint predictions.

When using \cref{alg:cpf-as} for other SLAM problems than the ones studied in this work, care should be taken in at least two respects. Firstly, for conditionally approximately linear measurement models, the performance of \cref{alg:cpf-as} naturally depends on the quality of the linearization and will therefore be model-dependent. Furthermore, in principle one would expect that we would have to discard the first samples of the smoother to allow for convergence of the Markov chain. However, because we did not observe a clear convergence effect, throughout \cref{sec:experiments} we did not discard any samples of the smoother. We suspect that we did not observe the convergence effect since the RBPF-AS initializes the Markov chain close to the smoothing distribution using a Rao-Blackwellized particle filter. However, care should be taken when using \cref{alg:cpf-as} for other smoothing problems and it should be checked if also in those cases it is not necessary to discard samples due to burn-in. 

There are several promising and more or less straightforward extensions of \cref{alg:cpf-as} that would be interesting to explore in future work. Firstly, we use bootstrap particle filtering, in which the proposal distribution is determined by the dynamic model. Our method can straightforwardly be extended to other proposal distributions \citep[\eg, the one used in][]{Montemerlo+Thrun+Koller+Wegbreit:2003}. Second, in the case of approximately linear models, the updates of the map in \cref{eq:measUpdateParam} are similar to measurement updates in an extended Kalman filter. These updates could be replaced, \eg\ with measurement updates similar to an unscented Kalman filter. Unscented Kalman filters have been shown to perform better than extended Kalman filters for visual SLAM in \citet{solin2022alook}. Further experiments would also be needed for studying how  smoothing approaches could improve the state-of-the-art in filtering-based visual-inertial odometry and SLAM \citep[see][]{seiskari2022hybvio}.

\section{Conclusion}
\label{sec:conclusions}
This paper introduced a framework for probabilistic SLAM using particle smoothing that does not only incorporate observed data in current state estimates, but also back-tracks the updated knowledge to correct for past drift and ambiguities in both the map and in the states. Our solution can handle both dense and sparse map representations by Rao-Blackwellization of conditionally linear and conditionally linearized models.
We structured the framework to cover a variety of SLAM paradigms, both for dense function-valued maps (typically used in magnetic and radio RSSI anomaly SLAM) and sparse feature-point based maps (typically used in visual and radio RSSI emitter SLAM). The algorithm allows for modelling that the map is constant over time, and for including {\it a~priori} assumptions regarding the map, \eg important for magnetic SLAM. 

The proposed algorithm alleviates particle degeneracy, results in smooth estimated trajectories, and is robust against calibration and initialization errors---features that were all demonstrated in the experiments. Interesting directions of future work include exploring ways to reduce the computational complexity of the algorithm and changing the proposal distribution of the particle filter or the measurement updates for conditionally approximately linear models. 

\appendix

\section{Estimating the parameters in information form}
\label{appsec:informationForm}
Any recursive least squares or Kalman filter problem can be written in information form \citep{gustafsson:2012}. It is therefore possible to slightly adapt the method from \cref{sec:methods} and estimate an information vector $\iota_t = I_t \hat{\theta}_t$ and information matrix $I_t = P_t^{-1}$ rather than estimating $\hat{\theta}_t$ and $P_t$. The time recursion \cref{eq:measUpdateParam} to update the parameter estimates in information form is 
\begin{equation}
\iota_t = \iota_{t-1} + (C(x_t))^\Transp \Sigma^{-1} y_t, \qquad I_t = I_{t-1} + (C(x_t))^\Transp \Sigma^{-1} C(x_t).
\label{eq:measUpdate-inf}
\end{equation}
To compute $p(y_{t} \mid x_{1:t}, y_{1:t-1})$ using the information vector and matrix, let us first write the Gaussian distribution $p(\theta \mid x_{1:t}, y_{1:t})$ as 
\begin{equation}
p(\theta \mid x_{1:t}, y_{1:t}) = \tfrac{\exp{ \left( -\tfrac{1}{2} \iota_t^\Transp I_t^{-1} \iota_t \right)}}{\sqrt{(2 \pi)^m \det I_t^{-1}}}  \exp{\left( -\tfrac{1}{2} \left( \theta^\Transp I_t \theta - 2 \theta^\Transp \iota_t \right) \right)},
\label{eq:Gaussian-infsplit}
\end{equation}
where we separated the terms that depend on $\theta$ from the terms that do not depend on $\theta$. We can then write $p(y_{t} \mid x_{1:t}, y_{1:t-1})$ as 
\begin{align}
p(y_{t} \mid x_{1:t}, y_{1:t-1}) &= \int p(y_{t} \mid x_{t}, \theta) p( \theta \mid x_{1:t-1}, y_{1:t-1}) \dint \theta \nonumber \\
&= \tfrac{\exp{ \left( -\tfrac{1}{2} \iota_{t-1}^\Transp I_{t-1}^{-1} \iota_{t-1} \right)}}{\sqrt{(2 \pi)^m \det I_{t-1}^{-1}}} \tfrac{\exp{ \left( -\tfrac{1}{2} y_{t}^\Transp \Sigma^{-1} y_{t} \right)}}{\sqrt{(2 \pi)^{n_y} \det \Sigma}} \nonumber \\ 
& \quad \int \exp{\left( -\tfrac{1}{2} \left( \theta^\Transp (I_{t-1} + (C(x_{t}))^\Transp \Sigma^{-1} C(x_{t})) \theta - 2 \theta^\Transp ( \iota_{t-1} + (C(x_{t}))^\Transp \Sigma^{-1} y_{t}) \right) \right)} \dint \theta \nonumber \\
&= \tfrac{\exp{ \left( -\tfrac{1}{2} \iota_{t-1}^\Transp I_{t-1}^{-1} \iota_{t-1} \right)}}{\sqrt{(2 \pi)^m \det I_{t-1}^{-1}}} \tfrac{\exp{ \left( -\tfrac{1}{2} y_{t}^\Transp \Sigma^{-1} y_{t} \right)}}{\sqrt{(2 \pi)^{n_y} \det \Sigma}} \tfrac{\sqrt{(2 \pi)^m \det I_{t}^{-1}}}{\exp{ \left( -\tfrac{1}{2} \iota_{t}^\Transp I_{t}^{-1} \iota_{t} \right)}},
\label{eq:weights-inf}
\end{align}
where we used the definition of the update equations from \cref{eq:measUpdate-inf}, the fact that the term in the integral in the second step of \cref{eq:weights-inf} has the same form as \cref{eq:Gaussian-infsplit} except for the terms that are independent of $\theta$, and the fact that probability distributions integrate to one. 

To compute the ancestor weights of the reference trajectory, we can compute $p(y_{t:T} \mid x^i_{1:t-1}, x'_{t:T}, y_{1:t-1})$ as
\begin{align}
p(y_{t:T} \mid x^i_{1:t-1}, x'_{t:T}, y_{1:t-1}) 
&= \prod_{\tau = t}^{T} p(y_{\tau} \mid x^i_{1:t-1}, x'_{t:\tau}, y_{1:\tau-1}) \nonumber \\
&\propto \tfrac{\exp{ \left( -\tfrac{1}{2} (\iota^i_{t-1})^\Transp (I^i_{t-1})^{-1} \iota^i_{t-1} \right)}}{\sqrt{(2 \pi)^m \det ((I^i_{t-1})^{-1})}} \tfrac{\sqrt{(2 \pi)^m \det ((I^i_{T})^{-1})}}{\exp{ \left( -\tfrac{1}{2} (\iota^i_{T})^\Transp (I^i_{T})^{-1} \iota^i_{T} \right)}},
\label{eq:weights-ref-inf}
\end{align}
where we used \cref{eq:weights-inf} and omitted the terms that only depend on $y_t$ and $\Sigma$ since they are the same for each particle $i$. Note that the particularly elegant expression \cref{eq:weights-ref-inf} is due to the cancellation of terms because of the special structure in \cref{eq:weights-inf}. Furthermore, note that 
\begin{subequations}
\begin{align}
\iota^i_{T} &= \iota^i_0 + \underbrace{\sum_{\tau = 1}^{t-1} C(x_\tau^i)^\Transp \Sigma^{-1} y_\tau}_{\bar{\iota}^i_T} + \underbrace{\sum_{\tau = t}^{T} C(x_\tau')^\Transp \Sigma^{-1} y_\tau}_{\bar{\iota}'_T}, \\
I^i_{T} &= I^i_0 + \underbrace{\sum_{\tau = 1}^{t-1} C(x_\tau^i)^\Transp \Sigma^{-1} C(x_\tau^i)}_{\bar{I}_T^i} + \underbrace{\sum_{\tau = t}^{T} C(x_\tau')^\Transp \Sigma^{-1} C(x_\tau')}_{\bar{I}'_T},
\end{align}
\end{subequations}
where $\bar{\iota}'_T$ and $\bar{I}'_T$ are independent of the particles and their parameters estimates. Because of this, it is possible to precompute each term in the sum and only add them once per iteration of the particle filter. 

The computational complexity of both \cref{eq:weights-inf} and \cref{eq:weights-ref-inf} is dominated by the inversion of the information matrix and is of order $\mathcal{O}(m^3)$. It is possible to avoid this cubic computational complexity in \cref{eq:weights-inf} by not only estimating $\iota_t$ and $I_t$ but also $P_t$ since this directly gives $I_t^{-1}$ with only a complexity $\mathcal{O}(m^2)$ and since
\begin{equation}
\det I_{t}^{-1} = \tfrac{1}{\det \left( I_{t-1} + (C(x_t))^\Transp \Sigma^{-1} C(x_t) \right)} = \tfrac{1}{\det \left( \Sigma + C(x_t) I_{t-1}^{-1} (C(x_t))^\Transp \right) \det \left( \Sigma^{-1} \right) \det \left( I_{t-1} \right)} = \tfrac{\det \left( \Sigma \right) \det \left( P_{t-1} \right)}{\det \left( \Sigma + C(x_t) P_{t-1} (C(x_t))^\Transp \right)}.
\end{equation}
It is, however, not possible to avoid the cubic complexity in \cref{eq:weights-ref-inf} without introducing additional scaling with $T$ as computing $P_T$ from $P_{t-1}$ is of order $\mathcal{O}(T m^2)$ for every particle and every time instance.

\begin{Backmatter}

\begin{mybackmatter}

\paragraph{Acknowledgments}
We would like to thank Anna Wigren and Fredrik Lindsten for the technical discussions related to the computational complexity of Line~\ref{as:ancWeights} in \cref{alg:cpf-as}. We would also like to thank the reviewers for their detailed and important suggestions that helped us improve the quality of the manuscript. 

\paragraph{Funding statement}
This publication is part of the project `\emph{Sensor Fusion for Indoor Localisation Using the Magnetic Field}' with project number 18213 of the research program Veni which is (partly) financed by the Dutch Research Council (NWO). The research was also supported by grants from the Academy of Finland (grant id: 339730), by the project `\emph{NewLEADS -- New Directions in Learning Dynamical Systems}' (contract number: 621-2016-06079) funded by the Swedish Research Council, and by the Kjell och M{\"a}rta Beijer Foundation. We acknowledge the computational resources provided by the Aalto Science-IT project. 

\paragraph{Competing interests}
None.%

\paragraph{Data availability statement}
We provide proof-of-concept reference implementations and data for replicating the experiments. This code and data can be found on \url{https://github.com/manonkok/Rao-Blackwellized-SLAM-smoothing}.

\paragraph{Ethical standards}
The research meets all ethical guidelines, including adherence to the legal requirements of the study country.

\paragraph{Author contributions}
Conceptualization: M.K; A.S; T.S. Methodology: M.K; A.S; T.S. Data curation: M.K; A.S. Data visualisation: M.K; A.S. Writing original draft: M.K; A.S. All authors approved the final submitted draft.

\end{mybackmatter}

\bibliographystyle{apalike}

\end{Backmatter}

\end{document}

%% file: fig/visual.tex
%
%
\definecolor{mycolor1}{rgb}{0.24220,0.15040,0.66030}%
\definecolor{mycolor2}{rgb}{0.26673,0.20283,0.80872}%
\definecolor{mycolor3}{rgb}{0.27974,0.26993,0.91412}%
\definecolor{mycolor4}{rgb}{0.27962,0.34352,0.97102}%
\definecolor{mycolor5}{rgb}{0.25671,0.41847,0.99618}%
\definecolor{mycolor6}{rgb}{0.18667,0.49808,0.98414}%
\definecolor{mycolor7}{rgb}{0.17002,0.56813,0.93741}%
\definecolor{mycolor8}{rgb}{0.13555,0.63146,0.89542}%
\definecolor{mycolor9}{rgb}{0.08307,0.68786,0.84942}%
\definecolor{mycolor10}{rgb}{0.00401,0.72955,0.77015}%
\definecolor{mycolor11}{rgb}{0.14657,0.75974,0.67971}%
\definecolor{mycolor12}{rgb}{0.22907,0.78804,0.57574}%
\definecolor{mycolor13}{rgb}{0.38016,0.80262,0.44482}%
\definecolor{mycolor14}{rgb}{0.56748,0.79364,0.29984}%
\definecolor{mycolor15}{rgb}{0.74549,0.76566,0.17894}%
\definecolor{mycolor16}{rgb}{0.89336,0.73476,0.17121}%
\definecolor{mycolor17}{rgb}{0.99607,0.74446,0.23623}%
\definecolor{mycolor18}{rgb}{0.98569,0.82278,0.18414}%
\definecolor{mycolor19}{rgb}{0.95954,0.90583,0.14626}%
\definecolor{mycolor20}{rgb}{0.97690,0.98390,0.08050}%
\begin{tikzpicture}

\begin{axis}[%
point meta min=0,
point meta max=255,
axis on top,
xmin=-5,
xmax=5,
ymin=-5,
ymax=5,
axis line style={draw=none},
ticks=none,
width=\figurewidth,
height=\figureheight,
scale only axis
]
\addplot [forget plot] graphics [xmin=-5.00357142857143, xmax=5.00357142857143, ymin=-5.00357142857143, ymax=5.00357142857143] {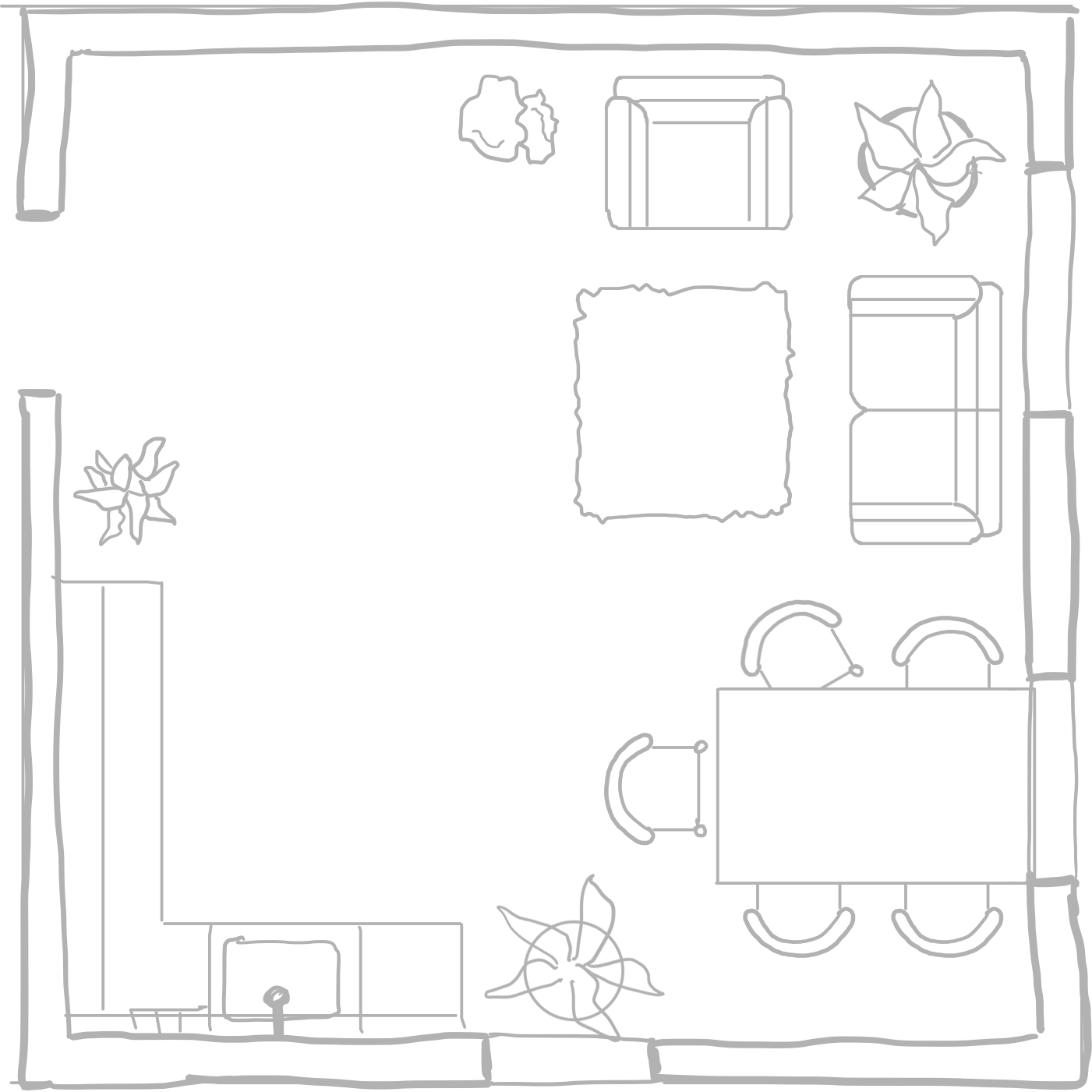};

\addplot[area legend, dashed, draw=black, fill=white!50!blue, fill opacity=0.05, forget plot]
table[row sep=crcr] {%
x	y\\
-0.812992281766924	3.31653269805099\\
3.26051863392883	-20.3727958402906\\
-21.1133298111223	-9.55491143118416\\
-0.812992281766924	3.31653269805099\\
}--cycle;
\addplot[only marks, mark=*, mark options={}, mark size=1.4142pt, color=mycolor1, fill=mycolor1, forget plot] table[row sep=crcr]{%
x	y\\
-3.4119	0.2365\\
};
\addplot[only marks, mark=*, mark options={}, mark size=1.4142pt, color=mycolor2, fill=mycolor2, forget plot] table[row sep=crcr]{%
x	y\\
-0.7658	-3.452\\
};
\addplot[only marks, mark=*, mark options={}, mark size=1.4142pt, color=mycolor3, fill=mycolor3, forget plot] table[row sep=crcr]{%
x	y\\
2.8426	0.0922\\
};
\addplot[only marks, mark=*, mark options={}, mark size=1.4142pt, color=mycolor4, fill=mycolor4, forget plot] table[row sep=crcr]{%
x	y\\
0.6455	2.9308\\
};
\addplot[only marks, mark=*, mark options={}, mark size=1.4142pt, color=mycolor5, fill=mycolor5, forget plot] table[row sep=crcr]{%
x	y\\
-3.5242	0.9903\\
};
\addplot[only marks, mark=*, mark options={}, mark size=1.4142pt, color=mycolor6, fill=mycolor6, forget plot] table[row sep=crcr]{%
x	y\\
3.2916	-1.0464\\
};
\addplot[only marks, mark=*, mark options={}, mark size=1.4142pt, color=mycolor7, fill=mycolor7, forget plot] table[row sep=crcr]{%
x	y\\
2.8105	2.4336\\
};
\addplot[only marks, mark=*, mark options={}, mark size=1.4142pt, color=mycolor8, fill=mycolor8, forget plot] table[row sep=crcr]{%
x	y\\
-4.4543	3.043\\
};
\addplot[only marks, mark=*, mark options={}, mark size=1.4142pt, color=mycolor9, fill=mycolor9, forget plot] table[row sep=crcr]{%
x	y\\
-3.5242	-0.3248\\
};
\addplot[only marks, mark=*, mark options={}, mark size=1.4142pt, color=mycolor10, fill=mycolor10, forget plot] table[row sep=crcr]{%
x	y\\
1.4153	-1.8162\\
};
\addplot[only marks, mark=*, mark options={}, mark size=1.4142pt, color=mycolor11, fill=mycolor11, forget plot] table[row sep=crcr]{%
x	y\\
2.8586	1.2629\\
};
\addplot[only marks, mark=*, mark options={}, mark size=1.4142pt, color=mycolor12, fill=mycolor12, forget plot] table[row sep=crcr]{%
x	y\\
0.421	-3.035\\
};
\addplot[only marks, mark=*, mark options={}, mark size=1.4142pt, color=mycolor13, fill=mycolor13, forget plot] table[row sep=crcr]{%
x	y\\
-0.413	-3.2916\\
};
\addplot[only marks, mark=*, mark options={}, mark size=1.4142pt, color=mycolor14, fill=mycolor14, forget plot] table[row sep=crcr]{%
x	y\\
-3.4921	-3.4199\\
};
\addplot[only marks, mark=*, mark options={}, mark size=1.4142pt, color=mycolor15, fill=mycolor15, forget plot] table[row sep=crcr]{%
x	y\\
1.9285	-1.1106\\
};
\addplot[only marks, mark=*, mark options={}, mark size=1.4142pt, color=mycolor16, fill=mycolor16, forget plot] table[row sep=crcr]{%
x	y\\
2.8907	3.2034\\
};
\addplot[only marks, mark=*, mark options={}, mark size=1.4142pt, color=mycolor17, fill=mycolor17, forget plot] table[row sep=crcr]{%
x	y\\
-4.326	4.5024\\
};
\addplot[only marks, mark=*, mark options={}, mark size=1.4142pt, color=mycolor18, fill=mycolor18, forget plot] table[row sep=crcr]{%
x	y\\
4.5104	-4.4142\\
};
\addplot[only marks, mark=*, mark options={}, mark size=1.4142pt, color=mycolor19, fill=mycolor19, forget plot] table[row sep=crcr]{%
x	y\\
-0.3007	3.5723\\
};
\addplot[only marks, mark=*, mark options={}, mark size=1.4142pt, color=mycolor20, fill=mycolor20, forget plot] table[row sep=crcr]{%
x	y\\
2.2171	2.9147\\
};
\addplot [color=black, only marks, mark size=5.0pt, mark=x, mark options={solid, black}, forget plot]
  table[row sep=crcr]{%
2.8426	0.0922\\
0.6455	2.9308\\
3.2916	-1.0464\\
2.8105	2.4336\\
-4.4543	3.043\\
1.4153	-1.8162\\
2.8586	1.2629\\
0.421	-3.035\\
1.9285	-1.1106\\
2.8907	3.2034\\
-4.326	4.5024\\
4.5104	-4.4142\\
-0.3007	3.5723\\
2.2171	2.9147\\
};
\addplot [color=black, only marks, mark size=5.0pt, mark=*, mark options={solid, fill=black, black}, forget plot]
  table[row sep=crcr]{%
-0.812992281766924	3.31653269805099\\
};
\addplot [color=black, line width=2.0pt, forget plot]
  table[row sep=crcr]{%
-0.812992281766924	3.31653269805099\\
-0.71115450887453	2.72429948459245\\
-1.32050072000081	2.99474659482011\\
-0.812992281766924	3.31653269805099\\
};
\addplot [color=black, dashed, line width=1.2pt, forget plot]
  table[row sep=crcr]{%
3.5	-1.5\\
3.47636095353085	-1.25456679474552\\
3.40838335801488	-1.0146583078651\\
3.30118235203317	-0.784977946290424\\
3.16063260129131	-0.569211294728816\\
2.99312404909659	-0.369933940805933\\
2.80530524513046	-0.188567834220353\\
2.60382428266956	-0.0253873254994235\\
2.39507739303089	0.120425942686067\\
2.18497487529805	0.250682053533507\\
1.97873329440414	0.368040327032806\\
1.78070179096999	0.47583154503926\\
1.59422895482957	0.577852501609632\\
1.4215750754922	0.678141577768131\\
1.26387276101681	0.780744783796319\\
1.12113698308925	0.889482083043456\\
0.992323635978317	1.00772379474996\\
0.875433767343679	1.13818647015975\\
0.767658824812068	1.28275686200894\\
0.665560634522176	1.44235149249745\\
0.56527844988794	1.61681791252078\\
0.462754334329808	1.80488209028887\\
0.353967412543315	2.004144534959\\
0.235167169440886	2.21112582198052\\
0.103096008097972	2.42136021708417\\
-0.044808303342732	2.62953417208612\\
-0.210238877044148	2.82966466301533\\
-0.393882096007434	3.01531072995885\\
-0.595322178620682	3.17981022151298\\
-0.812992281766924	3.31653269805099\\
};
\end{axis}
\end{tikzpicture}%

%% file: fig/radio.tex
%
%
\begin{tikzpicture}

\begin{axis}[%
axis on top,
xmin=-5,
xmax=5,
ymin=-5,
ymax=5,
axis line style={draw=none},
ticks=none,
axis x line*=bottom,
axis y line*=left,
width=\figurewidth,
height=\figureheight,
scale only axis
]
\addplot [forget plot] graphics [xmin=-5.00357142857143, xmax=5.00357142857143, ymin=-5.00357142857143, ymax=5.00357142857143] {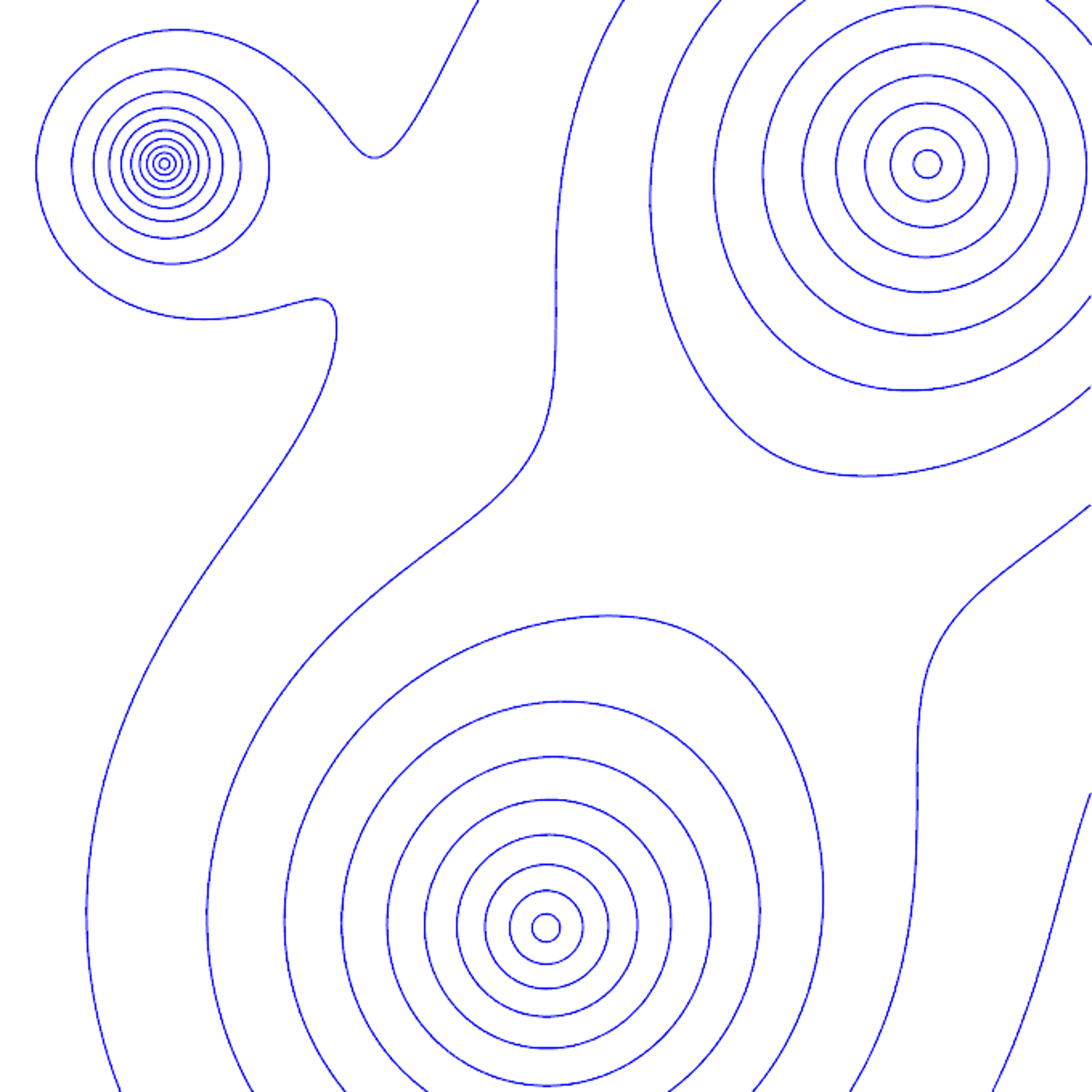};
\addplot [forget plot] graphics [xmin=-5.00357142857143, xmax=5.00357142857143, ymin=-5.00357142857143, ymax=5.00357142857143] {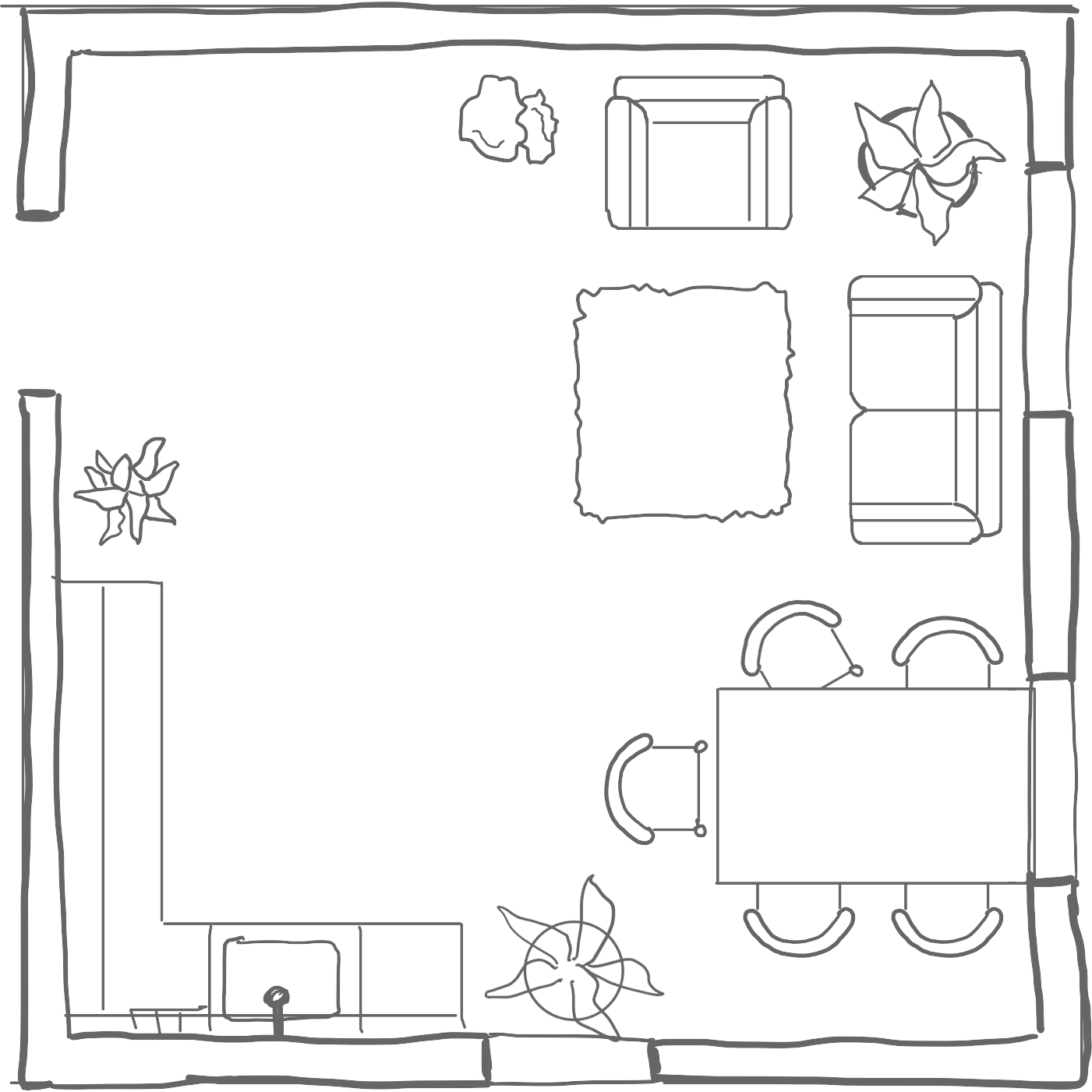};
\addplot[only marks, mark=*, mark options={}, mark size=7.0711pt, color=white!50!blue, fill=white!50!blue, forget plot] table[row sep=crcr]{%
x	y\\
-0.812992281766924	3.31653269805099\\
};
\addplot[only marks, mark=*, mark options={}, mark size=1.9365pt, color=black, fill=black, forget plot] table[row sep=crcr]{%
x	y\\
-0.812992281766924	3.31653269805099\\
};
\addplot [color=black, dashed, line width=1.2pt, forget plot]
  table[row sep=crcr]{%
3.5	-1.5\\
3.47636095353085	-1.25456679474552\\
3.40838335801488	-1.0146583078651\\
3.30118235203317	-0.784977946290424\\
3.16063260129131	-0.569211294728816\\
2.99312404909659	-0.369933940805933\\
2.80530524513046	-0.188567834220353\\
2.60382428266956	-0.0253873254994235\\
2.39507739303089	0.120425942686067\\
2.18497487529805	0.250682053533507\\
1.97873329440414	0.368040327032806\\
1.78070179096999	0.47583154503926\\
1.59422895482957	0.577852501609632\\
1.4215750754922	0.678141577768131\\
1.26387276101681	0.780744783796319\\
1.12113698308925	0.889482083043456\\
0.992323635978317	1.00772379474996\\
0.875433767343679	1.13818647015975\\
0.767658824812068	1.28275686200894\\
0.665560634522176	1.44235149249745\\
0.56527844988794	1.61681791252078\\
0.462754334329808	1.80488209028887\\
0.353967412543315	2.004144534959\\
0.235167169440886	2.21112582198052\\
0.103096008097972	2.42136021708417\\
-0.044808303342732	2.62953417208612\\
-0.210238877044148	2.82966466301533\\
-0.393882096007434	3.01531072995885\\
-0.595322178620682	3.17981022151298\\
-0.812992281766924	3.31653269805099\\
};
\node[align=center]
at (axis cs:-3.5,3.5) {\huge\faBroadcastTower};
\node[align=center]
at (axis cs:3.5,3.5) {\huge\faBroadcastTower};
\node[align=center]
at (axis cs:0,-3.5) {\huge\faBroadcastTower};
\end{axis}
\end{tikzpicture}%

%% file: fig/magnetic.tex
%
%
\begin{tikzpicture}

\begin{axis}[%
axis on top,
unbounded coords=jump,
xmin=-5,
xmax=5,
ymin=-5,
ymax=5,
axis line style={draw=none},
ticks=none,
axis x line*=bottom,
axis y line*=left,
width=\figurewidth,
height=\figureheight,
scale only axis
]
\addplot [forget plot] graphics [xmin=-5.00357142857143, xmax=5.00357142857143, ymin=-5.00357142857143, ymax=5.00357142857143] {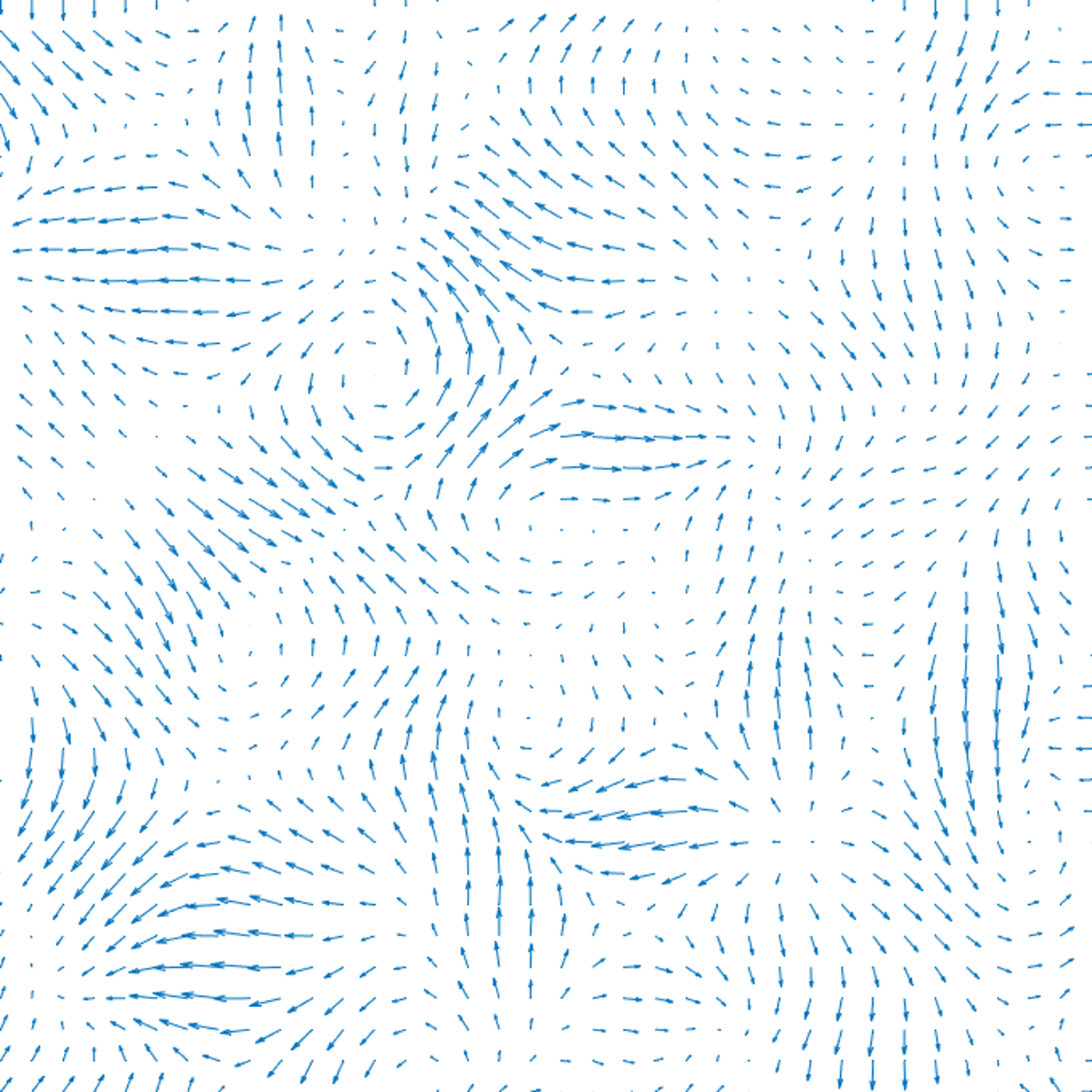};
\addplot [forget plot] graphics [xmin=-5.00357142857143, xmax=5.00357142857143, ymin=-5.00357142857143, ymax=5.00357142857143] {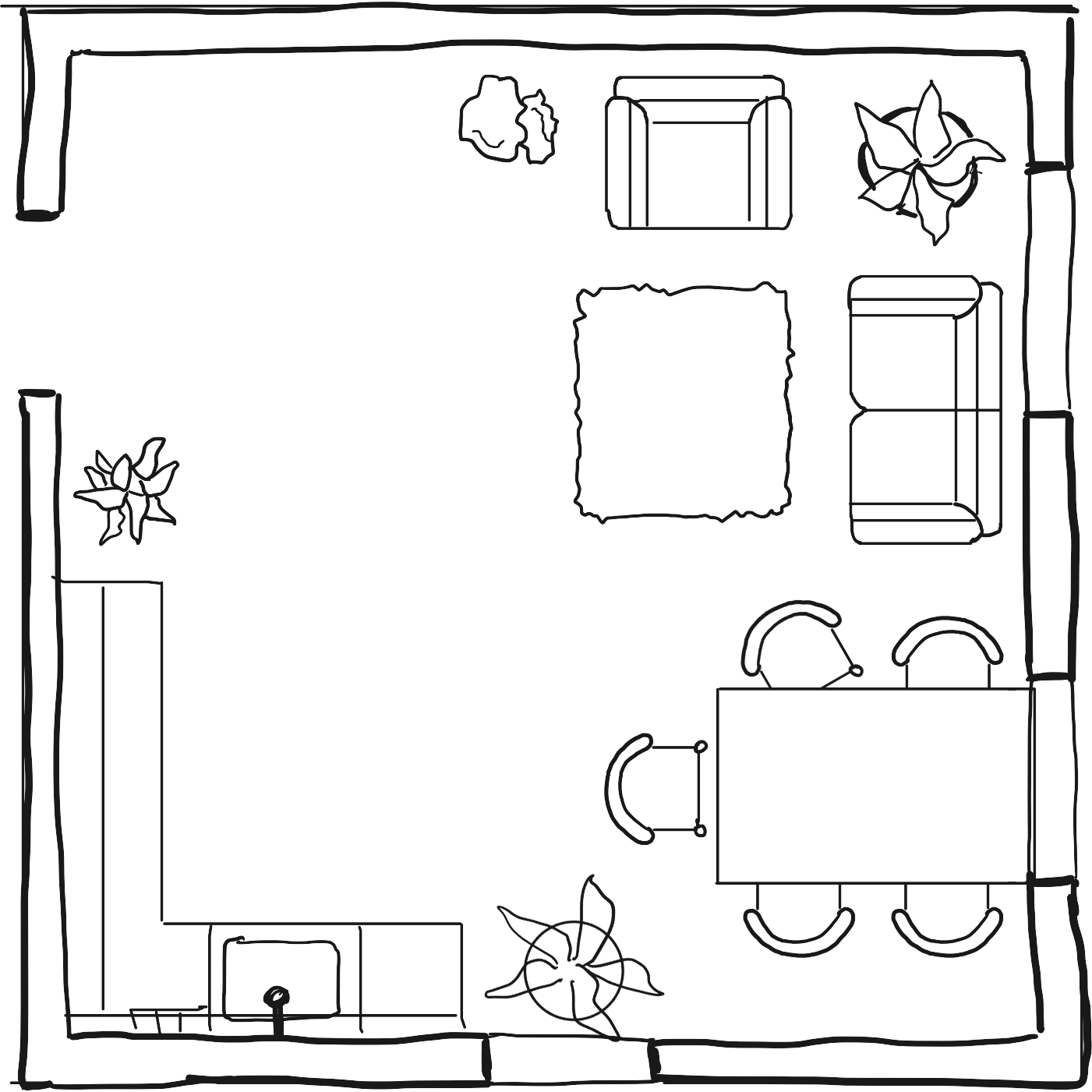};
\addplot[only marks, mark=o, mark options={}, mark size=7.0711pt, draw=black, forget plot] table[row sep=crcr]{%
x	y\\
-0.812992281766924	3.31653269805099\\
};
\addplot[only marks, mark=*, mark options={}, mark size=1.9365pt, color=black, fill=black, forget plot] table[row sep=crcr]{%
x	y\\
-0.812992281766924	3.31653269805099\\
};
\addplot [color=black, dashed, line width=1.2pt, forget plot]
  table[row sep=crcr]{%
3.5	-1.5\\
3.47636095353085	-1.25456679474552\\
3.40838335801488	-1.0146583078651\\
3.30118235203317	-0.784977946290424\\
3.16063260129131	-0.569211294728816\\
2.99312404909659	-0.369933940805933\\
2.80530524513046	-0.188567834220353\\
2.60382428266956	-0.0253873254994235\\
2.39507739303089	0.120425942686067\\
2.18497487529805	0.250682053533507\\
1.97873329440414	0.368040327032806\\
1.78070179096999	0.47583154503926\\
1.59422895482957	0.577852501609632\\
1.4215750754922	0.678141577768131\\
1.26387276101681	0.780744783796319\\
1.12113698308925	0.889482083043456\\
0.992323635978317	1.00772379474996\\
0.875433767343679	1.13818647015975\\
0.767658824812068	1.28275686200894\\
0.665560634522176	1.44235149249745\\
0.56527844988794	1.61681791252078\\
0.462754334329808	1.80488209028887\\
0.353967412543315	2.004144534959\\
0.235167169440886	2.21112582198052\\
0.103096008097972	2.42136021708417\\
-0.044808303342732	2.62953417208612\\
-0.210238877044148	2.82966466301533\\
-0.393882096007434	3.01531072995885\\
-0.595322178620682	3.17981022151298\\
-0.812992281766924	3.31653269805099\\
};
\addplot [color=black, line width=1.2pt, forget plot]
  table[row sep=crcr]{%
-0.812992281766924	3.31653269805099\\
-0.950320737827481	3.61660084890347\\
-0.982868676785864	3.52912152752086\\
nan	nan\\
-0.950320737827481	3.61660084890347\\
-0.862841416444874	3.58405290994508\\
};
\end{axis}
\end{tikzpicture}%

%% file: fig/line-smoother.tex
%
%
\begin{tikzpicture}

\begin{axis}[%
point meta min=-4.22960266259245,
point meta max=4.22960266259245,
axis on top,
xmin=-0.7,
xmax=0.7,
xtick={-0.5,0,0.5},
xticklabels={\empty},
ymin=-2,
ymax=2,
ytick={-2,-1.5,-1,-0.5,0,0.5,1,1.5,2},
yticklabels={\empty},
axis background/.style={fill=white},
width=\figurewidth,
height=\figureheight,
scale only axis
]
\addplot [forget plot] graphics [xmin=-0.702928870292887, xmax=0.702928870292887, ymin=-2.00292397660819, ymax=2.00292397660819] {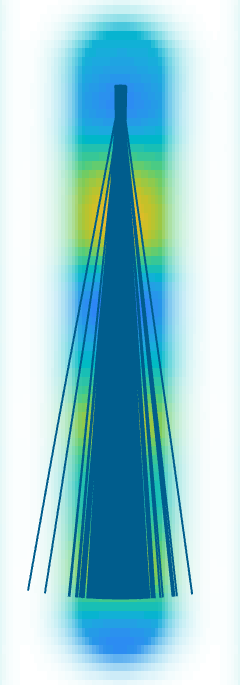};
\end{axis}
\end{tikzpicture}%

%% file: fig/mag-path-field.tex
%
%
\definecolor{mycolor1}{rgb}{0.00000,0.36471,0.55294}%
\begin{tikzpicture}

\begin{axis}[%
point meta min=8.55412011009386,
point meta max=79.0690737604038,
axis on top,
xmin=-12.0206409562931,
xmax=12.0206409562931,
xtick={-10,-5,0,5,10},
ymin=-12.0206421963723,
ymax=12.0206421963723,
ytick={-10,-5,0,5,10},
yticklabels={\empty},
axis background/.style={fill=white},
axis x line*=bottom,
axis y line*=left,
width=\figurewidth,
height=\figureheight,
scale only axis,
colormap={mymap}{[1pt] rgb(0pt)=(0.2422,0.1504,0.6603); rgb(1pt)=(0.2444,0.1534,0.6728); rgb(2pt)=(0.2464,0.1569,0.6847); rgb(3pt)=(0.2484,0.1607,0.6961); rgb(4pt)=(0.2503,0.1648,0.7071); rgb(5pt)=(0.2522,0.1689,0.7179); rgb(6pt)=(0.254,0.1732,0.7286); rgb(7pt)=(0.2558,0.1773,0.7393); rgb(8pt)=(0.2576,0.1814,0.7501); rgb(9pt)=(0.2594,0.1854,0.761); rgb(11pt)=(0.2628,0.1932,0.7828); rgb(12pt)=(0.2645,0.1972,0.7937); rgb(13pt)=(0.2661,0.2011,0.8043); rgb(14pt)=(0.2676,0.2052,0.8148); rgb(15pt)=(0.2691,0.2094,0.8249); rgb(16pt)=(0.2704,0.2138,0.8346); rgb(17pt)=(0.2717,0.2184,0.8439); rgb(18pt)=(0.2729,0.2231,0.8528); rgb(19pt)=(0.274,0.228,0.8612); rgb(20pt)=(0.2749,0.233,0.8692); rgb(21pt)=(0.2758,0.2382,0.8767); rgb(22pt)=(0.2766,0.2435,0.884); rgb(23pt)=(0.2774,0.2489,0.8908); rgb(24pt)=(0.2781,0.2543,0.8973); rgb(25pt)=(0.2788,0.2598,0.9035); rgb(26pt)=(0.2794,0.2653,0.9094); rgb(27pt)=(0.2798,0.2708,0.915); rgb(28pt)=(0.2802,0.2764,0.9204); rgb(29pt)=(0.2806,0.2819,0.9255); rgb(30pt)=(0.2809,0.2875,0.9305); rgb(31pt)=(0.2811,0.293,0.9352); rgb(32pt)=(0.2813,0.2985,0.9397); rgb(33pt)=(0.2814,0.304,0.9441); rgb(34pt)=(0.2814,0.3095,0.9483); rgb(35pt)=(0.2813,0.315,0.9524); rgb(36pt)=(0.2811,0.3204,0.9563); rgb(37pt)=(0.2809,0.3259,0.96); rgb(38pt)=(0.2807,0.3313,0.9636); rgb(39pt)=(0.2803,0.3367,0.967); rgb(40pt)=(0.2798,0.3421,0.9702); rgb(41pt)=(0.2791,0.3475,0.9733); rgb(42pt)=(0.2784,0.3529,0.9763); rgb(43pt)=(0.2776,0.3583,0.9791); rgb(44pt)=(0.2766,0.3638,0.9817); rgb(45pt)=(0.2754,0.3693,0.984); rgb(46pt)=(0.2741,0.3748,0.9862); rgb(47pt)=(0.2726,0.3804,0.9881); rgb(48pt)=(0.271,0.386,0.9898); rgb(49pt)=(0.2691,0.3916,0.9912); rgb(50pt)=(0.267,0.3973,0.9924); rgb(51pt)=(0.2647,0.403,0.9935); rgb(52pt)=(0.2621,0.4088,0.9946); rgb(53pt)=(0.2591,0.4145,0.9955); rgb(54pt)=(0.2556,0.4203,0.9965); rgb(55pt)=(0.2517,0.4261,0.9974); rgb(56pt)=(0.2473,0.4319,0.9983); rgb(57pt)=(0.2424,0.4378,0.9991); rgb(58pt)=(0.2369,0.4437,0.9996); rgb(59pt)=(0.2311,0.4497,0.9995); rgb(60pt)=(0.225,0.4559,0.9985); rgb(61pt)=(0.2189,0.462,0.9968); rgb(62pt)=(0.2128,0.4682,0.9948); rgb(63pt)=(0.2066,0.4743,0.9926); rgb(64pt)=(0.2006,0.4803,0.9906); rgb(65pt)=(0.195,0.4861,0.9887); rgb(66pt)=(0.1903,0.4919,0.9867); rgb(67pt)=(0.1869,0.4975,0.9844); rgb(68pt)=(0.1847,0.503,0.9819); rgb(69pt)=(0.1831,0.5084,0.9793); rgb(70pt)=(0.1818,0.5138,0.9766); rgb(71pt)=(0.1806,0.5191,0.9738); rgb(72pt)=(0.1795,0.5244,0.9709); rgb(73pt)=(0.1785,0.5296,0.9677); rgb(74pt)=(0.1778,0.5349,0.9641); rgb(75pt)=(0.1773,0.5401,0.9602); rgb(76pt)=(0.1768,0.5452,0.956); rgb(77pt)=(0.1764,0.5504,0.9516); rgb(78pt)=(0.1755,0.5554,0.9473); rgb(79pt)=(0.174,0.5605,0.9432); rgb(80pt)=(0.1716,0.5655,0.9393); rgb(81pt)=(0.1686,0.5705,0.9357); rgb(82pt)=(0.1649,0.5755,0.9323); rgb(83pt)=(0.161,0.5805,0.9289); rgb(84pt)=(0.1573,0.5854,0.9254); rgb(85pt)=(0.154,0.5902,0.9218); rgb(86pt)=(0.1513,0.595,0.9182); rgb(87pt)=(0.1492,0.5997,0.9147); rgb(88pt)=(0.1475,0.6043,0.9113); rgb(89pt)=(0.1461,0.6089,0.908); rgb(90pt)=(0.1446,0.6135,0.905); rgb(91pt)=(0.1429,0.618,0.9022); rgb(92pt)=(0.1408,0.6226,0.8998); rgb(93pt)=(0.1383,0.6272,0.8975); rgb(94pt)=(0.1354,0.6317,0.8953); rgb(95pt)=(0.1321,0.6363,0.8932); rgb(96pt)=(0.1288,0.6408,0.891); rgb(97pt)=(0.1253,0.6453,0.8887); rgb(98pt)=(0.1219,0.6497,0.8862); rgb(99pt)=(0.1185,0.6541,0.8834); rgb(100pt)=(0.1152,0.6584,0.8804); rgb(101pt)=(0.1119,0.6627,0.877); rgb(102pt)=(0.1085,0.6669,0.8734); rgb(103pt)=(0.1048,0.671,0.8695); rgb(104pt)=(0.1009,0.675,0.8653); rgb(105pt)=(0.0964,0.6789,0.8609); rgb(106pt)=(0.0914,0.6828,0.8562); rgb(107pt)=(0.0855,0.6865,0.8513); rgb(108pt)=(0.0789,0.6902,0.8462); rgb(109pt)=(0.0713,0.6938,0.8409); rgb(110pt)=(0.0628,0.6972,0.8355); rgb(111pt)=(0.0535,0.7006,0.8299); rgb(112pt)=(0.0433,0.7039,0.8242); rgb(113pt)=(0.0328,0.7071,0.8183); rgb(114pt)=(0.0234,0.7103,0.8124); rgb(115pt)=(0.0155,0.7133,0.8064); rgb(116pt)=(0.0091,0.7163,0.8003); rgb(117pt)=(0.0046,0.7192,0.7941); rgb(118pt)=(0.0019,0.722,0.7878); rgb(119pt)=(0.0009,0.7248,0.7815); rgb(120pt)=(0.0018,0.7275,0.7752); rgb(121pt)=(0.0046,0.7301,0.7688); rgb(122pt)=(0.0094,0.7327,0.7623); rgb(123pt)=(0.0162,0.7352,0.7558); rgb(124pt)=(0.0253,0.7376,0.7492); rgb(125pt)=(0.0369,0.74,0.7426); rgb(126pt)=(0.0504,0.7423,0.7359); rgb(127pt)=(0.0638,0.7446,0.7292); rgb(128pt)=(0.077,0.7468,0.7224); rgb(129pt)=(0.0899,0.7489,0.7156); rgb(130pt)=(0.1023,0.751,0.7088); rgb(131pt)=(0.1141,0.7531,0.7019); rgb(132pt)=(0.1252,0.7552,0.695); rgb(133pt)=(0.1354,0.7572,0.6881); rgb(134pt)=(0.1448,0.7593,0.6812); rgb(135pt)=(0.1532,0.7614,0.6741); rgb(136pt)=(0.1609,0.7635,0.6671); rgb(137pt)=(0.1678,0.7656,0.6599); rgb(138pt)=(0.1741,0.7678,0.6527); rgb(139pt)=(0.1799,0.7699,0.6454); rgb(140pt)=(0.1853,0.7721,0.6379); rgb(141pt)=(0.1905,0.7743,0.6303); rgb(142pt)=(0.1954,0.7765,0.6225); rgb(143pt)=(0.2003,0.7787,0.6146); rgb(144pt)=(0.2061,0.7808,0.6065); rgb(145pt)=(0.2118,0.7828,0.5983); rgb(146pt)=(0.2178,0.7849,0.5899); rgb(147pt)=(0.2244,0.7869,0.5813); rgb(148pt)=(0.2318,0.7887,0.5725); rgb(149pt)=(0.2401,0.7905,0.5636); rgb(150pt)=(0.2491,0.7922,0.5546); rgb(151pt)=(0.2589,0.7937,0.5454); rgb(152pt)=(0.2695,0.7951,0.536); rgb(153pt)=(0.2809,0.7964,0.5266); rgb(154pt)=(0.2929,0.7975,0.517); rgb(155pt)=(0.3052,0.7985,0.5074); rgb(156pt)=(0.3176,0.7994,0.4975); rgb(157pt)=(0.3301,0.8002,0.4876); rgb(158pt)=(0.3424,0.8009,0.4774); rgb(159pt)=(0.3548,0.8016,0.4669); rgb(160pt)=(0.3671,0.8021,0.4563); rgb(161pt)=(0.3795,0.8026,0.4454); rgb(162pt)=(0.3921,0.8029,0.4344); rgb(163pt)=(0.405,0.8031,0.4233); rgb(164pt)=(0.4184,0.803,0.4122); rgb(165pt)=(0.4322,0.8028,0.4013); rgb(166pt)=(0.4463,0.8024,0.3904); rgb(167pt)=(0.4608,0.8018,0.3797); rgb(168pt)=(0.4753,0.8011,0.3691); rgb(169pt)=(0.4899,0.8002,0.3586); rgb(170pt)=(0.5044,0.7993,0.348); rgb(171pt)=(0.5187,0.7982,0.3374); rgb(172pt)=(0.5329,0.797,0.3267); rgb(173pt)=(0.547,0.7957,0.3159); rgb(175pt)=(0.5748,0.7929,0.2941); rgb(176pt)=(0.5886,0.7913,0.2833); rgb(177pt)=(0.6024,0.7896,0.2726); rgb(178pt)=(0.6161,0.7878,0.2622); rgb(179pt)=(0.6297,0.7859,0.2521); rgb(180pt)=(0.6433,0.7839,0.2423); rgb(181pt)=(0.6567,0.7818,0.2329); rgb(182pt)=(0.6701,0.7796,0.2239); rgb(183pt)=(0.6833,0.7773,0.2155); rgb(184pt)=(0.6963,0.775,0.2075); rgb(185pt)=(0.7091,0.7727,0.1998); rgb(186pt)=(0.7218,0.7703,0.1924); rgb(187pt)=(0.7344,0.7679,0.1852); rgb(188pt)=(0.7468,0.7654,0.1782); rgb(189pt)=(0.759,0.7629,0.1717); rgb(190pt)=(0.771,0.7604,0.1658); rgb(191pt)=(0.7829,0.7579,0.1608); rgb(192pt)=(0.7945,0.7554,0.157); rgb(193pt)=(0.806,0.7529,0.1546); rgb(194pt)=(0.8172,0.7505,0.1535); rgb(195pt)=(0.8281,0.7481,0.1536); rgb(196pt)=(0.8389,0.7457,0.1546); rgb(197pt)=(0.8495,0.7435,0.1564); rgb(198pt)=(0.86,0.7413,0.1587); rgb(199pt)=(0.8703,0.7392,0.1615); rgb(200pt)=(0.8804,0.7372,0.165); rgb(201pt)=(0.8903,0.7353,0.1695); rgb(202pt)=(0.9,0.7336,0.1749); rgb(203pt)=(0.9093,0.7321,0.1815); rgb(204pt)=(0.9184,0.7308,0.189); rgb(205pt)=(0.9272,0.7298,0.1973); rgb(206pt)=(0.9357,0.729,0.2061); rgb(207pt)=(0.944,0.7285,0.2151); rgb(208pt)=(0.9523,0.7284,0.2237); rgb(209pt)=(0.9606,0.7285,0.2312); rgb(210pt)=(0.9689,0.7292,0.2373); rgb(211pt)=(0.977,0.7304,0.2418); rgb(212pt)=(0.9842,0.733,0.2446); rgb(213pt)=(0.99,0.7365,0.2429); rgb(214pt)=(0.9946,0.7407,0.2394); rgb(215pt)=(0.9966,0.7458,0.2351); rgb(216pt)=(0.9971,0.7513,0.2309); rgb(217pt)=(0.9972,0.7569,0.2267); rgb(218pt)=(0.9971,0.7626,0.2224); rgb(219pt)=(0.9969,0.7683,0.2181); rgb(220pt)=(0.9966,0.774,0.2138); rgb(221pt)=(0.9962,0.7798,0.2095); rgb(222pt)=(0.9957,0.7856,0.2053); rgb(223pt)=(0.9949,0.7915,0.2012); rgb(224pt)=(0.9938,0.7974,0.1974); rgb(225pt)=(0.9923,0.8034,0.1939); rgb(226pt)=(0.9906,0.8095,0.1906); rgb(227pt)=(0.9885,0.8156,0.1875); rgb(228pt)=(0.9861,0.8218,0.1846); rgb(229pt)=(0.9835,0.828,0.1817); rgb(230pt)=(0.9807,0.8342,0.1787); rgb(231pt)=(0.9778,0.8404,0.1757); rgb(232pt)=(0.9748,0.8467,0.1726); rgb(233pt)=(0.972,0.8529,0.1695); rgb(234pt)=(0.9694,0.8591,0.1665); rgb(235pt)=(0.9671,0.8654,0.1636); rgb(236pt)=(0.9651,0.8716,0.1608); rgb(237pt)=(0.9634,0.8778,0.1582); rgb(238pt)=(0.9619,0.884,0.1557); rgb(239pt)=(0.9608,0.8902,0.1532); rgb(240pt)=(0.9601,0.8963,0.1507); rgb(241pt)=(0.9596,0.9023,0.148); rgb(242pt)=(0.9595,0.9084,0.145); rgb(243pt)=(0.9597,0.9143,0.1418); rgb(244pt)=(0.9601,0.9203,0.1382); rgb(245pt)=(0.9608,0.9262,0.1344); rgb(246pt)=(0.9618,0.932,0.1304); rgb(247pt)=(0.9629,0.9379,0.1261); rgb(248pt)=(0.9642,0.9437,0.1216); rgb(249pt)=(0.9657,0.9494,0.1168); rgb(250pt)=(0.9674,0.9552,0.1116); rgb(251pt)=(0.9692,0.9609,0.1061); rgb(252pt)=(0.9711,0.9667,0.1001); rgb(253pt)=(0.973,0.9724,0.0938); rgb(254pt)=(0.9749,0.9782,0.0872); rgb(255pt)=(0.9769,0.9839,0.0805)},
colorbar
]
\addplot [forget plot] graphics [xmin=-12.1420615720132, xmax=12.1420615720132, ymin=-12.1420628246185, ymax=12.1420628246185] {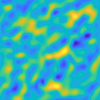};
\addplot [color=mycolor1, line width=1.0pt, forget plot]
  table[row sep=crcr]{%
9.62064095629306	-5.37733720384555\\
9.54954077347751	-4.6400813194855\\
9.34509549859762	-3.9194627777427\\
9.02271583630256	-3.22964148725436\\
8.600101143	-2.58170702001013\\
8.09650171505295	-1.98340074492785\\
7.53194363648938	-1.4389847320283\\
6.92644655552651	-0.94926086823213\\
6.29926480944675	-0.511737645851638\\
5.66818118592452	-0.120936220974944\\
5.04888034191613	0.231178179896973\\
4.45442558888979	0.554658385015126\\
3.8948585262442	0.860954018969469\\
3.37693602945526	1.16223323208053\\
2.90401357099291	1.47067602983174\\
2.47607798694104	1.79776892224463\\
2.08992683006262	2.15363054451894\\
1.73948560433359	2.54639665826321\\
1.41624868521555	2.98169057584278\\
1.10982480869271	3.462201651268\\
0.808563852614223	3.987390182429\\
0.500238398640696	4.55333203926947\\
0.172751378487678	5.15271076719526\\
-0.185159940421237	5.77495903209975\\
-0.583253246267293	6.40654530282157\\
-1.02917206674593	7.03139584214429\\
-1.52792413881667	7.63143662658554\\
-2.08146970787433	8.18723495178239\\
-2.68843785744779	8.67871639512573\\
-3.34398390458977	9.0859296611456\\
-4.03979530434646	9.38982975231225\\
-4.76424760733596	9.57304897144881\\
-5.50270603789064	9.62062551040414\\
-6.23796243996723	9.5206608055398\\
-6.95079190172225	9.26487939072707\\
-7.62060853013695	8.84906855600157\\
-8.22619579415339	8.27337958789875\\
-8.74648374780537	7.54247755423974\\
-9.16134340650074	6.66553230217551\\
-9.45236766120692	5.65604934375921\\
-9.60360841347137	4.53154537656198\\
-9.60224108957959	3.3130790934454\\
-9.43913028961018	2.02465344742303\\
-9.10927394799175	0.692510441596711\\
-8.61210788792095	-0.655656379138106\\
-7.9516578697602	-1.99154340738197\\
-7.13653196266292	-3.28710202380468\\
-6.17975208858032	-4.51539590513282\\
-5.09842966648079	-5.65141955532849\\
-3.91329618763772	-6.67284848853246\\
-2.64810505245693	-7.56069408190585\\
-1.32892588326322	-8.29983973290218\\
0.0166433927836849	-8.87943945913547\\
1.36031827841464	-9.29316529602563\\
2.67399281036825	-9.53929557642412\\
3.93059670920893	-9.62064219637229\\
5.10491356997374	-9.54432104980152\\
6.17433058937916	-9.32137571799673\\
7.11949294955542	-8.96626999960876\\
7.92483962457386	-8.49626975227504\\
8.5790019033491	-7.93073859981171\\
9.07505115577882	-7.29037518302313\\
9.41058810596426	-6.59642167827179\\
9.58767189539852	-5.86987419878676\\
9.61259328770619	-5.13072539685946\\
9.49550225069417	-4.39726811428386\\
9.24990562427323	-3.68548634290335\\
8.89205543348653	-3.00855615941738\\
8.44025244796337	-2.37647483040034\\
7.9140926676334	-1.7958311202252\\
7.3336864120125	-1.26972417857338\\
6.71888053137064	-0.797832456765685\\
6.08851391212662	-0.376628134974751\\
5.45973493150443	0.000273231432408316\\
4.84740688893641	0.341643493215708\\
4.26362380810014	0.658073490354403\\
3.71735450704334	0.96132972621438\\
3.21422765123577	1.26366337252595\\
2.75646483880263	1.57710071559974\\
2.34296284050197	1.91274487232796\\
1.96952016108838	2.28011814220442\\
1.62919733642982	2.68657273704199\\
1.31279505692537	3.13679490680806\\
1.00942952041194	3.63242376444834\\
0.707180549955575	4.17180154987025\\
0.393785114787387	4.74986684164703\\
0.057347079362053	5.35819652887437\\
-0.312966653009894	5.98519641756586\\
-0.726272913429557	6.61643439816855\\
-1.1893895865843	7.2351043758897\\
-1.70634832746899	7.82260388769761\\
-2.27802376324118	8.3592037059865\\
-2.90189566380719	8.82478394056666\\
-3.57195530037104	9.19960834706813\\
-4.27876149184632	9.46510684205029\\
-5.00964588799333	9.60463568159013\\
-5.74906108783415	9.60418540416947\\
-6.47905947488975	9.45300844746425\\
-7.17988539073236	9.14414125405151\\
-7.83065867130354	8.67479957235502\\
-8.41012381682913	8.04663038724266\\
-8.8974363037297	7.26580929787827\\
-9.27295588502914	6.3429779914689\\
-9.51901623107588	5.29302251566803\\
-9.62064095629306	4.13469909601427\\
-9.56617793540963	2.89012004456839\\
-9.34782676423577	1.58411763781353\\
-8.96203815307143	0.243508499841727\\
-8.40976880465671	-1.10371516922629\\
-7.69658074023559	-2.42923837688434\\
-6.83257988861541	-3.70528929812068\\
-5.83219481868546	-4.90548705780842\\
-4.71380254188944	-6.00564065937777\\
-3.49921410441806	-6.98447023040396\\
-2.21303800585695	-7.82422470074061\\
-0.881944116461955	-8.51117395830008\\
0.46614545980192	-9.0359582898366\\
1.80290936055318	-9.39378332528912\\
3.100492906995	-9.58445455553524\\
4.33235566217487	-9.61225155902113\\
5.47406997146867	-9.4856481187228\\
6.50404173507025	-9.21689020422238\\
7.40412764052468	-8.82144911124414\\
8.16012704078942	-8.31737168826212\\
8.76213144689155	-7.72455335874197\\
9.20472002903077	-7.06396242358513\\
9.48699537631951	-6.3568457957691\\
9.61245982704459	-5.62394681545548\\
9.58873871436796	-4.88476510188736\\
9.42716264402192	-4.15688654725434\\
9.1422262083575	-3.45540862138064\\
8.75094514155077	-2.7924822506805\\
8.2721376571439	-2.17698681396645\\
7.72565843832234	-1.61434944659725\\
7.13161536951895	-1.10651407267732\\
6.50959954439789	-0.652059618328823\\
5.87795834484482	-0.246460930632476\\
5.25313948921779	0.117519731797072\\
4.6491309716107	0.44932793859875\\
4.07701787497925	0.760020835477463\\
3.54467229472711	1.06160478652097\\
3.05658724207333	1.36633489879029\\
2.61385961857295	1.68600590936718\\
2.21432139072759	2.03126424051062\\
1.85281217992172	2.41097017455446\\
1.52158084914048	2.83163710014471\\
1.21079853421801	3.29697171355621\\
0.909161134294921	3.80753504300433\\
0.604555716899268	4.36053935694889\\
0.284762746650227	4.94979060715567\\
-0.0618353874734563	5.56578025512486\\
-0.445569336951593	6.19592436318842\\
-0.874850127891652	6.82494193246227\\
-1.35561568492433	7.43535886974204\\
-1.89088024490949	8.00811888369157\\
-2.48040629844919	8.52327824660388\\
-3.12051385170609	8.9607578831368\\
-3.80403636112839	9.30112379880308\\
-4.52042687167183	9.52636553587308\\
-5.25601191229368	9.62064219637229\\
-5.99438480436568	9.57096660895663\\
-6.71692444980859	9.36780040051654\\
-7.4034206049169	9.00553598170769\\
-8.03278231137433	8.48284564414761\\
-8.5838027181106	7.80288293508149\\
-9.0359511219472	6.97332703185926\\
-9.37016177694654	6.00626676955688\\
-9.5695889237875	4.91792705182512\\
-9.62029857735791	3.72824636303947\\
-9.51186984305852	2.46031976782823\\
-9.23788182554722	1.1397269130781\\
-8.79626642207437	-0.206231060686497\\
-8.18951229405872	-1.54935830691195\\
-7.42471089292042	-2.86142492508851\\
-6.51344136456545	-4.11503033630085\\
-5.47149724211745	-5.2844381218786\\
-4.3184638238903	-6.34635213721563\\
-3.07716079220466	-7.28060593581815\\
-1.7729697401046	-8.07074082640769\\
-0.433070640015131	-8.70445211154821\\
0.914385257676733	-9.17388805665454\\
2.24113620056654	-9.47579172035463\\
3.51967322084723	-9.61148172558587\\
4.72407468857199	-9.5866741360152\\
5.83078199845367	-9.41115358972869\\
6.8192885159864	-9.0983075009198\\
7.67271722810854	-8.66454225267285\\
8.37826679516826	-8.12860467336104\\
8.9275107201938	-7.51083554806036\\
9.31653994439363	-6.83238433253801\\
9.54594512830716	-6.11441551915548\\
9.62064095629306	-5.37733720384556\\
};
\end{axis}
\end{tikzpicture}%

%% file: fig/boxplot-mag.tex
%
%
\definecolor{mycolor1}{rgb}{0.00000,0.36471,0.55294}%
\begin{tikzpicture}

\begin{axis}[%
unbounded coords=jump,
xmin=0.5,
xmax=12.5,
xtick={1,2,3,4,5,6,7,8,9,10,11,12},
xticklabels={{EKF},{PF},{PS},{EKF},{PF},{PS},{EKF},{PF},{PS},{EKF},{PF},{PS}},
ymin=0,
ymax=0.4,
ytick={0, 0.1, 0.2, 0.3, 0.4},
ylabel style={font=\color{white!15!black}},
ylabel={RMSE in final SLAM path},
axis background/.style={fill=white},
width=\figurewidth,
height=\figureheight,
scale only axis
]

\addplot[area legend, draw=white!95!black, fill=white!90!black, forget plot]
table[row sep=crcr] {%
x	y\\
0.5	0\\
3.5	0\\
3.5	0.4\\
0.5	0.4\\
0.5	0\\
}--cycle;

\addplot[area legend, draw=white!95!black, fill=white!90!black, forget plot]
table[row sep=crcr] {%
x	y\\
6.5	0\\
9.5	0\\
9.5	0.4\\
6.5	0.4\\
6.5	0\\
}--cycle;
\addplot [color=mycolor1, forget plot]
  table[row sep=crcr]{%
1	0.134708447383603\\
1	0.178993363061122\\
};
\addplot [color=mycolor1, forget plot]
  table[row sep=crcr]{%
2	0.170606492141345\\
2	0.188133338189866\\
};
\addplot [color=mycolor1, forget plot]
  table[row sep=crcr]{%
3	0.124614541373392\\
3	0.15132839260034\\
};
\addplot [color=mycolor1, forget plot]
  table[row sep=crcr]{%
4	0.147754597569471\\
4	0.166659125294113\\
};
\addplot [color=mycolor1, forget plot]
  table[row sep=crcr]{%
5	0.151069877061738\\
5	0.174550601762206\\
};
\addplot [color=mycolor1, forget plot]
  table[row sep=crcr]{%
6	0.136571084357961\\
6	0.159757684920726\\
};
\addplot [color=mycolor1, forget plot]
  table[row sep=crcr]{%
7	0.214474839106715\\
7	0.256033840574501\\
};
\addplot [color=mycolor1, forget plot]
  table[row sep=crcr]{%
8	0.166280219359089\\
8	0.215816595164675\\
};
\addplot [color=mycolor1, forget plot]
  table[row sep=crcr]{%
9	0.114504437909787\\
9	0.127516304096007\\
};
\addplot [color=mycolor1, forget plot]
  table[row sep=crcr]{%
10	0.297551897380774\\
10	0.394576203024538\\
};
\addplot [color=mycolor1, forget plot]
  table[row sep=crcr]{%
11	0.159883939973454\\
11	0.196170768183959\\
};
\addplot [color=mycolor1, forget plot]
  table[row sep=crcr]{%
12	0.140997508111211\\
12	0.158444094087462\\
};
\addplot [color=mycolor1, forget plot]
  table[row sep=crcr]{%
1	0.0947820184504295\\
1	0.104764984419409\\
};
\addplot [color=mycolor1, forget plot]
  table[row sep=crcr]{%
2	0.10845117442936\\
2	0.122345122348075\\
};
\addplot [color=mycolor1, forget plot]
  table[row sep=crcr]{%
3	0.0926300442403827\\
3	0.105005429112276\\
};
\addplot [color=mycolor1, forget plot]
  table[row sep=crcr]{%
4	0.101267089145633\\
4	0.116219789006477\\
};
\addplot [color=mycolor1, forget plot]
  table[row sep=crcr]{%
5	0.0994766375253135\\
5	0.126617611777331\\
};
\addplot [color=mycolor1, forget plot]
  table[row sep=crcr]{%
6	0.0853226465421028\\
6	0.106054928339871\\
};
\addplot [color=mycolor1, forget plot]
  table[row sep=crcr]{%
7	0.103161506129855\\
7	0.12913754569139\\
};
\addplot [color=mycolor1, forget plot]
  table[row sep=crcr]{%
8	0.101766046718769\\
8	0.123139333484005\\
};
\addplot [color=mycolor1, forget plot]
  table[row sep=crcr]{%
9	0.0819507376644195\\
9	0.0963451144816952\\
};
\addplot [color=mycolor1, forget plot]
  table[row sep=crcr]{%
10	0.155908666015379\\
10	0.214581326511984\\
};
\addplot [color=mycolor1, forget plot]
  table[row sep=crcr]{%
11	0.117766487887697\\
11	0.129741703361657\\
};
\addplot [color=mycolor1, forget plot]
  table[row sep=crcr]{%
12	0.088882444273515\\
12	0.105825605368076\\
};
\addplot [color=mycolor1, forget plot]
  table[row sep=crcr]{%
0.875	0.178993363061122\\
1.125	0.178993363061122\\
};
\addplot [color=mycolor1, forget plot]
  table[row sep=crcr]{%
1.875	0.188133338189866\\
2.125	0.188133338189866\\
};
\addplot [color=mycolor1, forget plot]
  table[row sep=crcr]{%
2.875	0.15132839260034\\
3.125	0.15132839260034\\
};
\addplot [color=mycolor1, forget plot]
  table[row sep=crcr]{%
3.875	0.166659125294113\\
4.125	0.166659125294113\\
};
\addplot [color=mycolor1, forget plot]
  table[row sep=crcr]{%
4.875	0.174550601762206\\
5.125	0.174550601762206\\
};
\addplot [color=mycolor1, forget plot]
  table[row sep=crcr]{%
5.875	0.159757684920726\\
6.125	0.159757684920726\\
};
\addplot [color=mycolor1, forget plot]
  table[row sep=crcr]{%
6.875	0.256033840574501\\
7.125	0.256033840574501\\
};
\addplot [color=mycolor1, forget plot]
  table[row sep=crcr]{%
7.875	0.215816595164675\\
8.125	0.215816595164675\\
};
\addplot [color=mycolor1, forget plot]
  table[row sep=crcr]{%
8.875	0.127516304096007\\
9.125	0.127516304096007\\
};
\addplot [color=mycolor1, forget plot]
  table[row sep=crcr]{%
9.875	0.394576203024538\\
10.125	0.394576203024538\\
};
\addplot [color=mycolor1, forget plot]
  table[row sep=crcr]{%
10.875	0.196170768183959\\
11.125	0.196170768183959\\
};
\addplot [color=mycolor1, forget plot]
  table[row sep=crcr]{%
11.875	0.158444094087462\\
12.125	0.158444094087462\\
};
\addplot [color=mycolor1, forget plot]
  table[row sep=crcr]{%
0.875	0.0947820184504295\\
1.125	0.0947820184504295\\
};
\addplot [color=mycolor1, forget plot]
  table[row sep=crcr]{%
1.875	0.10845117442936\\
2.125	0.10845117442936\\
};
\addplot [color=mycolor1, forget plot]
  table[row sep=crcr]{%
2.875	0.0926300442403827\\
3.125	0.0926300442403827\\
};
\addplot [color=mycolor1, forget plot]
  table[row sep=crcr]{%
3.875	0.101267089145633\\
4.125	0.101267089145633\\
};
\addplot [color=mycolor1, forget plot]
  table[row sep=crcr]{%
4.875	0.0994766375253135\\
5.125	0.0994766375253135\\
};
\addplot [color=mycolor1, forget plot]
  table[row sep=crcr]{%
5.875	0.0853226465421028\\
6.125	0.0853226465421028\\
};
\addplot [color=mycolor1, forget plot]
  table[row sep=crcr]{%
6.875	0.103161506129855\\
7.125	0.103161506129855\\
};
\addplot [color=mycolor1, forget plot]
  table[row sep=crcr]{%
7.875	0.101766046718769\\
8.125	0.101766046718769\\
};
\addplot [color=mycolor1, forget plot]
  table[row sep=crcr]{%
8.875	0.0819507376644195\\
9.125	0.0819507376644195\\
};
\addplot [color=mycolor1, forget plot]
  table[row sep=crcr]{%
9.875	0.155908666015379\\
10.125	0.155908666015379\\
};
\addplot [color=mycolor1, forget plot]
  table[row sep=crcr]{%
10.875	0.117766487887697\\
11.125	0.117766487887697\\
};
\addplot [color=mycolor1, forget plot]
  table[row sep=crcr]{%
11.875	0.088882444273515\\
12.125	0.088882444273515\\
};
\addplot [color=mycolor1, forget plot]
  table[row sep=crcr]{%
0.875	0.104764984419409\\
0.875	0.134708447383603\\
1.125	0.134708447383603\\
1.125	0.104764984419409\\
0.875	0.104764984419409\\
};
\addplot [color=mycolor1, forget plot]
  table[row sep=crcr]{%
1.875	0.122345122348075\\
1.875	0.170606492141345\\
2.125	0.170606492141345\\
2.125	0.122345122348075\\
1.875	0.122345122348075\\
};
\addplot [color=mycolor1, forget plot]
  table[row sep=crcr]{%
2.875	0.105005429112276\\
2.875	0.124614541373392\\
3.125	0.124614541373392\\
3.125	0.105005429112276\\
2.875	0.105005429112276\\
};
\addplot [color=mycolor1, forget plot]
  table[row sep=crcr]{%
3.875	0.116219789006477\\
3.875	0.147754597569471\\
4.125	0.147754597569471\\
4.125	0.116219789006477\\
3.875	0.116219789006477\\
};
\addplot [color=mycolor1, forget plot]
  table[row sep=crcr]{%
4.875	0.126617611777331\\
4.875	0.151069877061738\\
5.125	0.151069877061738\\
5.125	0.126617611777331\\
4.875	0.126617611777331\\
};
\addplot [color=mycolor1, forget plot]
  table[row sep=crcr]{%
5.875	0.106054928339871\\
5.875	0.136571084357961\\
6.125	0.136571084357961\\
6.125	0.106054928339871\\
5.875	0.106054928339871\\
};
\addplot [color=mycolor1, forget plot]
  table[row sep=crcr]{%
6.875	0.12913754569139\\
6.875	0.214474839106715\\
7.125	0.214474839106715\\
7.125	0.12913754569139\\
6.875	0.12913754569139\\
};
\addplot [color=mycolor1, forget plot]
  table[row sep=crcr]{%
7.875	0.123139333484005\\
7.875	0.166280219359089\\
8.125	0.166280219359089\\
8.125	0.123139333484005\\
7.875	0.123139333484005\\
};
\addplot [color=mycolor1, forget plot]
  table[row sep=crcr]{%
8.875	0.0963451144816952\\
8.875	0.114504437909787\\
9.125	0.114504437909787\\
9.125	0.0963451144816952\\
8.875	0.0963451144816952\\
};
\addplot [color=mycolor1, forget plot]
  table[row sep=crcr]{%
9.875	0.214581326511984\\
9.875	0.297551897380774\\
10.125	0.297551897380774\\
10.125	0.214581326511984\\
9.875	0.214581326511984\\
};
\addplot [color=mycolor1, forget plot]
  table[row sep=crcr]{%
10.875	0.129741703361657\\
10.875	0.159883939973454\\
11.125	0.159883939973454\\
11.125	0.129741703361657\\
10.875	0.129741703361657\\
};
\addplot [color=mycolor1, forget plot]
  table[row sep=crcr]{%
11.875	0.105825605368076\\
11.875	0.140997508111211\\
12.125	0.140997508111211\\
12.125	0.105825605368076\\
11.875	0.105825605368076\\
};
\addplot [color=mycolor1, forget plot]
  table[row sep=crcr]{%
0.75	0.124844823393892\\
1.25	0.124844823393892\\
};
\addplot [color=mycolor1, forget plot]
  table[row sep=crcr]{%
1.75	0.140432442334196\\
2.25	0.140432442334196\\
};
\addplot [color=mycolor1, forget plot]
  table[row sep=crcr]{%
2.75	0.114948487705492\\
3.25	0.114948487705492\\
};
\addplot [color=mycolor1, forget plot]
  table[row sep=crcr]{%
3.75	0.126362637060196\\
4.25	0.126362637060196\\
};
\addplot [color=mycolor1, forget plot]
  table[row sep=crcr]{%
4.75	0.142198201758604\\
5.25	0.142198201758604\\
};
\addplot [color=mycolor1, forget plot]
  table[row sep=crcr]{%
5.75	0.118878691122415\\
6.25	0.118878691122415\\
};
\addplot [color=mycolor1, forget plot]
  table[row sep=crcr]{%
6.75	0.178738866025129\\
7.25	0.178738866025129\\
};
\addplot [color=mycolor1, forget plot]
  table[row sep=crcr]{%
7.75	0.147915151226\\
8.25	0.147915151226\\
};
\addplot [color=mycolor1, forget plot]
  table[row sep=crcr]{%
8.75	0.10234776069791\\
9.25	0.10234776069791\\
};
\addplot [color=mycolor1, forget plot]
  table[row sep=crcr]{%
9.75	0.26368108958602\\
10.25	0.26368108958602\\
};
\addplot [color=mycolor1, forget plot]
  table[row sep=crcr]{%
10.75	0.144080695072364\\
11.25	0.144080695072364\\
};
\addplot [color=mycolor1, forget plot]
  table[row sep=crcr]{%
11.75	0.127805217048059\\
12.25	0.127805217048059\\
};
\addplot [color=mycolor1, only marks, mark=+, mark options={solid, draw=red}, forget plot]
  table[row sep=crcr]{%
1	0.186993768638113\\
1	0.221783748840139\\
};
\addplot [color=mycolor1, only marks, mark=+, mark options={solid, draw=red}, forget plot]
  table[row sep=crcr]{%
nan	nan\\
};
\addplot [color=mycolor1, only marks, mark=+, mark options={solid, draw=red}, forget plot]
  table[row sep=crcr]{%
nan	nan\\
};
\addplot [color=mycolor1, only marks, mark=+, mark options={solid, draw=red}, forget plot]
  table[row sep=crcr]{%
4	0.224724932396514\\
4	0.24514952107459\\
};
\addplot [color=mycolor1, only marks, mark=+, mark options={solid, draw=red}, forget plot]
  table[row sep=crcr]{%
5	0.18906835705893\\
};
\addplot [color=mycolor1, only marks, mark=+, mark options={solid, draw=red}, forget plot]
  table[row sep=crcr]{%
nan	nan\\
};
\addplot [color=mycolor1, only marks, mark=+, mark options={solid, draw=red}, forget plot]
  table[row sep=crcr]{%
7	0.593635364365156\\
};
\addplot [color=mycolor1, only marks, mark=+, mark options={solid, draw=red}, forget plot]
  table[row sep=crcr]{%
nan	nan\\
};
\addplot [color=mycolor1, only marks, mark=+, mark options={solid, draw=red}, forget plot]
  table[row sep=crcr]{%
9	0.151864458143665\\
};
\addplot [color=mycolor1, only marks, mark=+, mark options={solid, draw=red}, forget plot]
  table[row sep=crcr]{%
10	0.478178955465915\\
};
\addplot [color=mycolor1, only marks, mark=+, mark options={solid, draw=red}, forget plot]
  table[row sep=crcr]{%
11	0.224716688909696\\
};
\addplot [color=mycolor1, only marks, mark=+, mark options={solid, draw=red}, forget plot]
  table[row sep=crcr]{%
nan	nan\\
};

\addplot[area legend, draw=mycolor1, fill=mycolor1, forget plot]
table[row sep=crcr] {%
x	y\\
2.875	0.105005429112276\\
2.875	0.124614541373392\\
3.125	0.124614541373392\\
3.125	0.105005429112276\\
2.875	0.105005429112276\\
}--cycle;

\addplot[area legend, draw=mycolor1, fill=mycolor1, forget plot]
table[row sep=crcr] {%
x	y\\
5.875	0.106054928339871\\
5.875	0.136571084357961\\
6.125	0.136571084357961\\
6.125	0.106054928339871\\
5.875	0.106054928339871\\
}--cycle;

\addplot[area legend, draw=mycolor1, fill=mycolor1, forget plot]
table[row sep=crcr] {%
x	y\\
8.875	0.0963451144816952\\
8.875	0.114504437909787\\
9.125	0.114504437909787\\
9.125	0.0963451144816952\\
8.875	0.0963451144816952\\
}--cycle;

\addplot[area legend, draw=mycolor1, fill=mycolor1, forget plot]
table[row sep=crcr] {%
x	y\\
11.875	0.105825605368076\\
11.875	0.140997508111211\\
12.125	0.140997508111211\\
12.125	0.105825605368076\\
11.875	0.105825605368076\\
}--cycle;
\node[align=center]
at (axis cs:2,0.38) {$o = 0$};
\node[align=center]
at (axis cs:5,0.38) {$o = 1$};
\node[align=center]
at (axis cs:8,0.38) {$o = 5$};
\node[align=center]
at (axis cs:11,0.38) {$o = 10$};
\end{axis}

\begin{axis}[%
xmin=0,
xmax=1,
ymin=0,
ymax=1,
axis line style={draw=none},
ticks=none,
axis x line*=bottom,
axis y line*=left,
width=\figurewidth,
height=\figureheight,
scale only axis
]
\end{axis}
\end{tikzpicture}%

%% file: fig/boxplot.tex
%
%
\definecolor{mycolor1}{rgb}{0.00000,0.36471,0.55294}%
\begin{tikzpicture}

\begin{axis}[%
unbounded coords=jump,
xmin=0.5,
xmax=20.5,
xtick={1,2,3,4,5,6,7,8,9,10,11,12,13,14,15,16,17,18,19,20},
xticklabels={{EKF},{PF},{EKS},{PS},{EKF},{PF},{EKS},{PS},{EKF},{PF},{EKS},{PS},{EKF},{PF},{EKS},{PS},{EKF},{PF},{EKS},{PS}},
xticklabel style={rotate=45},
ymin=0,
ymax=4,
ylabel style={font=\color{white!15!black}},
ylabel={RMSE in final SLAM path},
axis background/.style={fill=white}
]

\addplot[area legend, draw=white!95!black, fill=white!90!black, forget plot]
table[row sep=crcr] {%
x	y\\
0.5	0\\
4.5	0\\
4.5	4\\
0.5	4\\
0.5	0\\
}--cycle;

\addplot[area legend, draw=white!95!black, fill=white!90!black, forget plot]
table[row sep=crcr] {%
x	y\\
8.5	0\\
12.5	0\\
12.5	4\\
8.5	4\\
8.5	0\\
}--cycle;

\addplot[area legend, draw=white!95!black, fill=white!90!black, forget plot]
table[row sep=crcr] {%
x	y\\
16.5	0\\
20.5	0\\
20.5	4\\
16.5	4\\
16.5	0\\
}--cycle;
\addplot [color=mycolor1, forget plot]
  table[row sep=crcr]{%
1	0.201212684746366\\
1	0.225378089309081\\
};
\addplot [color=mycolor1, forget plot]
  table[row sep=crcr]{%
2	0.465188879505022\\
2	0.593896228971637\\
};
\addplot [color=mycolor1, forget plot]
  table[row sep=crcr]{%
3	0.0699164318271521\\
3	0.0781543042289592\\
};
\addplot [color=mycolor1, forget plot]
  table[row sep=crcr]{%
4	0.294547195598189\\
4	0.39783609464743\\
};
\addplot [color=mycolor1, forget plot]
  table[row sep=crcr]{%
5	0.789905482322303\\
5	1.15205891270714\\
};
\addplot [color=mycolor1, forget plot]
  table[row sep=crcr]{%
6	0.533761427672551\\
6	0.633147879128433\\
};
\addplot [color=mycolor1, forget plot]
  table[row sep=crcr]{%
7	0.439840747451059\\
7	0.760482188754949\\
};
\addplot [color=mycolor1, forget plot]
  table[row sep=crcr]{%
8	0.330910198096339\\
8	0.368844327401804\\
};
\addplot [color=mycolor1, forget plot]
  table[row sep=crcr]{%
9	1.38729714018565\\
9	1.82107588102734\\
};
\addplot [color=mycolor1, forget plot]
  table[row sep=crcr]{%
10	0.549286670715965\\
10	0.638160004621424\\
};
\addplot [color=mycolor1, forget plot]
  table[row sep=crcr]{%
11	0.847938624484079\\
11	1.69069399003407\\
};
\addplot [color=mycolor1, forget plot]
  table[row sep=crcr]{%
12	0.356503077063858\\
12	0.464173963647698\\
};
\addplot [color=mycolor1, forget plot]
  table[row sep=crcr]{%
13	1.88983950258309\\
13	2.163256165539\\
};
\addplot [color=mycolor1, forget plot]
  table[row sep=crcr]{%
14	0.742403166932043\\
14	0.929578025019634\\
};
\addplot [color=mycolor1, forget plot]
  table[row sep=crcr]{%
15	1.73880156556663\\
15	3.07076618018772\\
};
\addplot [color=mycolor1, forget plot]
  table[row sep=crcr]{%
16	0.392986652659016\\
16	0.447096739564454\\
};
\addplot [color=mycolor1, forget plot]
  table[row sep=crcr]{%
17	2.25583637230309\\
17	3.3136480552417\\
};
\addplot [color=mycolor1, forget plot]
  table[row sep=crcr]{%
18	1.07821975247756\\
18	1.29178820958696\\
};
\addplot [color=mycolor1, forget plot]
  table[row sep=crcr]{%
19	2.18407185920109\\
19	4.07261272894621\\
};
\addplot [color=mycolor1, forget plot]
  table[row sep=crcr]{%
20	0.614831711256669\\
20	0.749303763988976\\
};
\addplot [color=mycolor1, forget plot]
  table[row sep=crcr]{%
1	0.150380467941201\\
1	0.161878288900725\\
};
\addplot [color=mycolor1, forget plot]
  table[row sep=crcr]{%
2	0.311203274273146\\
2	0.36616912436275\\
};
\addplot [color=mycolor1, forget plot]
  table[row sep=crcr]{%
3	0.0490655443127494\\
3	0.0570215420841111\\
};
\addplot [color=mycolor1, forget plot]
  table[row sep=crcr]{%
4	0.154690085085368\\
4	0.220292693428581\\
};
\addplot [color=mycolor1, forget plot]
  table[row sep=crcr]{%
5	0.156246307276579\\
5	0.214612819272293\\
};
\addplot [color=mycolor1, forget plot]
  table[row sep=crcr]{%
6	0.275536269686901\\
6	0.363330505400861\\
};
\addplot [color=mycolor1, forget plot]
  table[row sep=crcr]{%
7	0.0739899454357007\\
7	0.0906841229407276\\
};
\addplot [color=mycolor1, forget plot]
  table[row sep=crcr]{%
8	0.183791849392802\\
8	0.225492907522627\\
};
\addplot [color=mycolor1, forget plot]
  table[row sep=crcr]{%
9	0.22480339612757\\
9	0.391682313250346\\
};
\addplot [color=mycolor1, forget plot]
  table[row sep=crcr]{%
10	0.360223783165039\\
10	0.421172064587149\\
};
\addplot [color=mycolor1, forget plot]
  table[row sep=crcr]{%
11	0.103326197654463\\
11	0.200555300587484\\
};
\addplot [color=mycolor1, forget plot]
  table[row sep=crcr]{%
12	0.215500780478672\\
12	0.282528949214905\\
};
\addplot [color=mycolor1, forget plot]
  table[row sep=crcr]{%
13	0.323855905075882\\
13	0.549272502858511\\
};
\addplot [color=mycolor1, forget plot]
  table[row sep=crcr]{%
14	0.348353627530024\\
14	0.481603546852736\\
};
\addplot [color=mycolor1, forget plot]
  table[row sep=crcr]{%
15	0.129100457119248\\
15	0.331153905271288\\
};
\addplot [color=mycolor1, forget plot]
  table[row sep=crcr]{%
16	0.203202523619603\\
16	0.300685359684869\\
};
\addplot [color=mycolor1, forget plot]
  table[row sep=crcr]{%
17	0.363507826882596\\
17	1.01263176433388\\
};
\addplot [color=mycolor1, forget plot]
  table[row sep=crcr]{%
18	0.413825873653882\\
18	0.562583417911194\\
};
\addplot [color=mycolor1, forget plot]
  table[row sep=crcr]{%
19	0.182087710326151\\
19	0.679531063906125\\
};
\addplot [color=mycolor1, forget plot]
  table[row sep=crcr]{%
20	0.191282985060182\\
20	0.325276677487548\\
};
\addplot [color=mycolor1, forget plot]
  table[row sep=crcr]{%
0.875	0.225378089309081\\
1.125	0.225378089309081\\
};
\addplot [color=mycolor1, forget plot]
  table[row sep=crcr]{%
1.875	0.593896228971637\\
2.125	0.593896228971637\\
};
\addplot [color=mycolor1, forget plot]
  table[row sep=crcr]{%
2.875	0.0781543042289592\\
3.125	0.0781543042289592\\
};
\addplot [color=mycolor1, forget plot]
  table[row sep=crcr]{%
3.875	0.39783609464743\\
4.125	0.39783609464743\\
};
\addplot [color=mycolor1, forget plot]
  table[row sep=crcr]{%
4.875	1.15205891270714\\
5.125	1.15205891270714\\
};
\addplot [color=mycolor1, forget plot]
  table[row sep=crcr]{%
5.875	0.633147879128433\\
6.125	0.633147879128433\\
};
\addplot [color=mycolor1, forget plot]
  table[row sep=crcr]{%
6.875	0.760482188754949\\
7.125	0.760482188754949\\
};
\addplot [color=mycolor1, forget plot]
  table[row sep=crcr]{%
7.875	0.368844327401804\\
8.125	0.368844327401804\\
};
\addplot [color=mycolor1, forget plot]
  table[row sep=crcr]{%
8.875	1.82107588102734\\
9.125	1.82107588102734\\
};
\addplot [color=mycolor1, forget plot]
  table[row sep=crcr]{%
9.875	0.638160004621424\\
10.125	0.638160004621424\\
};
\addplot [color=mycolor1, forget plot]
  table[row sep=crcr]{%
10.875	1.69069399003407\\
11.125	1.69069399003407\\
};
\addplot [color=mycolor1, forget plot]
  table[row sep=crcr]{%
11.875	0.464173963647698\\
12.125	0.464173963647698\\
};
\addplot [color=mycolor1, forget plot]
  table[row sep=crcr]{%
12.875	2.163256165539\\
13.125	2.163256165539\\
};
\addplot [color=mycolor1, forget plot]
  table[row sep=crcr]{%
13.875	0.929578025019634\\
14.125	0.929578025019634\\
};
\addplot [color=mycolor1, forget plot]
  table[row sep=crcr]{%
14.875	3.07076618018772\\
15.125	3.07076618018772\\
};
\addplot [color=mycolor1, forget plot]
  table[row sep=crcr]{%
15.875	0.447096739564454\\
16.125	0.447096739564454\\
};
\addplot [color=mycolor1, forget plot]
  table[row sep=crcr]{%
16.875	3.3136480552417\\
17.125	3.3136480552417\\
};
\addplot [color=mycolor1, forget plot]
  table[row sep=crcr]{%
17.875	1.29178820958696\\
18.125	1.29178820958696\\
};
\addplot [color=mycolor1, forget plot]
  table[row sep=crcr]{%
18.875	4.07261272894621\\
19.125	4.07261272894621\\
};
\addplot [color=mycolor1, forget plot]
  table[row sep=crcr]{%
19.875	0.749303763988976\\
20.125	0.749303763988976\\
};
\addplot [color=mycolor1, forget plot]
  table[row sep=crcr]{%
0.875	0.150380467941201\\
1.125	0.150380467941201\\
};
\addplot [color=mycolor1, forget plot]
  table[row sep=crcr]{%
1.875	0.311203274273146\\
2.125	0.311203274273146\\
};
\addplot [color=mycolor1, forget plot]
  table[row sep=crcr]{%
2.875	0.0490655443127494\\
3.125	0.0490655443127494\\
};
\addplot [color=mycolor1, forget plot]
  table[row sep=crcr]{%
3.875	0.154690085085368\\
4.125	0.154690085085368\\
};
\addplot [color=mycolor1, forget plot]
  table[row sep=crcr]{%
4.875	0.156246307276579\\
5.125	0.156246307276579\\
};
\addplot [color=mycolor1, forget plot]
  table[row sep=crcr]{%
5.875	0.275536269686901\\
6.125	0.275536269686901\\
};
\addplot [color=mycolor1, forget plot]
  table[row sep=crcr]{%
6.875	0.0739899454357007\\
7.125	0.0739899454357007\\
};
\addplot [color=mycolor1, forget plot]
  table[row sep=crcr]{%
7.875	0.183791849392802\\
8.125	0.183791849392802\\
};
\addplot [color=mycolor1, forget plot]
  table[row sep=crcr]{%
8.875	0.22480339612757\\
9.125	0.22480339612757\\
};
\addplot [color=mycolor1, forget plot]
  table[row sep=crcr]{%
9.875	0.360223783165039\\
10.125	0.360223783165039\\
};
\addplot [color=mycolor1, forget plot]
  table[row sep=crcr]{%
10.875	0.103326197654463\\
11.125	0.103326197654463\\
};
\addplot [color=mycolor1, forget plot]
  table[row sep=crcr]{%
11.875	0.215500780478672\\
12.125	0.215500780478672\\
};
\addplot [color=mycolor1, forget plot]
  table[row sep=crcr]{%
12.875	0.323855905075882\\
13.125	0.323855905075882\\
};
\addplot [color=mycolor1, forget plot]
  table[row sep=crcr]{%
13.875	0.348353627530024\\
14.125	0.348353627530024\\
};
\addplot [color=mycolor1, forget plot]
  table[row sep=crcr]{%
14.875	0.129100457119248\\
15.125	0.129100457119248\\
};
\addplot [color=mycolor1, forget plot]
  table[row sep=crcr]{%
15.875	0.203202523619603\\
16.125	0.203202523619603\\
};
\addplot [color=mycolor1, forget plot]
  table[row sep=crcr]{%
16.875	0.363507826882596\\
17.125	0.363507826882596\\
};
\addplot [color=mycolor1, forget plot]
  table[row sep=crcr]{%
17.875	0.413825873653882\\
18.125	0.413825873653882\\
};
\addplot [color=mycolor1, forget plot]
  table[row sep=crcr]{%
18.875	0.182087710326151\\
19.125	0.182087710326151\\
};
\addplot [color=mycolor1, forget plot]
  table[row sep=crcr]{%
19.875	0.191282985060182\\
20.125	0.191282985060182\\
};
\addplot [color=mycolor1, forget plot]
  table[row sep=crcr]{%
0.875	0.161878288900725\\
0.875	0.201212684746366\\
1.125	0.201212684746366\\
1.125	0.161878288900725\\
0.875	0.161878288900725\\
};
\addplot [color=mycolor1, forget plot]
  table[row sep=crcr]{%
1.875	0.36616912436275\\
1.875	0.465188879505022\\
2.125	0.465188879505022\\
2.125	0.36616912436275\\
1.875	0.36616912436275\\
};
\addplot [color=mycolor1, forget plot]
  table[row sep=crcr]{%
2.875	0.0570215420841111\\
2.875	0.0699164318271521\\
3.125	0.0699164318271521\\
3.125	0.0570215420841111\\
2.875	0.0570215420841111\\
};
\addplot [color=mycolor1, forget plot]
  table[row sep=crcr]{%
3.875	0.220292693428581\\
3.875	0.294547195598189\\
4.125	0.294547195598189\\
4.125	0.220292693428581\\
3.875	0.220292693428581\\
};
\addplot [color=mycolor1, forget plot]
  table[row sep=crcr]{%
4.875	0.214612819272293\\
4.875	0.789905482322303\\
5.125	0.789905482322303\\
5.125	0.214612819272293\\
4.875	0.214612819272293\\
};
\addplot [color=mycolor1, forget plot]
  table[row sep=crcr]{%
5.875	0.363330505400861\\
5.875	0.533761427672551\\
6.125	0.533761427672551\\
6.125	0.363330505400861\\
5.875	0.363330505400861\\
};
\addplot [color=mycolor1, forget plot]
  table[row sep=crcr]{%
6.875	0.0906841229407276\\
6.875	0.439840747451059\\
7.125	0.439840747451059\\
7.125	0.0906841229407276\\
6.875	0.0906841229407276\\
};
\addplot [color=mycolor1, forget plot]
  table[row sep=crcr]{%
7.875	0.225492907522627\\
7.875	0.330910198096339\\
8.125	0.330910198096339\\
8.125	0.225492907522627\\
7.875	0.225492907522627\\
};
\addplot [color=mycolor1, forget plot]
  table[row sep=crcr]{%
8.875	0.391682313250346\\
8.875	1.38729714018565\\
9.125	1.38729714018565\\
9.125	0.391682313250346\\
8.875	0.391682313250346\\
};
\addplot [color=mycolor1, forget plot]
  table[row sep=crcr]{%
9.875	0.421172064587149\\
9.875	0.549286670715965\\
10.125	0.549286670715965\\
10.125	0.421172064587149\\
9.875	0.421172064587149\\
};
\addplot [color=mycolor1, forget plot]
  table[row sep=crcr]{%
10.875	0.200555300587484\\
10.875	0.847938624484079\\
11.125	0.847938624484079\\
11.125	0.200555300587484\\
10.875	0.200555300587484\\
};
\addplot [color=mycolor1, forget plot]
  table[row sep=crcr]{%
11.875	0.282528949214905\\
11.875	0.356503077063858\\
12.125	0.356503077063858\\
12.125	0.282528949214905\\
11.875	0.282528949214905\\
};
\addplot [color=mycolor1, forget plot]
  table[row sep=crcr]{%
12.875	0.549272502858511\\
12.875	1.88983950258309\\
13.125	1.88983950258309\\
13.125	0.549272502858511\\
12.875	0.549272502858511\\
};
\addplot [color=mycolor1, forget plot]
  table[row sep=crcr]{%
13.875	0.481603546852736\\
13.875	0.742403166932043\\
14.125	0.742403166932043\\
14.125	0.481603546852736\\
13.875	0.481603546852736\\
};
\addplot [color=mycolor1, forget plot]
  table[row sep=crcr]{%
14.875	0.331153905271288\\
14.875	1.73880156556663\\
15.125	1.73880156556663\\
15.125	0.331153905271288\\
14.875	0.331153905271288\\
};
\addplot [color=mycolor1, forget plot]
  table[row sep=crcr]{%
15.875	0.300685359684869\\
15.875	0.392986652659016\\
16.125	0.392986652659016\\
16.125	0.300685359684869\\
15.875	0.300685359684869\\
};
\addplot [color=mycolor1, forget plot]
  table[row sep=crcr]{%
16.875	1.01263176433388\\
16.875	2.25583637230309\\
17.125	2.25583637230309\\
17.125	1.01263176433388\\
16.875	1.01263176433388\\
};
\addplot [color=mycolor1, forget plot]
  table[row sep=crcr]{%
17.875	0.562583417911194\\
17.875	1.07821975247756\\
18.125	1.07821975247756\\
18.125	0.562583417911194\\
17.875	0.562583417911194\\
};
\addplot [color=mycolor1, forget plot]
  table[row sep=crcr]{%
18.875	0.679531063906125\\
18.875	2.18407185920109\\
19.125	2.18407185920109\\
19.125	0.679531063906125\\
18.875	0.679531063906125\\
};
\addplot [color=mycolor1, forget plot]
  table[row sep=crcr]{%
19.875	0.325276677487548\\
19.875	0.614831711256669\\
20.125	0.614831711256669\\
20.125	0.325276677487548\\
19.875	0.325276677487548\\
};
\addplot [color=mycolor1, forget plot]
  table[row sep=crcr]{%
0.75	0.187371353633691\\
1.25	0.187371353633691\\
};
\addplot [color=mycolor1, forget plot]
  table[row sep=crcr]{%
1.75	0.406718530317351\\
2.25	0.406718530317351\\
};
\addplot [color=mycolor1, forget plot]
  table[row sep=crcr]{%
2.75	0.0610655334598717\\
3.25	0.0610655334598717\\
};
\addplot [color=mycolor1, forget plot]
  table[row sep=crcr]{%
3.75	0.269727679396264\\
4.25	0.269727679396264\\
};
\addplot [color=mycolor1, forget plot]
  table[row sep=crcr]{%
4.75	0.478508375341999\\
5.25	0.478508375341999\\
};
\addplot [color=mycolor1, forget plot]
  table[row sep=crcr]{%
5.75	0.450099804653308\\
6.25	0.450099804653308\\
};
\addplot [color=mycolor1, forget plot]
  table[row sep=crcr]{%
6.75	0.26634611436952\\
7.25	0.26634611436952\\
};
\addplot [color=mycolor1, forget plot]
  table[row sep=crcr]{%
7.75	0.26448803596721\\
8.25	0.26448803596721\\
};
\addplot [color=mycolor1, forget plot]
  table[row sep=crcr]{%
8.75	0.86035773864253\\
9.25	0.86035773864253\\
};
\addplot [color=mycolor1, forget plot]
  table[row sep=crcr]{%
9.75	0.450307283874421\\
10.25	0.450307283874421\\
};
\addplot [color=mycolor1, forget plot]
  table[row sep=crcr]{%
10.75	0.452614417971173\\
11.25	0.452614417971173\\
};
\addplot [color=mycolor1, forget plot]
  table[row sep=crcr]{%
11.75	0.320178916696779\\
12.25	0.320178916696779\\
};
\addplot [color=mycolor1, forget plot]
  table[row sep=crcr]{%
12.75	1.27479733879609\\
13.25	1.27479733879609\\
};
\addplot [color=mycolor1, forget plot]
  table[row sep=crcr]{%
13.75	0.553162163390452\\
14.25	0.553162163390452\\
};
\addplot [color=mycolor1, forget plot]
  table[row sep=crcr]{%
14.75	0.734985558433206\\
15.25	0.734985558433206\\
};
\addplot [color=mycolor1, forget plot]
  table[row sep=crcr]{%
15.75	0.353044077738919\\
16.25	0.353044077738919\\
};
\addplot [color=mycolor1, forget plot]
  table[row sep=crcr]{%
16.75	1.45465118312075\\
17.25	1.45465118312075\\
};
\addplot [color=mycolor1, forget plot]
  table[row sep=crcr]{%
17.75	0.756783069391329\\
18.25	0.756783069391329\\
};
\addplot [color=mycolor1, forget plot]
  table[row sep=crcr]{%
18.75	1.19587896001775\\
19.25	1.19587896001775\\
};
\addplot [color=mycolor1, forget plot]
  table[row sep=crcr]{%
19.75	0.42633624314031\\
20.25	0.42633624314031\\
};
\addplot [color=mycolor1, only marks, mark=+, mark options={solid, draw=red}, forget plot]
  table[row sep=crcr]{%
1	0.271605608538446\\
};
\addplot [color=mycolor1, only marks, mark=+, mark options={solid, draw=red}, forget plot]
  table[row sep=crcr]{%
nan	nan\\
};
\addplot [color=mycolor1, only marks, mark=+, mark options={solid, draw=red}, forget plot]
  table[row sep=crcr]{%
nan	nan\\
};
\addplot [color=mycolor1, only marks, mark=+, mark options={solid, draw=red}, forget plot]
  table[row sep=crcr]{%
4	0.41482530324253\\
};
\addplot [color=mycolor1, only marks, mark=+, mark options={solid, draw=red}, forget plot]
  table[row sep=crcr]{%
5	1.7440916683466\\
5	2.04334617175009\\
};
\addplot [color=mycolor1, only marks, mark=+, mark options={solid, draw=red}, forget plot]
  table[row sep=crcr]{%
nan	nan\\
};
\addplot [color=mycolor1, only marks, mark=+, mark options={solid, draw=red}, forget plot]
  table[row sep=crcr]{%
7	1.66143067591336\\
};
\addplot [color=mycolor1, only marks, mark=+, mark options={solid, draw=red}, forget plot]
  table[row sep=crcr]{%
nan	nan\\
};
\addplot [color=mycolor1, only marks, mark=+, mark options={solid, draw=red}, forget plot]
  table[row sep=crcr]{%
9	3.60946586007805\\
};
\addplot [color=mycolor1, only marks, mark=+, mark options={solid, draw=red}, forget plot]
  table[row sep=crcr]{%
10	0.742179790299045\\
10	0.916699497390951\\
};
\addplot [color=mycolor1, only marks, mark=+, mark options={solid, draw=red}, forget plot]
  table[row sep=crcr]{%
11	2.03888164974407\\
11	3.07531508098835\\
};
\addplot [color=mycolor1, only marks, mark=+, mark options={solid, draw=red}, forget plot]
  table[row sep=crcr]{%
nan	nan\\
};
\addplot [color=mycolor1, only marks, mark=+, mark options={solid, draw=red}, forget plot]
  table[row sep=crcr]{%
13	8.30106827655757\\
};
\addplot [color=mycolor1, only marks, mark=+, mark options={solid, draw=red}, forget plot]
  table[row sep=crcr]{%
14	1.22433674803077\\
};
\addplot [color=mycolor1, only marks, mark=+, mark options={solid, draw=red}, forget plot]
  table[row sep=crcr]{%
nan	nan\\
};
\addplot [color=mycolor1, only marks, mark=+, mark options={solid, draw=red}, forget plot]
  table[row sep=crcr]{%
nan	nan\\
};
\addplot [color=mycolor1, only marks, mark=+, mark options={solid, draw=red}, forget plot]
  table[row sep=crcr]{%
17	6.60617184317549\\
17	7.62438431903933\\
17	10.6452824197403\\
};
\addplot [color=mycolor1, only marks, mark=+, mark options={solid, draw=red}, forget plot]
  table[row sep=crcr]{%
18	2.38301159035218\\
};
\addplot [color=mycolor1, only marks, mark=+, mark options={solid, draw=red}, forget plot]
  table[row sep=crcr]{%
19	4.99298816749996\\
};
\addplot [color=mycolor1, only marks, mark=+, mark options={solid, draw=red}, forget plot]
  table[row sep=crcr]{%
20	1.28015403555592\\
20	1.32424437947007\\
};

\addplot[area legend, draw=mycolor1, fill=mycolor1, forget plot]
table[row sep=crcr] {%
x	y\\
2.875	0.0570215420841111\\
2.875	0.0699164318271521\\
3.125	0.0699164318271521\\
3.125	0.0570215420841111\\
2.875	0.0570215420841111\\
}--cycle;

\addplot[area legend, draw=mycolor1, fill=mycolor1, forget plot]
table[row sep=crcr] {%
x	y\\
3.875	0.220292693428581\\
3.875	0.294547195598189\\
4.125	0.294547195598189\\
4.125	0.220292693428581\\
3.875	0.220292693428581\\
}--cycle;

\addplot[area legend, draw=mycolor1, fill=mycolor1, forget plot]
table[row sep=crcr] {%
x	y\\
6.875	0.0906841229407276\\
6.875	0.439840747451059\\
7.125	0.439840747451059\\
7.125	0.0906841229407276\\
6.875	0.0906841229407276\\
}--cycle;

\addplot[area legend, draw=mycolor1, fill=mycolor1, forget plot]
table[row sep=crcr] {%
x	y\\
7.875	0.225492907522627\\
7.875	0.330910198096339\\
8.125	0.330910198096339\\
8.125	0.225492907522627\\
7.875	0.225492907522627\\
}--cycle;

\addplot[area legend, draw=mycolor1, fill=mycolor1, forget plot]
table[row sep=crcr] {%
x	y\\
10.875	0.200555300587484\\
10.875	0.847938624484079\\
11.125	0.847938624484079\\
11.125	0.200555300587484\\
10.875	0.200555300587484\\
}--cycle;

\addplot[area legend, draw=mycolor1, fill=mycolor1, forget plot]
table[row sep=crcr] {%
x	y\\
11.875	0.282528949214905\\
11.875	0.356503077063858\\
12.125	0.356503077063858\\
12.125	0.282528949214905\\
11.875	0.282528949214905\\
}--cycle;

\addplot[area legend, draw=mycolor1, fill=mycolor1, forget plot]
table[row sep=crcr] {%
x	y\\
14.875	0.331153905271288\\
14.875	1.73880156556663\\
15.125	1.73880156556663\\
15.125	0.331153905271288\\
14.875	0.331153905271288\\
}--cycle;

\addplot[area legend, draw=mycolor1, fill=mycolor1, forget plot]
table[row sep=crcr] {%
x	y\\
15.875	0.300685359684869\\
15.875	0.392986652659016\\
16.125	0.392986652659016\\
16.125	0.300685359684869\\
15.875	0.300685359684869\\
}--cycle;

\addplot[area legend, draw=mycolor1, fill=mycolor1, forget plot]
table[row sep=crcr] {%
x	y\\
18.875	0.679531063906125\\
18.875	2.18407185920109\\
19.125	2.18407185920109\\
19.125	0.679531063906125\\
18.875	0.679531063906125\\
}--cycle;

\addplot[area legend, draw=mycolor1, fill=mycolor1, forget plot]
table[row sep=crcr] {%
x	y\\
19.875	0.325276677487548\\
19.875	0.614831711256669\\
20.125	0.614831711256669\\
20.125	0.325276677487548\\
19.875	0.325276677487548\\
}--cycle;
\node[align=center]
at (axis cs:2.5,3.8) {$\sigma^2 = 0.01$};
\node[align=center]
at (axis cs:6.5,3.8) {$\sigma^2 = 1.00$};
\node[align=center]
at (axis cs:10.5,3.8) {$\sigma^2 = 1.69$};
\node[align=center]
at (axis cs:14.5,3.8) {$\sigma^2 = 2.56$};
\node[align=center]
at (axis cs:18.5,3.8) {$\sigma^2 = 4.00$};
\end{axis}
\end{tikzpicture}%

%% file: fig/dpo-odometry.tex
%
%
\begin{tikzpicture}

\begin{axis}[%
xmin=-5.0505,
xmax=12.5505,
xtick={-5,  0,  5, 10},
ymin=-9.3806,
ymax=22.371,
ytick={-5,  0,  5, 10, 15, 20},
axis line style={draw=none},
ticks=none,
axis x line*=bottom,
axis y line*=left,
width=\figurewidth,
height=\figureheight,
scale only axis
]
\addplot [color=black, forget plot]
  table[row sep=crcr]{%
0	0\\
-0.00979799999999997	0.0640840000000003\\
-0.0148160000000002	0.163538\\
-0.0185079999999997	0.258725\\
-0.0165950000000001	0.348149\\
-0.0161669999999998	0.447501\\
-0.0159629999999997	0.542149\\
-0.0178370000000001	0.643704\\
-0.0263619999999998	0.735078\\
-0.037277	0.820318\\
-0.0481560000000001	0.913494\\
-0.0563609999999999	1.004879\\
-0.0627	1.104922\\
-0.066865	1.193739\\
-0.068114	1.28105\\
-0.068997	1.368067\\
-0.074122	1.452102\\
-0.0741429999999998	1.540176\\
-0.0790850000000001	1.625466\\
-0.0792440000000001	1.705049\\
-0.083421	1.787677\\
-0.0846040000000001	1.865222\\
-0.0834730000000001	1.967034\\
-0.0808759999999999	2.052306\\
-0.0793879999999998	2.136159\\
-0.0762709999999998	2.221759\\
-0.0719439999999998	2.312907\\
-0.070811	2.401622\\
-0.0711689999999998	2.483483\\
-0.076838	2.561433\\
-0.08392	2.648431\\
-0.0899529999999999	2.735239\\
-0.0946029999999998	2.848355\\
-0.0940289999999999	2.940281\\
-0.0919080000000001	3.026839\\
-0.0908519999999999	3.112532\\
-0.0896279999999998	3.199146\\
-0.0897969999999999	3.289701\\
-0.09232	3.379028\\
-0.0968550000000001	3.460425\\
-0.104221	3.549008\\
-0.109246	3.634782\\
-0.117298	3.743464\\
-0.121138	3.837546\\
-0.122567	3.925686\\
-0.121788	4.017428\\
-0.121575	4.110022\\
-0.123663	4.20488\\
-0.126406	4.29347\\
-0.133256	4.375458\\
-0.140771	4.45918\\
-0.147763	4.541785\\
-0.152103	4.641112\\
-0.15212	4.731414\\
-0.151987	4.818904\\
-0.15265	4.90891\\
-0.15184	4.999915\\
-0.150538	5.09903\\
-0.150397	5.184606\\
-0.152324	5.265454\\
-0.152334	5.349217\\
-0.154778	5.436087\\
-0.163773	5.540436\\
-0.159303	5.630246\\
-0.14955	5.719195\\
-0.142314	5.809324\\
-0.133402	5.903724\\
-0.13028	6.004323\\
-0.129505	6.09654\\
-0.132239	6.187908\\
-0.130275	6.277512\\
-0.132872	6.369078\\
-0.134017	6.486836\\
-0.132427	6.585622\\
-0.124078	6.679664\\
-0.121058	6.77456\\
-0.117994	6.869114\\
-0.114722	6.968874\\
-0.118292	7.055861\\
-0.123524	7.139303\\
-0.125442	7.222268\\
-0.126039	7.310755\\
-0.128577	7.428479\\
-0.128205	7.517136\\
-0.122562	7.610133\\
-0.11643	7.701799\\
-0.108871	7.79974\\
-0.106239	7.90587\\
-0.112786	8.006814\\
-0.119217	8.105286\\
-0.122157	8.200728\\
-0.13131	8.297339\\
-0.139287	8.413581\\
-0.144336	8.515727\\
-0.146225	8.617829\\
-0.148688	8.719255\\
-0.148168	8.8173\\
-0.149625	8.916245\\
-0.15373	9.006726\\
-0.16282	9.096052\\
-0.17256	9.182078\\
-0.180302	9.275349\\
-0.191217	9.388142\\
-0.192264	9.481053\\
-0.193321	9.577734\\
-0.197289	9.672842\\
-0.197397	9.766248\\
-0.201171	9.865462\\
-0.208913	9.954997\\
-0.217768	10.042899\\
-0.223456	10.135341\\
-0.232544	10.230133\\
-0.240847	10.340627\\
-0.241773	10.43419\\
-0.239097	10.523858\\
-0.235346	10.61633\\
-0.232944	10.711421\\
-0.228781	10.810629\\
-0.230704	10.901546\\
-0.2352	10.98529\\
-0.238182	11.07364\\
-0.241674	11.162507\\
-0.245333	11.264009\\
-0.245126	11.357018\\
-0.245895	11.445748\\
-0.245258	11.530668\\
-0.241751	11.619389\\
-0.240278	11.710546\\
-0.243159	11.793839\\
-0.245373	11.875743\\
-0.245366	11.960127\\
-0.252334	12.045046\\
-0.259457	12.152541\\
-0.254901	12.252701\\
-0.248959	12.354291\\
-0.247537	12.447913\\
-0.244149	12.544575\\
-0.24167	12.635827\\
-0.244538	12.724433\\
-0.248046	12.817822\\
-0.251186	12.916708\\
-0.256673	13.01957\\
-0.259208	13.122402\\
-0.252812	13.218479\\
-0.250201	13.310938\\
-0.248105	13.396112\\
-0.246857	13.494157\\
-0.247742	13.582267\\
-0.2561	13.67188\\
-0.262489	13.752502\\
-0.270537	13.840309\\
-0.279458	13.940108\\
-0.279197	14.031813\\
-0.27059	14.129539\\
-0.264684	14.221162\\
-0.257924	14.319655\\
-0.251915	14.430303\\
-0.248418	14.523621\\
-0.245171	14.619468\\
-0.241849	14.703166\\
-0.239843	14.788041\\
-0.223366	14.875699\\
-0.20217	14.956523\\
-0.1708	15.036756\\
-0.13998	15.106812\\
-0.101701	15.180505\\
-0.058878	15.252637\\
-0.0168589999999997	15.318776\\
0.0224600000000001	15.370875\\
0.0697200000000002	15.420564\\
0.115963	15.467095\\
0.169845	15.515021\\
0.227712	15.556195\\
0.292849	15.587169\\
0.356812	15.606767\\
0.421871	15.62416\\
0.496563	15.639361\\
0.573881	15.647405\\
0.64765	15.658601\\
0.726147	15.671163\\
0.799874	15.680013\\
0.882164	15.692185\\
0.969261	15.702527\\
1.05643	15.713643\\
1.138767	15.71312\\
1.223787	15.706656\\
1.310169	15.698045\\
1.4016	15.682298\\
1.496206	15.671712\\
1.59034	15.668695\\
1.681871	15.662019\\
1.774968	15.655309\\
1.865747	15.646783\\
1.979172	15.636947\\
2.083173	15.627904\\
2.179191	15.613822\\
2.272263	15.597766\\
2.363002	15.582929\\
2.455744	15.567822\\
2.540127	15.563235\\
2.62442	15.556624\\
2.715261	15.550279\\
2.804191	15.552474\\
2.918807	15.555133\\
3.019862	15.552943\\
3.11131	15.544997\\
3.200134	15.537263\\
3.293962	15.529635\\
3.388246	15.525462\\
3.479418	15.525435\\
3.567544	15.52468\\
3.663972	15.527011\\
3.753423	15.529849\\
3.853123	15.527591\\
3.947219	15.521785\\
4.032073	15.513894\\
4.117932	15.507659\\
4.215216	15.508981\\
4.310299	15.505281\\
4.402844	15.49977\\
4.487387	15.489726\\
4.580382	15.480422\\
4.676962	15.469298\\
4.775278	15.45102\\
4.873522	15.420562\\
4.962533	15.390687\\
5.057476	15.365222\\
5.152647	15.338624\\
5.238304	15.30713\\
5.315601	15.267613\\
5.381452	15.220557\\
5.450842	15.161601\\
5.511306	15.099502\\
5.564229	15.034423\\
5.615915	14.960594\\
5.659028	14.883382\\
5.698835	14.806806\\
5.736594	14.731019\\
5.772387	14.652777\\
5.805544	14.583432\\
5.836627	14.511106\\
5.866232	14.439483\\
5.899793	14.365947\\
5.936118	14.286553\\
5.971075	14.197743\\
5.9987	14.106997\\
6.0251	14.018199\\
6.047598	13.930227\\
6.067608	13.835492\\
6.086042	13.746249\\
6.111506	13.657952\\
6.134193	13.573537\\
6.151738	13.485284\\
6.173837	13.395726\\
6.19405	13.290362\\
6.207542	13.193624\\
6.217331	13.100305\\
6.226204	13.0051\\
6.234784	12.91022\\
6.247595	12.814079\\
6.25908	12.725686\\
6.267641	12.638028\\
6.279555	12.550838\\
6.291775	12.462613\\
6.304208	12.34714\\
6.31086	12.250667\\
6.312031	12.155273\\
6.31316	12.061303\\
6.312203	11.964745\\
6.316369	11.864731\\
6.324365	11.77385\\
6.330544	11.684428\\
6.334822	11.591346\\
6.343509	11.508773\\
6.35381	11.396132\\
6.354857	11.286649\\
6.353971	11.182779\\
6.353325	11.080266\\
6.350397	10.976591\\
6.350822	10.873646\\
6.355723	10.779748\\
6.361033	10.688113\\
6.364543	10.590851\\
6.3733	10.497059\\
6.380189	10.381347\\
6.383759	10.269463\\
6.38363	10.166564\\
6.387983	10.059504\\
6.389012	9.950045\\
6.393812	9.832826\\
6.400742	9.723813\\
6.405604	9.618077\\
6.408638	9.513678\\
6.42092	9.412787\\
6.434423	9.29852\\
6.443167	9.186497\\
6.444187	9.084171\\
6.445435	8.97693\\
6.443011	8.868258\\
6.446676	8.748712\\
6.452387	8.639513\\
6.460765	8.526589\\
6.469395	8.410763\\
6.481265	8.301166\\
6.49422	8.175781\\
6.500141	8.054357\\
6.499579	7.938683\\
6.500123	7.824459\\
6.497032	7.715238\\
6.494739	7.608327\\
6.495411	7.511783\\
6.495046	7.418103\\
6.494386	7.322688\\
6.503334	7.229611\\
6.511951	7.110407\\
6.514287	6.999068\\
6.511134	6.891034\\
6.508762	6.785434\\
6.504668	6.677044\\
6.49892	6.560566\\
6.496853	6.452422\\
6.494565	6.347969\\
6.491385	6.239606\\
6.49531	6.139364\\
6.496216	6.01561\\
6.490043	5.8924\\
6.478208	5.777789\\
6.465625	5.667093\\
6.45195	5.557343\\
6.443559	5.444072\\
6.44124	5.341096\\
6.439189	5.24186\\
6.440258	5.139712\\
6.45119	5.045567\\
6.458878	4.934913\\
6.460735	4.823124\\
6.459575	4.714757\\
6.460984	4.603771\\
6.457581	4.493241\\
6.460129	4.376437\\
6.464515	4.265167\\
6.471509	4.159664\\
6.480539	4.053031\\
6.498185	3.93797\\
6.513391	3.82704\\
6.523065	3.706733\\
6.52933	3.589532\\
6.533684	3.477781\\
6.534831	3.371776\\
6.536148	3.264242\\
6.541914	3.163165\\
6.549799	3.058107\\
6.559541	2.948889\\
6.576925	2.836295\\
6.588713	2.733554\\
6.596551	2.627025\\
6.60229	2.524987\\
6.603853	2.419299\\
6.606288	2.312118\\
6.611639	2.216566\\
6.622812	2.13049\\
6.635253	2.04858\\
6.649963	1.967655\\
6.659849	1.86849\\
6.663585	1.777059\\
6.663949	1.675077\\
6.667602	1.579653\\
6.670366	1.472585\\
6.67127	1.357144\\
6.666449	1.255574\\
6.669963	1.153031\\
6.672623	1.054242\\
6.680796	0.952876\\
6.688972	0.827499\\
6.689819	0.720470000000001\\
6.684854	0.612803\\
6.681549	0.509252\\
6.677753	0.405853\\
6.67336	0.297415\\
6.67089	0.208935\\
6.674739	0.114199\\
6.675156	0.0213700000000001\\
6.683836	-0.0752689999999996\\
6.688295	-0.20203\\
6.690304	-0.305253\\
6.688982	-0.406224\\
6.688579	-0.50507\\
6.686695	-0.608128\\
6.686228	-0.724027\\
6.689067	-0.817671\\
6.69457	-0.918945\\
6.699966	-1.019988\\
6.715487	-1.114189\\
6.729486	-1.250894\\
6.734425	-1.362575\\
6.739349	-1.472422\\
6.742592	-1.576092\\
6.747256	-1.678963\\
6.75411	-1.78742\\
6.76449	-1.886519\\
6.773149	-1.988229\\
6.776552	-2.08966\\
6.78664	-2.183893\\
6.796841	-2.302525\\
6.798695	-2.407744\\
6.79898	-2.505759\\
6.80032	-2.609696\\
6.798843	-2.723407\\
6.798345	-2.83059\\
6.798457	-2.925022\\
6.794853	-3.026392\\
6.782389	-3.13192\\
6.768018	-3.237888\\
6.746346	-3.353345\\
6.70485	-3.463624\\
6.64725	-3.557216\\
6.577746	-3.639259\\
6.502505	-3.701288\\
6.42377	-3.753602\\
6.353408	-3.78951\\
6.2814	-3.824824\\
6.210732	-3.846034\\
6.142239	-3.867849\\
6.06831	-3.897839\\
5.991198	-3.93295\\
5.908971	-3.971053\\
5.820385	-4.007654\\
5.724533	-4.04489\\
5.61909	-4.0804\\
5.510172	-4.109728\\
5.410727	-4.139916\\
5.310013	-4.174906\\
5.214126	-4.206111\\
5.117634	-4.236557\\
5.024477	-4.256608\\
4.929642	-4.272346\\
4.831934	-4.282849\\
4.731318	-4.298025\\
4.631061	-4.31077\\
4.527654	-4.320042\\
4.432594	-4.329385\\
4.33406	-4.340571\\
4.239495	-4.347404\\
4.144585	-4.357196\\
4.03274	-4.36806\\
3.938157	-4.379745\\
3.834448	-4.383638\\
3.740682	-4.381136\\
3.644839	-4.379025\\
3.548192	-4.376895\\
3.459313	-4.374706\\
3.366566	-4.381321\\
3.276747	-4.38917\\
3.186837	-4.405663\\
3.077805	-4.420921\\
2.985295	-4.432597\\
2.879519	-4.435441\\
2.782405	-4.435274\\
2.678001	-4.432451\\
2.572119	-4.430077\\
2.473942	-4.433234\\
2.367701	-4.441004\\
2.27083	-4.447876\\
2.170555	-4.458925\\
2.055849	-4.469661\\
1.959334	-4.480919\\
1.857716	-4.480738\\
1.76551	-4.481024\\
1.667792	-4.483282\\
1.576172	-4.483989\\
1.488812	-4.486203\\
1.403565	-4.493093\\
1.318032	-4.498813\\
1.226454	-4.504622\\
1.12407	-4.507786\\
1.025259	-4.507234\\
0.927989	-4.497537\\
0.840636	-4.478091\\
0.744987	-4.446245\\
0.653915	-4.40264\\
0.571687	-4.349342\\
0.49899	-4.299498\\
0.429313	-4.247085\\
0.372697	-4.191232\\
0.321628	-4.133978\\
0.281244	-4.068727\\
0.249792	-3.994892\\
0.226198	-3.915255\\
0.200572	-3.837136\\
0.171505	-3.760228\\
0.138791	-3.67837\\
0.10487	-3.601524\\
0.0739040000000002	-3.516518\\
0.0465850000000003	-3.433692\\
0.0132020000000002	-3.345573\\
-0.0232649999999999	-3.250879\\
-0.0556839999999998	-3.150789\\
-0.0792799999999998	-3.052434\\
-0.0953710000000001	-2.957236\\
-0.108321	-2.858534\\
-0.121269	-2.756352\\
-0.134905	-2.652694\\
-0.152433	-2.554151\\
-0.170479	-2.462464\\
-0.191093	-2.371106\\
-0.208359	-2.279042\\
-0.223705	-2.176163\\
-0.236928	-2.073489\\
-0.246059	-1.972523\\
-0.251638	-1.868715\\
-0.2606	-1.767735\\
-0.268306	-1.659064\\
-0.279744	-1.564612\\
-0.292723	-1.470792\\
-0.309251	-1.370632\\
-0.325108	-1.279553\\
-0.339889	-1.177894\\
-0.350477	-1.073207\\
-0.359635	-0.975070999999999\\
-0.362029	-0.875807999999998\\
-0.367333	-0.779312999999998\\
-0.371529	-0.675113999999999\\
-0.38082	-0.580020999999999\\
-0.389172	-0.490273999999999\\
-0.396241	-0.397321999999999\\
-0.403382	-0.310612999999998\\
-0.413789	-0.209502999999999\\
-0.420118	-0.111608999999999\\
-0.422297	-0.0173219999999983\\
-0.422297	-0.0173219999999983\\
-0.421967	0.0719400000000014\\
-0.421933	0.163374000000002\\
-0.426322	0.246136000000002\\
-0.430306	0.326626000000002\\
-0.434115	0.412363000000002\\
-0.439298	0.494423000000001\\
-0.447797	0.594475000000001\\
-0.454174	0.677229000000001\\
-0.456705	0.759482000000002\\
-0.45675	0.838933000000001\\
-0.456469	0.914557000000002\\
-0.454466	0.999248000000001\\
-0.458757	1.071879\\
-0.463619	1.146378\\
-0.462789	1.225248\\
-0.466747	1.292042\\
-0.47222	1.388216\\
-0.478961	1.469829\\
-0.478955	1.553781\\
-0.482419	1.637967\\
-0.48484	1.720807\\
-0.491288	1.814868\\
-0.492893	1.899929\\
-0.499437	1.984856\\
-0.50504	2.066808\\
-0.515427	2.149387\\
-0.52628	2.255721\\
-0.532663	2.351204\\
-0.533216	2.448707\\
-0.534755	2.546398\\
-0.534763	2.64946\\
-0.539486	2.763634\\
-0.542819	2.870649\\
-0.551736	2.97327\\
-0.557134	3.069803\\
-0.561186	3.165184\\
-0.564399	3.269544\\
-0.569741	3.359748\\
-0.570576	3.453646\\
-0.571818	3.546112\\
-0.570885	3.640166\\
-0.576869	3.740886\\
-0.577127	3.840113\\
-0.580819	3.935832\\
-0.581472	4.025752\\
-0.586632	4.122497\\
-0.593738	4.230249\\
-0.597297	4.332839\\
-0.590798	4.439699\\
-0.586878	4.540878\\
-0.581913	4.64499\\
-0.581516	4.751271\\
-0.581192	4.849364\\
-0.582242	4.948287\\
-0.578999	5.041311\\
-0.578613	5.144055\\
-0.579318	5.252182\\
-0.583362	5.346222\\
-0.580565	5.445661\\
-0.577463	5.535787\\
-0.574233	5.631357\\
-0.574454	5.728212\\
-0.576976	5.815109\\
-0.580044	5.908399\\
-0.58145	5.996106\\
-0.589587	6.082342\\
-0.596015	6.19738\\
-0.59988	6.29322\\
-0.595919	6.391815\\
-0.591011	6.485981\\
-0.590057	6.58623\\
-0.588988	6.691691\\
-0.592122	6.784058\\
-0.597053	6.875477\\
-0.604118	6.96677\\
-0.609864	7.059858\\
-0.614278	7.173578\\
-0.617511	7.274301\\
-0.614488	7.369033\\
-0.610206	7.470222\\
-0.608324	7.570219\\
-0.606486	7.674834\\
-0.608671	7.771237\\
-0.606789	7.862187\\
-0.608209	7.958491\\
-0.610865	8.052893\\
-0.612937	8.151619\\
-0.607951	8.253042\\
-0.602745	8.347094\\
-0.595292	8.442562\\
-0.591641	8.538617\\
-0.590938	8.633868\\
-0.590021	8.731953\\
-0.586734	8.822228\\
-0.587507	8.913127\\
-0.590806	9.016765\\
-0.594979	9.105873\\
-0.590293	9.209232\\
-0.584706	9.304633\\
-0.578849	9.398743\\
-0.575053	9.488799\\
-0.575911	9.578617\\
-0.578692	9.669135\\
-0.581074	9.756575\\
-0.589725	9.848369\\
-0.595615	9.953881\\
-0.602247	10.045517\\
-0.602368	10.142713\\
-0.604195	10.233806\\
-0.603426	10.328838\\
-0.602626	10.430389\\
-0.603751	10.519769\\
-0.609037	10.607758\\
-0.610894	10.693844\\
-0.615591	10.77372\\
-0.624233	10.868213\\
-0.627797	10.960941\\
-0.624747	11.045707\\
-0.620384	11.129411\\
-0.615706	11.212758\\
-0.613226	11.2982\\
-0.61144	11.380191\\
-0.618268	11.46209\\
-0.622652	11.54066\\
-0.626488	11.622229\\
-0.633402	11.71012\\
-0.637476	11.793572\\
-0.634817	11.881968\\
-0.631234	11.966576\\
-0.626991	12.048598\\
-0.624538	12.134751\\
-0.624637	12.218877\\
-0.627531	12.299153\\
-0.630061	12.376006\\
-0.634213	12.458342\\
-0.643132	12.565649\\
-0.646334	12.656253\\
-0.644341	12.746615\\
-0.64559	12.831605\\
-0.64405	12.921397\\
-0.64141	13.012195\\
-0.631273	13.094867\\
-0.620000000000001	13.169562\\
-0.609791	13.244801\\
-0.598589	13.32376\\
-0.589244	13.402837\\
-0.570796000000001	13.485143\\
-0.536865000000001	13.564689\\
-0.486311000000001	13.629184\\
-0.424539	13.695302\\
-0.355080000000001	13.751445\\
-0.285060000000001	13.793116\\
-0.216099	13.829183\\
-0.153192000000001	13.856032\\
-0.0823980000000004	13.875567\\
-0.0194870000000003	13.89416\\
0.0429549999999996	13.91101\\
0.109197	13.923673\\
0.182779	13.930873\\
0.257133	13.93064\\
0.337947	13.92698\\
0.428878999999999	13.921077\\
0.522637	13.914769\\
0.610231	13.911969\\
0.700600999999999	13.904614\\
0.789072	13.896506\\
0.876491999999999	13.890317\\
0.972446999999999	13.881204\\
1.074489	13.874223\\
1.16838	13.866092\\
1.261659	13.857526\\
1.356502	13.846921\\
1.451276	13.836015\\
1.546339	13.825234\\
1.644704	13.818131\\
1.741878	13.815419\\
1.846627	13.813205\\
1.950973	13.811812\\
2.056097	13.810735\\
2.153706	13.806909\\
2.250663	13.796443\\
2.344963	13.783303\\
2.441913	13.772227\\
2.539711	13.762965\\
2.46809	13.416858\\
2.54866	13.417485\\
2.632479	13.417712\\
2.708573	13.417281\\
2.802666	13.415969\\
2.886325	13.413983\\
2.974243	13.410153\\
3.060804	13.406729\\
3.147583	13.402739\\
3.236668	13.399352\\
3.318771	13.401\\
3.402283	13.403703\\
3.479448	13.399939\\
3.554631	13.398568\\
3.648477	13.396811\\
3.734027	13.387945\\
3.828046	13.375157\\
3.919217	13.363759\\
4.011374	13.353815\\
4.106395	13.342378\\
4.188459	13.337344\\
4.276674	13.338513\\
4.359997	13.339896\\
4.449314	13.344174\\
4.55581	13.351426\\
4.640769	13.345416\\
4.73143	13.327903\\
4.81192	13.304892\\
4.892741	13.278424\\
4.979672	13.245467\\
5.059672	13.201724\\
5.134902	13.152427\\
5.195982	13.090635\\
5.253129	13.0194\\
5.305219	12.934689\\
5.343229	12.84957\\
5.374028	12.762105\\
5.394016	12.672213\\
5.415041	12.583461\\
5.441573	12.497702\\
5.470733	12.412446\\
5.495174	12.342393\\
5.519116	12.267415\\
5.538549	12.193494\\
5.5556	12.115103\\
5.569861	12.025634\\
5.584284	11.933114\\
5.59125	11.838866\\
5.595807	11.745252\\
5.600943	11.651969\\
5.607664	11.554761\\
5.614458	11.46223\\
5.627871	11.373944\\
5.634324	11.289294\\
5.642497	11.200146\\
5.652156	11.097036\\
5.660536	10.995463\\
5.663483	10.893183\\
5.662305	10.797144\\
5.6613	10.70402\\
5.660001	10.599734\\
5.66076	10.508059\\
5.667252	10.4144\\
5.671774	10.325904\\
5.68239	10.235775\\
5.693777	10.124862\\
5.700012	10.026317\\
5.701031	9.92244\\
5.701824	9.825931\\
5.701208	9.727818\\
5.701202	9.625751\\
5.703774	9.532634\\
5.710297	9.440347\\
5.711325	9.354926\\
5.71606	9.269702\\
5.723937	9.165996\\
5.730052	9.064521\\
5.733538	8.966362\\
5.739306	8.86832\\
5.7411	8.767096\\
5.746714	8.661782\\
5.756834	8.562773\\
5.771177	8.466221\\
5.782502	8.372524\\
5.795199	8.278952\\
5.810617	8.168779\\
5.823513	8.066094\\
5.833202	7.957778\\
5.843398	7.847865\\
5.850124	7.737179\\
5.855647	7.626498\\
5.861671	7.52197\\
5.87036	7.418283\\
5.876036	7.319255\\
5.883448	7.218381\\
5.893762	7.106716\\
5.901055	6.995521\\
5.904485	6.883093\\
5.906227	6.774426\\
5.90465	6.665975\\
5.903538	6.552809\\
5.901585	6.445486\\
5.903059	6.335563\\
5.898207	6.243763\\
5.893358	6.14817\\
5.896265	6.035762\\
5.900698	5.931113\\
5.897692	5.828163\\
5.890072	5.732311\\
5.881305	5.631853\\
5.87021	5.532262\\
5.863793	5.437409\\
5.864134	5.345698\\
5.864725	5.256723\\
5.868004	5.17486\\
5.87739	5.083866\\
5.885322	4.992474\\
5.888932	4.903967\\
5.890024	4.81356\\
5.890318	4.717189\\
5.889339	4.623742\\
5.886596	4.523559\\
5.884733	4.426613\\
5.883539	4.334855\\
5.884849	4.24314\\
5.892283	4.144494\\
5.900489	4.047871\\
5.907047	3.950017\\
5.90634	3.855058\\
5.910302	3.761463\\
5.914944	3.664296\\
5.91593	3.557611\\
5.917206	3.446491\\
5.926411	3.349402\\
5.93316	3.250885\\
5.946681	3.152634\\
5.966112	3.04432\\
5.974878	2.93764\\
5.983635	2.83256\\
5.991079	2.730511\\
5.995601	2.629763\\
6.00082	2.519441\\
6.00052	2.418621\\
6.007648	2.318147\\
6.010609	2.221894\\
6.016819	2.124696\\
6.027135	2.010971\\
6.034743	1.899968\\
6.038094	1.791961\\
6.038231	1.684848\\
6.038709	1.582354\\
6.037863	1.475684\\
6.034169	1.372798\\
6.03711	1.272841\\
6.040819	1.180838\\
6.046022	1.089437\\
6.054114	0.990570000000002\\
6.057666	0.888276000000001\\
6.061081	0.787956000000001\\
6.060896	0.690141000000001\\
6.060746	0.593535000000001\\
6.059802	0.497550000000001\\
6.058151	0.403802000000002\\
6.05697	0.314758000000001\\
6.061427	0.230212000000002\\
6.062631	0.147399000000002\\
6.06878	0.0658320000000021\\
6.081886	-0.0228629999999992\\
6.089341	-0.113883999999999\\
6.094366	-0.204595999999999\\
6.100133	-0.296106999999998\\
6.101399	-0.387207999999998\\
6.103054	-0.485247999999999\\
6.102509	-0.579758999999998\\
6.112974	-0.669411999999999\\
6.116888	-0.759928999999999\\
6.121987	-0.847284999999999\\
6.132745	-0.948937999999998\\
6.14128	-1.050674\\
6.146039	-1.146481\\
6.144852	-1.242609\\
6.144239	-1.338615\\
6.140732	-1.44004\\
6.138081	-1.538553\\
6.145969	-1.630389\\
6.151984	-1.719026\\
6.15564	-1.810732\\
6.165339	-1.902802\\
6.175968	-2.002323\\
6.181147	-2.101368\\
6.187799	-2.196107\\
6.194758	-2.290305\\
6.199536	-2.392728\\
6.203064	-2.488311\\
6.211987	-2.584732\\
6.22234	-2.677139\\
6.23363	-2.772772\\
6.248222	-2.868994\\
6.26557	-2.978711\\
6.279362	-3.081551\\
6.287087	-3.179784\\
6.298481	-3.276428\\
6.302844	-3.377712\\
6.306361	-3.478315\\
6.311168	-3.572392\\
6.319204	-3.665821\\
6.32215	-3.7591\\
6.329999	-3.852776\\
6.345163	-3.958581\\
6.35202	-4.058935\\
6.354029	-4.155476\\
6.357318	-4.25039\\
6.354464	-4.347272\\
6.351529	-4.450183\\
6.342836	-4.543109\\
6.340832	-4.634038\\
6.338381	-4.725259\\
6.342986	-4.818804\\
6.350066	-4.925904\\
6.353277	-5.028439\\
6.349398	-5.127229\\
6.345943	-5.223422\\
6.337243	-5.32256\\
6.319673	-5.434872\\
6.295077	-5.53458\\
6.271125	-5.626843\\
6.246278	-5.713361\\
6.220206	-5.79575\\
6.189488	-5.888837\\
6.145517	-5.971411\\
6.096908	-6.042748\\
6.036979	-6.100243\\
5.971099	-6.151146\\
5.904023	-6.192002\\
5.829875	-6.233523\\
5.769318	-6.27245\\
5.695714	-6.314969\\
5.624572	-6.352251\\
5.545161	-6.386038\\
5.465337	-6.415979\\
5.378315	-6.444985\\
5.284725	-6.466979\\
5.19169	-6.482762\\
5.098575	-6.495874\\
4.999463	-6.506721\\
4.896916	-6.528713\\
4.801309	-6.553669\\
4.703422	-6.577653\\
4.606199	-6.600101\\
4.509251	-6.621462\\
4.410997	-6.64495\\
4.312191	-6.660448\\
4.208077	-6.666077\\
4.109355	-6.669416\\
4.010426	-6.669743\\
3.91231	-6.67644\\
3.821189	-6.680223\\
3.728184	-6.690929\\
3.639341	-6.698894\\
3.551988	-6.707385\\
3.449108	-6.723174\\
3.35237	-6.736093\\
3.253116	-6.739616\\
3.156139	-6.741392\\
3.059847	-6.74434\\
2.957066	-6.749191\\
2.855996	-6.754397\\
2.758962	-6.764919\\
2.668907	-6.77424\\
2.576452	-6.784584\\
2.468797	-6.800688\\
2.365799	-6.817392\\
2.259143	-6.827058\\
2.156219	-6.824426\\
2.050803	-6.825516\\
1.946851	-6.824145\\
1.841285	-6.819957\\
1.742589	-6.813961\\
1.649915	-6.809834\\
1.556627	-6.806538\\
1.461849	-6.806076\\
1.360903	-6.808415\\
1.267239	-6.805495\\
1.171424	-6.799001\\
1.080136	-6.793798\\
0.989847999999999	-6.785534\\
0.900647999999999	-6.777104\\
0.814516	-6.761587\\
0.729355	-6.749924\\
0.643219	-6.731868\\
0.551458	-6.712268\\
0.451493	-6.690065\\
0.361073999999999	-6.654033\\
0.268626	-6.603908\\
0.188334999999999	-6.54524\\
0.109927	-6.474987\\
0.0343699999999996	-6.399355\\
-0.0302670000000003	-6.322548\\
-0.0871060000000003	-6.252737\\
-0.139507	-6.17917\\
-0.187297	-6.10656\\
-0.235013	-6.027128\\
-0.273807	-5.940973\\
-0.309632	-5.855207\\
-0.342112	-5.758754\\
-0.37252	-5.663516\\
-0.401483	-5.568506\\
-0.430061	-5.470518\\
-0.454364	-5.380934\\
-0.479602	-5.292828\\
-0.503171	-5.20394\\
-0.529727	-5.110043\\
-0.560499	-4.999894\\
-0.590282	-4.899094\\
-0.613236000000001	-4.79825\\
-0.629219	-4.699721\\
-0.643354	-4.602308\\
-0.654771	-4.504066\\
-0.668171	-4.406451\\
-0.687174	-4.314331\\
-0.70468	-4.221093\\
-0.721364	-4.126832\\
-0.737498	-4.036829\\
-0.754737	-3.93565\\
-0.772399000000001	-3.843594\\
-0.783471	-3.743333\\
-0.789229000000001	-3.648326\\
-0.797023	-3.551688\\
-0.81029	-3.450163\\
-0.818452000000001	-3.359482\\
-0.835681	-3.26775\\
-0.847826	-3.174706\\
-0.86116	-3.082698\\
-0.881812	-2.96512\\
-0.894567	-2.870339\\
-0.900335	-2.762207\\
-0.900335	-2.762207\\
};
\addplot [color=black, forget plot]
  table[row sep=crcr]{%
1.25	20.871\\
1.25	21.371\\
2.25	21.371\\
3.25	21.371\\
4.25	21.371\\
5.25	21.371\\
6.25	21.371\\
6.25	20.871\\
};
\node[align=center]
at (axis cs:3.75,20.371) {5 m};
\node[right, align=left]
at (axis cs:0.25,0.5) {$\uparrow$};
\end{axis}
\end{tikzpicture}%

%% file: fig/dpo-smoother.tex
%
%
\begin{tikzpicture}

\begin{axis}[%
point meta min=30,
point meta max=60,
axis on top,
xmin=-5.05050505050505,
xmax=12.5505050505051,
xtick={-5,  0,  5, 10},
y dir=reverse,
ymin=-9.38063206942724,
ymax=22.3710131261938,
ytick={-5,  0,  5, 10, 15, 20},
axis line style={draw=none},
ticks=none,
axis x line*=bottom,
axis y line*=left,
width=\figurewidth,
height=\figureheight,
scale only axis
]
\addplot [forget plot] graphics [xmin=-5.06274497408016, xmax=12.5627449740802, ymin=-9.3928536418659, ymax=22.3832346986325] {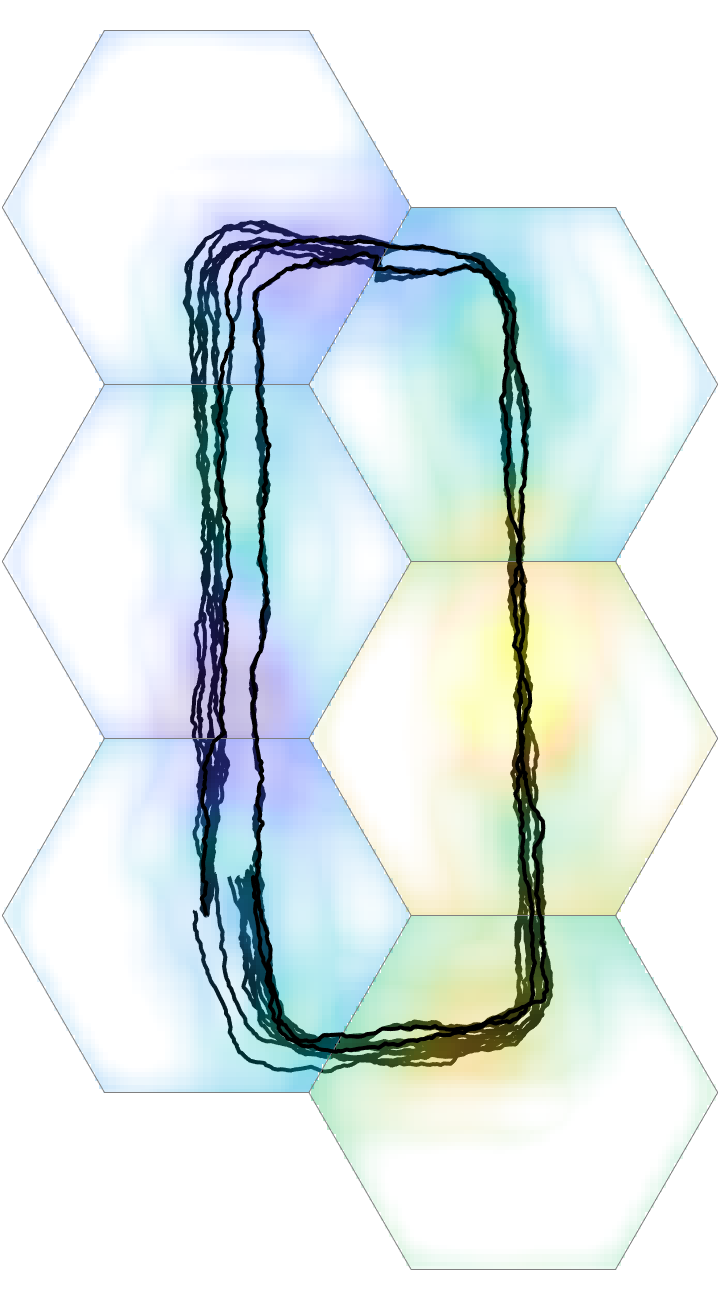};
\end{axis}

\begin{axis}[%
xmin=0,
xmax=1,
ymin=0,
ymax=1,
axis line style={draw=none},
ticks=none,
axis x line*=bottom,
axis y line*=left,
width=\figurewidth,
height=\figureheight,
scale only axis
]
\end{axis}
\end{tikzpicture}%